\documentclass[a4paper,fleqn,review]{cas-sc}

\usepackage{amssymb}
\usepackage{amsmath}
\usepackage{pdfpages}
\usepackage{makecell}  
\usepackage{graphicx}
\usepackage{lineno}
\usepackage{subcaption}
\usepackage[numbers,sort&compress]{natbib}
\usepackage{comment}
\usepackage[dvipsnames,table]{xcolor}
\usepackage{booktabs}      
\usepackage{caption}    
\usepackage{multirow}
\usepackage{placeins}   
\usepackage{colortbl}     
\definecolor{bestcell}{rgb}{0.85, 0.94, 0.83}   
\definecolor{rowgray}{rgb}{0.95, 0.95, 0.95}    
\usepackage{tabularx} 
\definecolor{ncblue}{HTML}{2166AC}
\definecolor{adorange}{HTML}{D6604D}
\usepackage{microtype}
\usepackage{hyperref}
\usepackage{float}
\usepackage{pgfplots}

\usepgfplotslibrary{groupplots}
\pgfplotsset{compat=1.18}

\pgfplotsset{
    colormap={pastelcka}{
        rgb255=(250,248,245)
        rgb255=(236,228,242)
        rgb255=(215,226,242)
        rgb255=(198,226,216)
        rgb255=(239,217,183)
        rgb255=(199,143,164)
    },
}
\def\tsc#1{\csdef{#1}{\textsc{\lowercase{#1}}\xspace}}
\tsc{WGM}
\tsc{QE}

\begin{document}
\let\WriteBookmarks\relax

\shorttitle{}    

\shortauthors{Ergün et al.}  

\title [mode = title]{Masked and Predictive Self-Supervised Pretraining for Brain MRI: Effects of Objective Design on Disease Detection}  



%

\author[1]{Esra Ergün}[orcid=0000-0001-5000-8543]

\cormark[1]

\ead{ergunesr@itu.edu.tr}

\credit{ Conceptualization, Methodology, Software, Formal Analysis, Investigation, Validation, Visualization, Writing – Original Draft, Writing – Review \& Editing}

\affiliation[1]{organization={Department of Computer Engineering, Istanbul Technical University},
            postcode={34469}, 
            state={Istanbul},
            country={Turkey}}

\affiliation[2]{organization={Department of Radiology, NYU Grossman School of Medicine},
            postcode={10016}, 
            state={New York},
            country={United States}}

\author[2]{Hersh Chandarana}[orcid=0000-0002-6807-4589]

\ead{hersh.chandarana@nyulangone.org}


\credit{Project Administration, Funding Acquisition}

\author[2]{Dan Sodickson}

\ead{daniel.sodickson@nyulangone.org}

\credit{Conceptualization, Supervision, Project Administration, Funding Acquisition}


\author[3]{Gozde Unal}[orcid=0000-0001-5942-8966]

\ead{unalgo@itu.edu.tr}

\credit{Conceptualization, Methodology, Software, Investigation, Funding Acquisition, Project Administration, Supervision, Writing – Original Draft, Writing – Review \& Editing}

\affiliation[3]{
    organization={Department of AI and Data Engineering, Istanbul Technical University},
    postcode={34469},
    city={Istanbul},
    country={Türkiye}
}

\cortext[1]{Corresponding author}



\begin{abstract}
MRI self-supervised learning (SSL) research has largely emphasized segmentation and 
dense prediction, while controlled studies examining how alternative 
pretraining objectives shape transferable representations for MRI-based 
disease detection remain comparatively scarce. In this study, we compare 
reconstruction-based learning via Masked Autoencoders (MAE) and latent-predictive 
learning via Joint Embedding Predictive Architectures (JEPA) for 3D structural MRI. We study 
auxiliary objectives by introducing a spectral-domain 
reconstruction loss for MAE to promote sensitivity to fine-grained anatomical 
structure and integrating variance--covariance regularization (VCR) within JEPA 
to encourage broader, less redundant latent representations.

Models are pretrained on heterogeneous MRI datasets using individual 
single-contrast inputs. Across seven dataset--task settings spanning four clinical 
problem domains, we find that downstream transfer depends not only on the broad 
pretraining paradigm but also on how the representational bias induced by a 
specific objective aligns with the downstream task. Spectral regularization 
provides its clearest gains for Alzheimer's disease classification and tumor 
grading, consistent with sensitivity
to fine anatomical detail, while its benefit is smaller for NC vs.\ MCI and ASD. 
VCR yields its strongest improvements when discrimination appears to involve sources of structural variation. Overall, MAE-based pretraining generally outperforms baseline JEPA, with SL-MAE providing the strongest 
gains in several disease-detection settings.

Representation analyses show that fine-tuning reorganizes learned 
representations in an objective-, depth-, and task-dependent manner. VCR-JEPA 
maintains broader effective dimensionality across depth and exhibits greater 
cross-task representational alignment than baseline JEPA, suggesting greater representational reuse. We publicly release all subject-level 
pretraining and downstream evaluation splits to support reproducible comparison.
\end{abstract}



\begin{keywords}
Self-Supervised Learning \sep
Representation Learning \sep
3D Brain MRI \sep
Masked Autoencoders \sep
Joint Embedding Predictive Architectures \sep
Disease Classification
\end{keywords}

\maketitle

\section{Introduction}
Self-supervised learning (SSL) methods trained on large-scale unlabeled data
have transformed computer vision and are increasingly influencing medical
imaging by reducing reliance on costly and labor-intensive annotations. This is
particularly important in clinical neuroimaging, where high-quality labels are
scarce, expensive to obtain, and often require specialized medical expertise.

Recent work has demonstrated strong results in 3D computed tomography (CT),
with SSL models trained on hundreds of thousands of scans achieving broad
transfer across clinical tasks \cite{zhu2025foundation}. Interest in
self-supervised learning for MRI has also grown rapidly. However, much of the
existing MRI SSL literature has focused on segmentation and dense prediction,
while systematic evaluation of transferable representations across
heterogeneous disease-detection and clinical classification tasks remains
comparatively limited. MRI additionally presents substantial acquisition
heterogeneity, including multiple contrast weightings (e.g., T1, T2, FLAIR,
and T2*) and anisotropic spatial resolutions that vary across protocols,
datasets, and institutions, complicating robust representation learning across
heterogeneous sources.

Within the SSL literature, two paradigms have gained particular traction.
Masked Autoencoders (MAE) \cite{he2022masked} learn by reconstructing masked
image regions in voxel space and have been widely adopted across vision tasks.
More recently, Joint Embedding Predictive Architectures (JEPA)
\cite{assran2023ijepa} have emerged as an alternative that predicts latent
representations of masked targets rather than reconstructing raw voxel values. The
JEPA framework has since been extended beyond natural images to video
\cite{drozdov2024videorepresentationlearningjointembedding},
ultrasound \cite{usjepa}, and brain signals \cite{brainjepa}. Concurrent work
has also demonstrated the potential of neuroimaging-specific JEPA adaptations
at scale \cite{huang}. Despite this progress, controlled comparisons of MAE
and JEPA as alternative pretraining paradigms for MRI-based disease detection
remain limited.

Beyond the choice of pretraining paradigm, an important open question is how
objective design influences the properties of learned representations and
their utility for downstream clinical tasks. Auxiliary objectives are
frequently introduced to modify optimization or representation structure, yet
their benefits may depend on the statistical characteristics of the target
problem and on the anatomical information emphasized during pretraining. This
is particularly relevant in MRI, where clinically discriminative signals may
range from subtle and spatially diffuse alterations to localized,
fine-grained, or high-frequency pathological structure. An auxiliary objective
that benefits one task may therefore provide substantially less benefit for
another.

Many recent medical SSL efforts emphasize either scaling models and datasets
or introducing new architectural and objective-level enhancements. Aggregate
benchmark improvements, however, do not reveal whether a particular
pretraining choice benefits qualitatively different downstream tasks in the
same way. Comparatively less attention has been devoted to characterizing how
reconstruction-based, latent-predictive, and auxiliary objectives interact
with downstream problems whose discriminative signals differ in spatial
extent, frequency content, and pathological heterogeneity. Understanding this
variation is important both for interpreting existing SSL results and for
designing pretraining strategies whose representational biases are well
matched to the intended target task.

Motivated by this gap, we conduct a controlled paradigm-level comparison of
reconstruction-based MAE and latent-predictive JEPA in their defining
formulations. Rather than seeking a single pretraining configuration that
maximizes average performance across a broad benchmark, we examine how the two
paradigms and their auxiliary objectives affect transfer across clinical tasks
with distinct anatomical and pathological characteristics. In addition to the
baseline formulations, we introduce a spectral-domain reconstruction loss for
MAE to promote sensitivity to fine-grained anatomical structure and
incorporate variance--covariance regularization (VCR) into JEPA to encourage feature variance while reducing redundancy across latent dimensions. We evaluate all models
under both frozen and fine-tuned transfer settings. Frozen evaluation assesses
the transferability of pretrained representations while keeping the encoder
fixed, whereas fine-tuning reveals how those representations reorganize under
task-specific supervision. Finally, we directly analyze the learned
representation spaces using centered kernel alignment (CKA) and Shannon
effective rank to characterize depth-wise representational reorganization and
changes in effective dimensional structure during downstream adaptation.

Our framework is pretrained on heterogeneous MRI volumes using individual
single-contrast inputs (T1-weighted, T2-weighted, FLAIR, or T2*, depending on
the dataset) without contrast concatenation. We employ randomized native
in-plane orientation sampling, standardized cropping, and depth interpolation,
while intentionally avoiding spatial normalization, bias-field correction,
and intensity harmonization. This design enables pretraining across
heterogeneous MRI datasets without requiring simultaneous availability of
multiple contrasts. The pretraining corpus comprises seven datasets: ADNI,
NACC/SCAN, PPMI, OASIS-3, IXI, MOOD, and BraTS-2024
\cite{adni_dataset, scan_dataset, ppmi_dataset, oasis3_dataset, ixi_dataset,
mood_dataset, BraTS24_dataset}. Downstream evaluation is conducted on ADNI,
NACC/SCAN, UCSF, and ABIDE
\cite{adni_dataset, scan_dataset, ucsf_dataset, abide_dataset}. We publicly
release the code and subject-level data splits at
\url{https://github.com/ituvisionlab/mjepa}.

Our core contributions are as follows:
\begin{itemize}

\item A controlled comparison of reconstruction-based MAE and latent-predictive
JEPA for MRI-based disease detection, preserving the defining formulation of
each framework to isolate differences associated with their pretraining
mechanisms.

\item A novel spectral-domain auxiliary loss for MAE that promotes sensitivity
to high-frequency anatomical structure, together with the integration of
variance--covariance regularization into JEPA to support more distributed
feature usage and lower redundancy across latent dimensions.

\item A systematic downstream evaluation demonstrating that the effects of SSL
objective design are task-dependent, with performance patterns indicating that
different objectives emphasize different forms of clinically relevant
anatomical information.

\item A representation-level analysis using CKA and Shannon effective rank,
showing that pretraining objective, encoder depth, and downstream task jointly
shape representational reorganization and effective dimensional structure
during adaptation.

\item A contrast-agnostic single-input MRI pretraining framework operating
across heterogeneous acquisition settings without contrast concatenation,
spatial normalization, or intensity harmonization, together with strict
subject-level separation between pretraining and downstream cohorts and public
release of all dataset partitions.

\end{itemize}
 
\section{Related Work}

\subsection{Self-Supervised Pretraining Paradigms in 3D Medical Imaging}

Self-supervised learning in 3D medical imaging has developed along several
complementary directions, including reconstruction-based learning, contrastive
and invariance-based learning, self-distillation, and latent predictive learning.
Early benchmarking studies compared classical pretext tasks such as rotation
prediction, jigsaw puzzles, and contrastive predictive coding across MRI and CT
datasets \cite{taleb-neurips2020}. More recent approaches have increasingly
adopted transformer architectures and large-scale masked or distillation-based
pretraining.

In CT, Swin UNETR combines masked volume inpainting, rotation prediction,
and contrastive learning for volumetric segmentation
\cite{tang2022swintransformers}. Large-scale CT pretraining studies have
subsequently demonstrated the benefits of training on hundreds of thousands of
clinical scans for broad downstream transfer and out-of-distribution
generalization \cite{zhu2025foundation}. Other work has investigated
properties such as demographic information leakage and fairness in pretrained
medical representations \cite{zheng2025fair}. Together, these studies demonstrate
the scalability and transferability of volumetric self-supervised pretraining
across medical imaging tasks.

Alternative masked pretraining strategies have also questioned whether raw voxel reconstruction provides the most effective target for medical representation learning. FedATA, for example, replaces low-level pixel reconstruction with masked self-distillation toward latent representations in a federated volumetric medical-imaging setting \cite{fedata}.

Masked autoencoding has also become an important paradigm for 3D MRI.
SPARK3D applies MAE pretraining to large-scale brain MRI and demonstrates
consistent gains over training from scratch across multiple segmentation
datasets \cite{wald2024spark3d}. The study further shows that transferring
both encoder and decoder parameters can benefit dense downstream prediction.
By contrast, image-level disease classification typically relies on encoder
representations rather than the reconstruction decoder, motivating evaluation
of the pretrained encoder under both frozen and fine-tuned settings.

Latent predictive learning provides a complementary alternative to
observation-space reconstruction. JEPA predicts representations of masked or
unobserved targets from visible context and has been extended from natural
images to video, ultrasound, and brain signals
\cite{assran2023ijepa,drozdov2024videorepresentationlearningjointembedding,usjepa,brainjepa}.
Despite the increasing adoption of both MAE and JEPA, controlled comparisons
of their defining reconstruction-based and latent-predictive objectives remain
limited for 3D structural MRI disease detection.

\subsection{MRI-Specific and Multimodal Representation Learning}

MRI introduces substantial heterogeneity in contrast weighting, spatial
resolution, orientation, and acquisition protocol. Recent MRI-specific methods
have addressed this variability through multimodal reconstruction, contrast
alignment, and large-scale sequence-aware pretraining. BrainMVP uses
cross-modal reconstruction and modality-aware distillation across predefined
MRI modality templates \cite{rui2025brainmvp}, whereas sequence-invariant
contrastive learning has been explored using simulated quantitative MRI
contrasts to improve cross-protocol generalization \cite{chalcroft2025}.
BrainFound adopts large-scale DINO-style self-distillation for structural brain
MRI and demonstrates strong transfer across multiple downstream neuroimaging
tasks \cite{Mazher2025}. Its use of a large transformer backbone,
natural-image initialization, and multimodal MRI inputs provides a complementary
example of scaling invariance-based representation learning for brain imaging. BrainIAC uses contrastive self-supervised pretraining on heterogeneous
structural brain MRI with a single-sequence encoder and demonstrates
transfer to clinical classification tasks, including MCI versus healthy
controls on OASIS-1 and IDH mutation prediction on UCSF-PDGM
\cite{Tak2026}. It provides a complementary example of learning
transferable representations across MRI contrasts without requiring
their simultaneous fusion as input. Relatedly, MRBC employs a self-supervised masked convolutional autoencoder with a morphology-aware reconstruction objective that combines local normalized cross-correlation and mean squared error to preserve fine-grained intensity and textural information in structural MRI representations \cite{yu2026mrbc}. These representations are subsequently used for brain-connectome construction and downstream clinical prediction.

Several works have also investigated general-purpose 3D medical
representations. 3DINO scales self-distillation across more than 100,000 scans
covering multiple organs and imaging modalities \cite{xu20253dino}, while the
Medical Slice Transformer adapts 2D DINO-style models to volumetric medical
data \cite{mullerfranzes2024}. Other approaches exploit local and global
slice relationships when transferring 2D models to 3D data
\cite{li-2024-isbi}. Comparisons among 2D, 2.5D, and 3D approaches generally
support the value of fully volumetric modeling for tasks requiring coherent
three-dimensional anatomical representations \cite{avesta2023}.

Cross-modal approaches have pursued complementary information across
modalities or views. CADS aligns 3D CT with generated 2D radiographic views
and combines this alignment with self-distillation
\cite{ye2025cads}. DrasCLR incorporates disease- and anatomy-specific
conditioning into contrastive learning for lung CT
\cite{yu2024drasclr}. Wild pretraining on diverse, uncurated medical datasets
has further demonstrated that heterogeneous pretraining data can improve
robustness beyond narrowly matched source--target settings
\cite{jiang2024self}. In contrast to explicit multimodal fusion, the present
study processes each MRI contrast independently and investigates whether
representations learned from heterogeneous single-contrast volumes transfer
across disease-detection tasks.

\subsection{Benchmarking Transfer in Medical Self-Supervised Learning}

Recent benchmarks have highlighted the importance of both downstream task
composition and evaluation protocol when assessing medical SSL. MRI-specific
foundation models trained on millions of 2D slices have reported strong
segmentation and probing performance, although many classification targets
represent acquisition metadata, anatomical location, or coarse severity
attributes rather than disease-specific discrimination
\cite{dong2025mricoreFM}.

The OpenMind benchmark provides a large-scale resource for comparing 3D
medical SSL methods across segmentation and selected classification tasks
\cite{wald2025openmind}. Its results suggest that reconstruction-based
methods are particularly effective for dense prediction, whereas objectives
favoring global representations may be advantageous for image-level
classification. However, the benchmark is dominated by segmentation tasks,
and its classification evaluation is comparatively limited. Our study instead
focuses on disease-detection problems involving subtle and heterogeneous
structural signals, including Alzheimer's disease, mild cognitive impairment,
autism spectrum disorder, and tumor grading.

The distinction between frozen and fine-tuned evaluation is also important.
In-domain SSL representations may show clear advantages under linear probing
that diminish after full fine-tuning \cite{rajamohan2026ssl}. Conversely,
semantically enriched pretraining objectives can produce frozen
representations that remain highly effective with limited downstream
supervision \cite{gao2026mass}. These observations indicate that the gap
between frozen and fine-tuned performance depends on the pretraining
objective, distribution shift, task difficulty, and alignment between the
learned representation and the downstream discriminative signal. 
Accordingly, we evaluate both settings to distinguish intrinsic representation utility from task-specific adaptation.

Transformer-based models have additionally been explored for longitudinal MRI
prognosis, including prediction of future knee replacement from 3D MRI
\cite{deniz2025mrtransformer}. Supervised studies on ADNI and external NACC
cohorts further demonstrate both the feasibility of structural MRI disease
classification and the persistent difficulty of distinguishing mild cognitive
impairment from normal aging \cite{liu2022razavian}. These findings motivate
self-supervised representations that can transfer across cohorts and disease
settings without relying exclusively on task-specific annotation.

\subsection{Positioning of the Present Study}
Existing medical SSL work has demonstrated substantial gains through both
large-scale pretraining and increasingly sophisticated architectural and
objective design. Less well characterized is whether the effects of these
choices remain consistent across downstream clinical tasks whose discriminative
signals differ in spatial extent, frequency content, and pathological
heterogeneity.

Concurrent work by Huang et al.\ introduces Neuro-JEPA, a large-scale
multimodal neuroimaging model combining JEPA pretraining with
mixture-of-experts routing, foreground-aware prediction, and multiscale
masking \cite{huang}. Neuro-JEPA demonstrates the potential of architectural
and objective innovations for improving transfer at scale. Our study addresses
a complementary question: under a controlled structural-MRI setting, how do
reconstruction-based and latent-predictive pretraining, together with
paradigm-specific auxiliary objectives, differ in their transfer behavior across
clinical tasks with distinct discriminative characteristics?

We therefore compare MAE and JEPA in their defining formulations, introduce
spectral-domain supervision for MAE, and evaluate variance--covariance
regularization within JEPA under both frozen and fine-tuned transfer. In
addition to downstream performance, we use CKA and Shannon effective rank to
examine how pretraining objective, encoder depth, and downstream task jointly
shape representational reorganization during adaptation. This design allows us
to study not only whether an objective improves transfer, but also how its
representational effects vary across clinical problems.
 
\section{Datasets}

We use nine research-accessible neuroimaging datasets for self-supervised
pretraining and downstream evaluation: the Alzheimer's Disease Neuroimaging
Initiative (ADNI) \cite{adni_dataset}, the National Alzheimer's Coordinating Center / Standardized Centralized Alzheimer's and Related Dementias Neuroimaging (NACC/SCAN) dataset \cite{scan_dataset}, the
Medical Out-of-Distribution Analysis Challenge (MOOD)
\cite{mood_dataset}, the Brain Tumor Segmentation Challenge 2024
(BraTS-2024) \cite{BraTS24_dataset}, the Parkinson's Progression Markers
Initiative (PPMI) \cite{ppmi_dataset}, the Autism Brain Imaging Data
Exchange (ABIDE) \cite{abide_dataset}, the Open Access Series of Imaging
Studies 3 (OASIS-3) \cite{oasis3_dataset}, the IXI dataset \cite{ixi_dataset}, and the University of California,
San Francisco Preoperative Diffuse Glioma MRI dataset (UCSF-PDGM)
\cite{ucsf_dataset}.

\subsection{Pretraining Cohort}

The self-supervised pretraining cohort comprises 58,781 MRI volumes from
8,769 subjects across ADNI, NACC/SCAN, PPMI, OASIS-3, IXI, MOOD, and
BraTS-2024. Each volume is processed as an individual single-contrast input.
The available contrasts include T1-weighted, T2-weighted, FLAIR, and T2*
MRI, depending on the source dataset. The cohort therefore contains
substantial variation in acquisition protocol, contrast weighting, pathology,
and subject population, including healthy controls and subjects with
neurodegenerative disease, mild cognitive impairment, Parkinson's disease,
and diffuse glioma. Supplementary Table~S1 summarizes the complete composition of the pretraining cohort.

\subsection{Downstream Evaluation Cohorts}

Downstream evaluation is performed on disease-detection and tumor-grading
tasks using ADNI, NACC/SCAN, UCSF-PDGM, and ABIDE. ADNI and NACC/SCAN are
used for normal control (NC) versus Alzheimer's disease (AD) and NC versus
mild cognitive impairment (MCI) classification. UCSF-PDGM is used for binary
and multiclass tumor grading, and ABIDE is used for autism spectrum disorder
classification. UCSF-PDGM and ABIDE are excluded from self-supervised
pretraining, thereby providing fully external downstream evaluation cohorts.
Supplementary Table~S1 summarizes the datasets, MRI contrasts, subject
counts, and volume counts used at each stage of the study.

\subsection{Subject-Level Separation and Data Splits}

For datasets contributing to both self-supervised pretraining and downstream
evaluation, all partitions are created at the subject level to prevent leakage
between pretraining and downstream cohorts.

For ADNI, subjects with more than 24 longitudinal MRI volumes were excluded
to reduce extreme subject-level imbalance. A total of 496 subjects were then
reserved exclusively for downstream evaluation, while the remaining subjects
were used for self-supervised pretraining. By volume count, this allocation corresponds to approximately 85\% of ADNI MRI volumes used for pretraining and 15\% reserved for downstream evaluation. The downstream cohort was evaluated using
stratified five-fold cross-validation, with all scans from a given subject
assigned to the same fold.

For NACC/SCAN, subjects whose cognitive-status labels did not correspond to
the downstream NC, MCI, or AD classification tasks were assigned exclusively
to pretraining. Among the remaining eligible subjects, 75\% were allocated to
pretraining and 25\% were reserved for downstream evaluation. The downstream
subset was then divided at the subject level into 70\% training, 15\%
validation, and 15\% test partitions.

\subsection{Reproducibility}

Standardized subject-level partitions are not available for many medical
imaging datasets, which makes direct comparison across studies difficult. We
therefore publicly release all subject identifiers used for self-supervised
pretraining and downstream train, validation, and test partitions at
\url{https://github.com/ituvisionlab/mjepa}. The released files contain only
anonymized identifiers supplied by the respective dataset custodians; no
imaging data or protected health information are redistributed.


\section{Methods}

Self-supervised learning methods for computer vision broadly fall into three families: invariance-based approaches, which learn representations by enforcing consistency across different views of the same input; reconstruction-based approaches, which learn by reconstructing masked or corrupted inputs; and predictive representation learning approaches, which predict latent representations of masked regions without reconstructing the input signal. In this work, we compare two representative methods from the latter two families for learning 3D MRI representations: Masked Autoencoders (MAE) and Joint Embedding Predictive Architectures (JEPA). MAE operates in the voxel reconstruction space while JEPA operates in the latent prediction space. Both methods enable pretraining without explicit labels. 

To support a controlled paradigm-level comparison, we implement MAE and JEPA according to their defining formulations, 
retaining the defining architectural components of each paradigm (the reconstruction decoder for MAE and the latent predictor for JEPA) while avoiding additional neuroimaging-specific architectural enhancements.
We first compare these base formulations and subsequently evaluate paradigm-specific auxiliary-loss variants, allowing us to examine how reconstruction-based and latent-predictive pretraining mechanisms influence downstream transfer.

\subsection{Masked Autoencoders (MAE)}

Masked Autoencoders (MAE) \cite{he2022masked} are reconstruction-based self-supervised learning models that learn representations by reconstructing masked regions of the input from sparse visible context. Unlike contrastive learning methods such as SimCLR \cite{chen2020simclr}, which align representations of different augmented views, MAE directly reconstructs missing image content. During pretraining, a high proportion (e.g., 75\%) of 3D input patches is masked, and the encoder--decoder architecture is trained to reconstruct the complete image. The encoder processes only the visible patches, while a lightweight decoder reconstructs the full volume using both encoded visible tokens and learned mask tokens. By avoiding dense encoding of the entire input, MAE achieves both computational efficiency and strong scalability.

Classical MAE divides an image into non-overlapping patches, samples a subset of patches as visible context, and masks the remaining ones. In our study, both MAE and JEPA employ the same volumetric multi-block masking framework adapted from V-JEPA \cite{drozdov2024videorepresentationlearningjointembedding} to 3D MRI. In our implementation, each sampled block corresponds to a single $8\times8\times8$ volumetric patch, and the number of sampled blocks is set to 75\% of the total number of patch positions. The sampled patches form the prediction or reconstruction targets, while the remaining visible patches provide the encoder context.

\begin{figure}[pos=!t]
    \centering
    \includegraphics[width=0.6\linewidth]{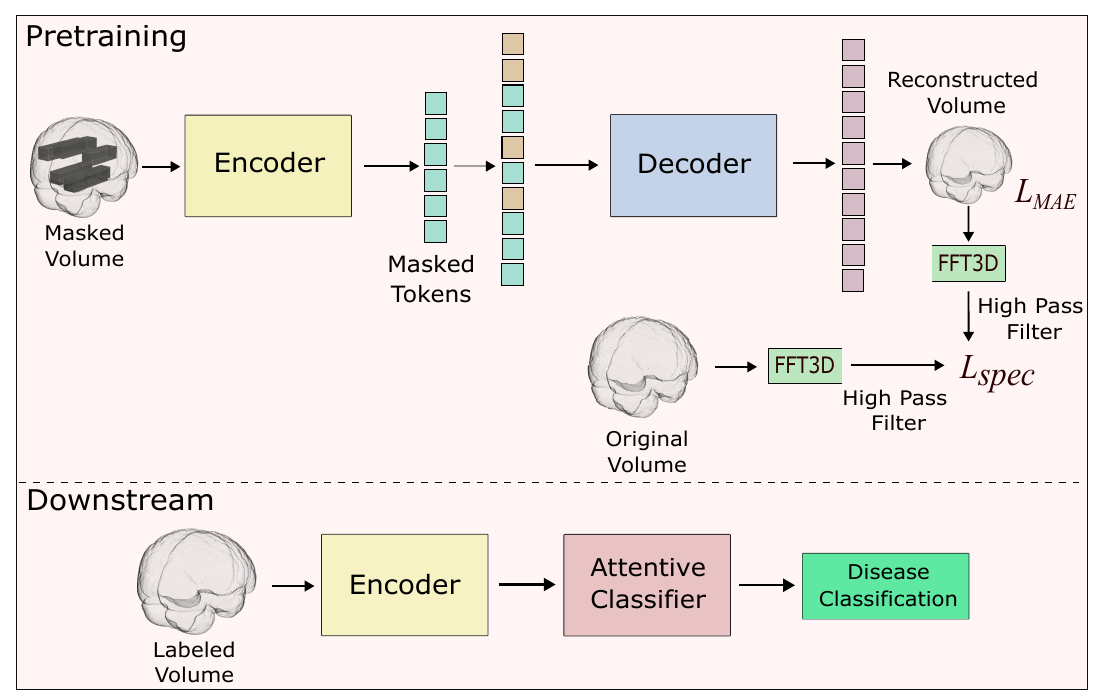}
    \caption{Overview of the proposed MAE pretraining and downstream evaluation framework. \textbf{Top:} During pretraining, the encoder processes visible tokens from masked 3D MRI patches, and the lightweight decoder reconstructs the full volume. The reconstruction loss $\mathcal{L}_{\text{MAE}}$
 is computed in the spatial domain, while the spectral loss $\mathcal{L}_{\text{spec}}$ is computed in the Fourier domain by applying a 3D FFT and a high-pass filter to both the reconstructed and original volumes, encouraging the encoder to capture fine-grained, high-frequency anatomical structure. \textbf{Bottom:} During downstream adaptation, the pretrained encoder is coupled with an Attentive Classifier and evaluated under both frozen and fine-tuned settings for disease classification. }
    \label{fig:mae_spectral_architecture}
\end{figure}

Figure \ref{fig:mae_spectral_architecture} illustrates the proposed MAE pretraining and downstream evaluation framework. During pretraining, the encoder processes only the visible tokens from the masked 3D MRI volume, while the lightweight decoder reconstructs the full volume. The spatial reconstruction loss $\mathcal{L}_{\text{MAE}}$
 penalizes voxel-wise discrepancies between the reconstructed and original volumes. Simultaneously, both the reconstructed and original volumes are transformed into the frequency domain via a 3D FFT, and the spectral loss $\mathcal{L}_{\text{spec}}$
is computed on their high-pass filtered log-magnitude spectra, encouraging the encoder to preserve fine-grained, high-frequency anatomical detail. During downstream adaptation, the pretrained encoder is kept either frozen or fine-tuned, and coupled with a lightweight Attentive Classifier for disease classification.

\subsection{Masked Autoencoding with Spectral Loss (SL-MAE)}\label{sec:MAE-SL}

Traditional masked autoencoders (MAE) optimize reconstruction loss in the spatial domain using voxel-wise mean squared error (MSE) over masked patches. This objective encourages fidelity in voxel-wise appearance by reconstructing the content of masked patches. Let $x_p \in \mathbb{R}^{D}$ denote the flattened vector of voxel values in patch $p$, where $D$ is the number of voxel values per patch, and let $\hat{x}_p$ denote the corresponding decoder prediction. The MAE reconstruction loss is computed as

\begin{equation}
\mathcal{L}_{\mathrm{MAE}}
=
\frac{1}{|\Omega_{\mathrm{mask}}|D}
\sum_{p\in\Omega_{\mathrm{mask}}}
\left\|
\hat{x}_p - x_p
\right\|_2^2 ,
\end{equation}

where $\Omega_{\mathrm{mask}}$ denotes the set of masked patch indices.

While effective, this objective does not explicitly constrain the frequency content of the reconstructed regions. Fine anatomical structures and sharp spatial transitions contribute to higher spatial-frequency components, motivating us to complement the voxel-domain reconstruction objective with frequency-domain supervision.

To address this limitation, we introduce an additional spectral loss term
computed in the frequency domain. The decoder predictions for the masked
patches are first combined with the original visible patches to form a full
reconstructed volume $\tilde{x}$. Before computing the Fourier transform,
both $\tilde{x}$ and the ground-truth volume $x$ are independently normalized
to zero mean and unit variance. Let $\mathcal{N}(\cdot)$ denote this
normalization and $\mathcal{F}(\cdot)$ the 3D discrete Fourier transform.

We apply a binary high-pass mask to the Fourier coefficients and compute the
mean squared error between the resulting log-magnitude spectra:

\begin{equation}
\begin{aligned}
\mathcal{L}_{\mathrm{spec}}
&=
\frac{1}{N_{\mathrm{freq}}}
\sum_{u,v,w}
M_{u,v,w}
\Bigg[
\log\left(
1+
\left|
\mathcal{F}(\mathcal{N}(\tilde{x}))_{u,v,w}
\right|
+\epsilon
\right)
\\
&\qquad\qquad
-
\log\left(
1+
\left|
\mathcal{F}(\mathcal{N}(x))_{u,v,w}
\right|
+\epsilon
\right)
\Bigg]^2 .
\end{aligned}
\end{equation}

where $N_{\mathrm{freq}}$ denotes the total number of frequency coefficients
in the 3D Fourier spectrum, $|\cdot|$ denotes magnitude, and $\epsilon$ is a
small constant for numerical stability. The binary high-pass mask
$M_{u,v,w}\in\{0,1\}$ is constructed by thresholding the normalized radial
frequency at $0.3$: coefficients with normalized radial frequency below
$0.3$ are suppressed, while those at or above the cutoff are retained.
The mask is constructed using the unshifted FFT frequency coordinates.

We use the log-magnitude spectrum to compress the large dynamic range of
spectral magnitudes and reduce the dominance of high-energy frequency
components. The additional high-pass filtering further suppresses
low-frequency components and places greater emphasis on higher-frequency
structure that may receive limited emphasis under voxel-space reconstruction
alone.

Because constructing the full reconstructed volume and computing its 3D
Fourier transform introduces additional computational cost, we evaluate
$\mathcal{L}_{\mathrm{spec}}$ periodically, once every $n$ training
iterations. Spectral gradients are therefore applied only on iterations in
which the spectral loss is recomputed; on the remaining iterations,
optimization proceeds using the MAE reconstruction objective alone.

Accordingly, on iterations in which the spectral loss is evaluated, the
training objective is
\begin{equation}
\mathcal{L}_{\mathrm{Total}}
=
\mathcal{L}_{\mathrm{MAE}}
+
\frac{\lambda_{\mathrm{FFT}}}{n}
\mathcal{L}_{\mathrm{spec}},
\end{equation}
where $n$ denotes the interval between successive spectral-loss evaluations
and $\lambda_{\mathrm{FFT}}$ controls the contribution of the spectral
objective. The effective spectral-loss coefficient is gradually increased
during its warm-up period. On iterations in which
$\mathcal{L}_{\mathrm{spec}}$ is not recomputed, the spectral term does not
contribute gradients.

We implement $\mathcal{F}(\cdot)$ using a 3D FFT operation. 

MRI acquisition noise and contrast variation can affect both voxel-space appearance and spectral characteristics. Combining spatial-domain reconstruction with spectral-domain supervision therefore provides complementary constraints: $\mathcal{L}_{\text{MAE}}$ promotes voxel-wise reconstruction fidelity, while $\mathcal{L}_{\text{spec}}$ emphasizes preservation of fine-scale frequency content that may otherwise receive limited weight under spatial-domain MSE alone.

\subsection{Joint Embedding Predictive Architecture (JEPA) and
Variance--Covariance Regularization (VCR)}

JEPA learns representations by predicting the latent representation of masked
target regions from visible context rather than reconstructing voxel values.
Our implementation follows a student--teacher formulation consisting of an
online encoder $E_\theta$, a predictor $P_\phi$, and a target encoder
$\bar{E}_\xi$. The target encoder is updated as an exponential moving average
(EMA) of the online encoder parameters and provides slowly evolving prediction
targets.

Given a 3D MRI volume $x$, we partition the volume into spatial patch tokens
and sample disjoint context and target regions using the volumetric masking
strategy described above. Let $\mathcal{C}$ and $\mathcal{T}$ denote the
context and target token sets, respectively. The online encoder processes only
the visible context tokens, while the target encoder processes the original
volume and provides latent representations corresponding to the masked target
regions. The predictor maps the encoded context, together with mask tokens
corresponding to the target locations, to predictions of these target
representations. Given context tokens $\mathcal{C}$ and masked
target tokens $\mathcal{T}$, the predictor estimates
\begin{equation}
\hat{\mathbf{H}}_{\mathcal{T}}
=
P_\phi\!\left(E_\theta(x_{\mathcal{C}}),z_{\mathcal{T}}\right),
\end{equation}
and is trained against stop-gradient target representations:
\begin{equation}
\mathcal{L}_{\mathrm{JEPA}}
=
\left\|
\hat{\mathbf{H}}_{\mathcal{T}}
-
\operatorname{sg}\!\left(\bar{E}_\xi(x)_{\mathcal{T}}\right)
\right\|_1 .
\end{equation}
The target encoder is updated by EMA,
\begin{equation}
\xi \leftarrow \mu\xi+(1-\mu)\theta,
\end{equation}
where $\mu$ is the momentum coefficient.

To improve representation quality and prevent feature collapse, we incorporate
\textbf{Variance--Covariance Regularization (VCR)}
\cite{drozdov2024videorepresentationlearningjointembedding}. VCR encourages each embedding dimension to maintain sufficient variance across
samples while discouraging redundancy between latent features.

VCR is applied jointly to the online-encoder context representations and the
stop-gradient target-encoder representations. For each context--target mask
pair, let
\begin{equation}
\mathbf H=
\left[
\mathbf H_{\mathcal C}^{\mathrm{online}},
\operatorname{sg}(\mathbf H_{\mathcal T}^{\mathrm{target}})
\right]
\in\mathbb{R}^{B\times T\times d},
\end{equation}
where $B$ is the batch size, $T$ the total number of context and target tokens,
and $d$ the embedding dimension. The context and
target representations are concatenated along the token dimension. Predictor
outputs are not included in the VCR term; gradients from VCR therefore
propagate through the online-encoder context representations but not through
the target-encoder representations.
The variance regularization term is
\begin{equation}
\mathcal{L}_{\mathrm{var}}(\mathbf{H})
=
\frac{1}{Td}
\sum_{t=1}^{T}
\sum_{k=1}^{d}
\max
\left(
0,
1-
\sqrt{
\mathrm{Var}(\mathbf{H}_{t,k})
+
\epsilon
}
\right),
\end{equation}
where $\epsilon=10^{-4}$ is used for numerical stability.

For the covariance term, rather than computing a separate covariance matrix
for each token position, we flatten the batch and token dimensions:
\begin{equation}
\mathbf{H}_{\mathrm{flat}}
=
\operatorname{reshape}(\mathbf{H})
\in
\mathbb{R}^{BT \times d}.
\end{equation}
Let
\begin{equation}
\bar{\mathbf{h}}_{\mathrm{flat}}
=
\frac{1}{BT}
\sum_{m=1}^{BT}
(\mathbf{H}_{\mathrm{flat}})_{m,:}
\end{equation}
denote the feature-wise mean. After centering the flattened representations,
the covariance matrix is computed as
\begin{equation}
\mathbf{C}
=
\frac{1}{BT-1}
\left(
\mathbf{H}_{\mathrm{flat}}
-
\bar{\mathbf{h}}_{\mathrm{flat}}
\right)^\top
\left(
\mathbf{H}_{\mathrm{flat}}
-
\bar{\mathbf{h}}_{\mathrm{flat}}
\right).
\end{equation}
The covariance regularization term penalizes the squared off-diagonal entries
of $\mathbf{C}$:
\begin{equation}
\mathcal{L}_{\mathrm{cov}}(\mathbf{H})
=
\frac{1}{d}
\sum_{i\neq j}
C_{ij}^{2}.
\end{equation}

The complete VCR term is
\begin{equation}
\mathcal{L}_{\mathrm{VCR}}(\mathbf{H})
=
\alpha_{\mathrm{var}}
\mathcal{L}_{\mathrm{var}}(\mathbf{H})
+
\beta_{\mathrm{cov}}
\mathcal{L}_{\mathrm{cov}}(\mathbf{H}),
\end{equation}
where $\alpha_{\mathrm{var}}$ and $\beta_{\mathrm{cov}}$ control the relative
contributions of the variance and covariance terms. When multiple
context--target mask pairs are used, the corresponding regularization terms
are averaged across mask pairs.

In practice, the variance and covariance components of VCR use different
aggregation strategies. The variance term is computed across the batch
independently for each token position and feature dimension. For the covariance
term, we flatten the batch and token dimensions and compute a single covariance
matrix over the resulting latent representations, rather than maintaining a
separate covariance matrix for each token position as in
\citet{drozdov2024videorepresentationlearningjointembedding}. This reduces
memory and computational overhead while still encouraging decorrelation across
feature dimensions.

The final JEPA--VCR training objective is
\begin{equation}
\mathcal{L}_{\mathrm{JEPA+VCR}}
=
\mathcal{L}_{\mathrm{JEPA}}
+
\lambda_{\mathrm{VCR}}
\mathcal{L}_{\mathrm{VCR}}(\mathbf{H}),
\end{equation}
where $\lambda_{\mathrm{VCR}}$ denotes the overall weight assigned to the VCR
regularization term during training.

\begin{figure}[pos=htbp]
    \centering
    \includegraphics[width=0.4\linewidth]{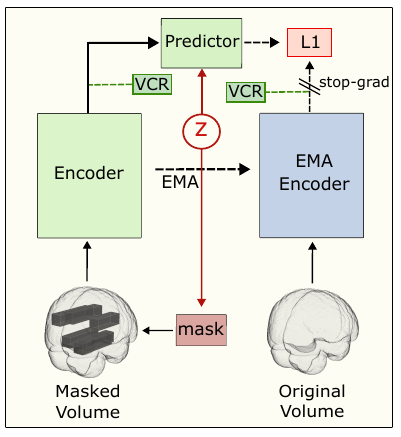}
    \caption{Overview of our JEPA framework. The online encoder processes
    visible tokens from masked 3D MRI patches, while the predictor learns to
    predict target-encoder representations corresponding to masked target
    regions. Variance--covariance regularization (VCR) is applied to the
    concatenated online context and stop-gradient target representations to
    encourage embedding diversity and reduce redundancy across feature
    dimensions.}
    \label{fig:mjepa_architecture}
\end{figure}

Figure~\ref{fig:mjepa_architecture} summarizes the resulting framework. The
online encoder processes visible context tokens, while the predictor estimates
the target-encoder representations corresponding to masked spatial regions.
VCR is applied to the concatenated online context and stop-gradient target
representations, providing an additional latent-space regularization signal
that encourages sufficient feature variance while reducing redundancy across
embedding dimensions.

\subsection{Downstream Transfer Evaluation}

The pretrained representations are evaluated across four downstream clinical classification problems, corresponding to seven dataset--task settings: AD vs.\ NC classification (ADNI and NACC/SCAN), MCI vs.\ NC classification (ADNI and NACC/SCAN), brain tumor grade prediction (binary and multiclass; UCSF), and autism classification (ABIDE). Performance is evaluated using accuracy, macro-averaged F1 score, and ROC-AUC. For multiclass tumor grading, ROC-AUC is computed using a one-vs-rest (OvR) strategy with macro averaging.

\section{Experimental Settings}

This section describes the data preprocessing pipeline, implementation details, self-supervised pretraining protocols, and downstream evaluation settings used throughout all experiments.

\paragraph{Data Preprocessing}

Each MRI volume is first cropped to the brain bounding box with a margin of six voxels. For anisotropic scans, the native in-plane acquisition orientation is preserved, whereas one of the axial, coronal, or sagittal orientations is randomly selected for isotropic volumes. The selected in-plane slices are center-cropped or resized to a fixed spatial resolution, while the through-plane dimension is interpolated to obtain standardized volumetric inputs. The resulting volumes are partitioned into non-overlapping 3D patches of size $8\times8\times8$. Intensity values are clipped to the 1st--99th percentile range and normalized to zero mean and unit variance on a per-volume basis. We do not perform atlas-based spatial normalization or resampling to a common physical voxel spacing.

\paragraph{Implementation Details}
Unless otherwise specified, all self-supervised models employ a ViT-Base 3D backbone and are pretrained for 300 epochs on four NVIDIA A100 GPUs using the AdamW optimizer with a cosine annealing learning rate schedule. Downstream fine-tuning is performed on two NVIDIA A100 GPUs. 

\paragraph{Masking Strategy}
All MAE- and JEPA-based models employ the same volumetric patch-masking
strategy adapted from V-JEPA. In our implementation, each sampled block
corresponds to a single $8\times8\times8$ volumetric patch, and the number
of sampled blocks is set to 75\% of the total number of patch positions.

\subsection{MAE Pretraining}

MAE models employ a 3D ViT-Base encoder with an input size of $144\times144\times144$ and non-overlapping patches of size
$8\times8\times8$ voxels. Following the original MAE framework, 75\% of the input patches are masked during pretraining. The learning rate is linearly increased from $1.0\times10^{-4}$ to $5.0\times10^{-4}$ during the first five warmup epochs and subsequently decayed to $5.0\times10^{-5}$ using a cosine schedule. 
SL-MAE follows the same architecture and optimization protocol as baseline
MAE. The spectral loss is applied every ten iterations beginning after epoch
five, with the frequency cutoff threshold set to 0.3. All remaining
pretraining hyperparameters are kept identical to the baseline MAE
configuration.

\subsection{JEPA Pretraining}
JEPA models employ the same 3D ViT-Base backbone with an input size of
$160\times160\times160$ voxels and non-overlapping patches of size
$8\times8\times8$ voxels. The models are trained to predict latent target
representations from visible context using the same 75\% masking ratio as MAE.
The learning rate is linearly increased from $1.0\times10^{-4}$ to
$2.0\times10^{-4}$ during the first 30 warmup epochs and subsequently decayed
to $1.0\times10^{-6}$ using a cosine schedule. VCR-JEPA follows the same
architecture and optimization protocol, with
$\alpha_{\mathrm{var}}=0.5$ and $\beta_{\mathrm{cov}}=0.1$.

\subsection{Downstream Training}

We evaluate pretrained representations across four downstream clinical
classification problems spanning Alzheimer's disease, mild cognitive
impairment, brain tumor grading, and autism spectrum disorder. For the ADNI
and NACC/SCAN datasets, both NC vs.\ AD and NC vs.\ MCI classification tasks
are considered. NC vs.\ MCI represents a substantially more challenging
discrimination problem because MCI is a heterogeneous clinical condition
associated with subtler structural alterations than established Alzheimer's
disease.

All downstream experiments are performed under both frozen and fine-tuned
evaluation protocols. For both settings, we use the Attentive Classifier
introduced in V-JEPA \cite{vjepa24} for MAE- and JEPA-based models. This
lightweight attention-based pooling classifier aggregates token-level
representations before classification, and a classifier depth of one is used
throughout all experiments.

\subsubsection{MAE Downstream Evaluation}

Unlike \citet{wald2024spark3d}, who utilize both the
pretrained encoder and decoder during downstream adaptation, we retain only the
pretrained encoder for downstream evaluation. Under fine-tuning, both the
encoder and Attentive Classifier are optimized. The learning rate is initialized
at $1.0\times10^{-5}$ and decayed to $1.0\times10^{-6}$ using a cosine
schedule over 400 epochs. We do not employ layer-wise learning rate decay, as
preliminary experiments did not demonstrate consistent improvements.

For frozen evaluation, the encoder parameters remain fixed and only the
Attentive Classifier is optimized. The learning rate is initialized at
$1.0\times10^{-4}$ and decayed to $1.0\times10^{-6}$ using a cosine
schedule.

\subsubsection{JEPA Downstream Evaluation}

For downstream evaluation, we retain the pretrained student encoder.
Fine-tuning follows the same optimization strategy used for MAE-based models.
For frozen evaluation, however, we found that JEPA-based models required
different classifier optimization settings on the external downstream cohorts.

In particular, frozen evaluation on cross-domain tasks, UCSF-PDGM tumor grading and ABIDE autism classification, required longer optimization schedules of up to 800 epochs and a smaller learning rate of $1.0\times10^{-5}$, compared with
$1.0\times10^{-4}$ used in the standard frozen setting. For these experiments,
the learning rate is warmed up from $1.0\times10^{-6}$ during the first 20
epochs and subsequently decayed using a cosine schedule.

\section{Experimental Evaluation}

We first compare the transfer performance of MAE and JEPA under frozen and fine-tuned evaluation protocols. Throughout this section, we primarily discuss ROC-AUC (referred to as AUC), as it provides a consistent basis for comparing downstream discriminative performance across tasks. Accuracy and macro-averaged F1 scores are reported alongside AUC in all tables.

\subsection{Masked Autoencoding and Spectral Pretraining} \label{subsec:MAE-SL}

Table~\ref{tab:mae_slmae_unfrozen} summarizes the fine-tuned evaluation
results comparing MAE and SL-MAE across all downstream settings. On ADNI
NC vs.\ AD classification, SL-MAE improves mean AUC from 0.866 to 0.945,
with lower variation across the five folds (standard deviation 0.024 vs.\
0.130). The benefit also appears on NACC/SCAN NC vs.\ AD, where AUC increases
from 0.761 to 0.841. On UCSF tumor grading, SL-MAE improves binary grading
AUC from 0.770 to 0.833 and multiclass grading AUC from 0.786 to 0.797.
A smaller improvement is observed on ABIDE (0.580 to 0.597). Across the
seven dataset--task settings, SL-MAE improves AUC in six, with a small
decrease on ADNI NC vs.\ MCI.

As reported in Table~\ref{tab:mae_slmae_unfrozen}, the improvement associated with spectral loss is substantially larger for NC vs.\ AD classification than for NC vs.\ MCI. On ADNI, the mean AUC changes marginally from 0.780 to 0.774 for NC vs.\ MCI, whereas NC vs.\ AD shows a substantial improvement from 0.866 to 0.945. A similar trend is observed on NACC/SCAN, where the gain is modest for NC vs.\ MCI (0.677 to 0.686) but more pronounced for NC vs.\ AD (0.761 to 0.841).

This pattern is consistent with the properties of frequency and texture features in structural MRI: while such features carry discriminative information for AD classification \cite{Zhang2022GFNet}, their discriminative power is consistently weaker for NC vs. MCI than for NC vs. AD \cite{Leandrou2020,Wang2022,Feng2018}. Although hippocampal texture and structural heterogeneity do evolve throughout the Alzheimer's continuum \cite{Wearn2023}, these alterations remain comparatively subtle and heterogeneous in cross-sectional NC vs.\ MCI discrimination settings. Prior work has further suggested that texture-related biomarkers may be more informative for predicting MCI-to-AD conversion than for separating stable MCI from normal controls \cite{Lee2020}. Our findings are consistent with spectral-domain auxiliary supervision preferentially emphasizing higher-frequency anatomical information, whose discriminative contribution may become more pronounced in established neurodegeneration.

Substantial AUC gains are also observed on UCSF tumor grading, where SL-MAE improves AUC from 0.770 to 0.833 on binary grading and
from 0.786 to 0.797 on multiclass classification. This pattern is consistent
with a favorable alignment between the type of representations encouraged by
spectral pretraining and the requirements of the tumor grading task. Brain
tumors introduce sharp intensity boundaries, necrotic cores, and heterogeneous
intratumoral texture, all of which correspond to high-frequency spatial
structure in MRI~\cite{Kinoshita2016, Shao2025}. In contrast, SL-MAE yields a
moderate AUC improvement on ABIDE (0.580 to 0.597). ASD is not characterized by focal lesions or sharp structural boundaries; reported structural differences are generally subtler and more spatially distributed than those associated with brain tumors~\cite{Chaddad2017}. Accordingly, we hypothesize that the
high-frequency sensitivity induced by spectral pretraining offers less benefit
for the ASD downstream task than for tumor grading.

\begin{figure}[pos=htbp]
    \centering
    \includegraphics[width=\linewidth]{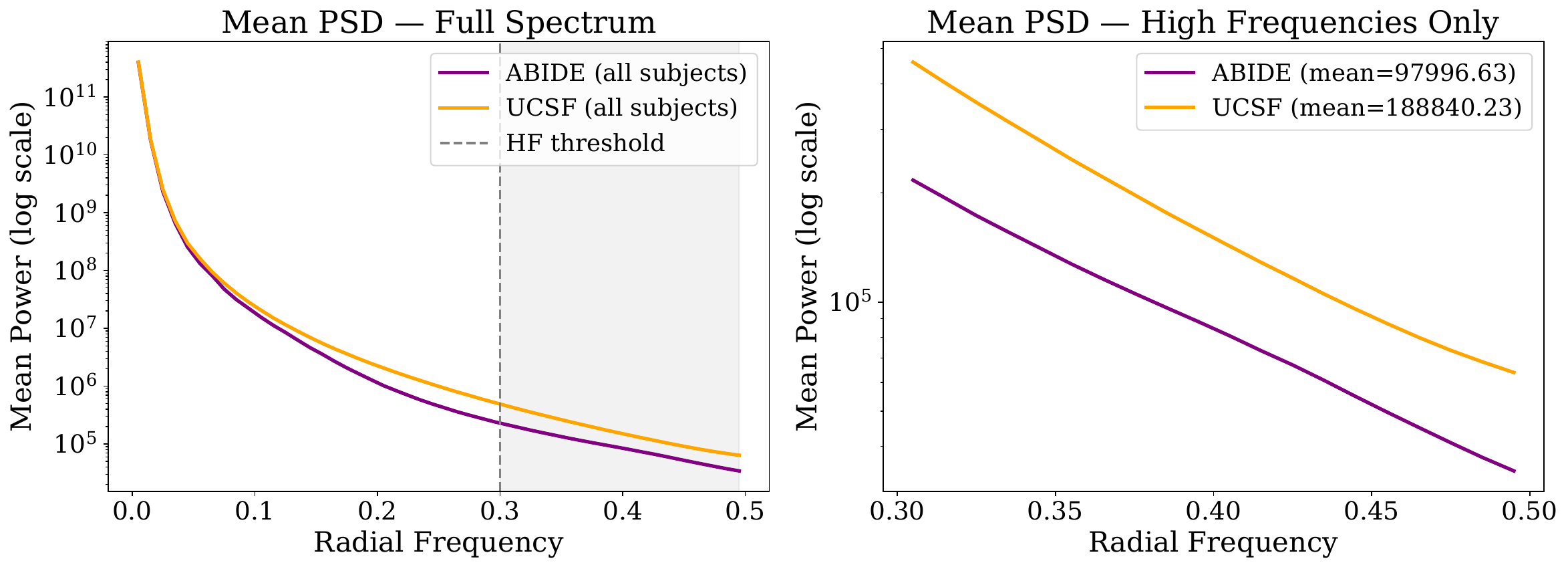}
    \caption{%
Mean high-frequency power spectral density (PSD) comparison between
UCSF tumor grading and ABIDE ASD datasets. Left: full radially averaged
PSD curves for both datasets collapsed across classes. Right: high-frequency
region only (radial frequency $\geq 0.3$), showing that UCSF volumes contain
approximately $1.93\times$ higher mean high-frequency spatial energy than
ABIDE volumes under the applied preprocessing (UCSF: $188840.23$, ABIDE:
$97996.63$).%
}
    \label{fig:psd_comparison}
\end{figure}

As a descriptive comparison of the spectral characteristics of the two
downstream datasets, we computed the mean high-frequency power spectral density
after applying the same preprocessing pipeline. UCSF volumes exhibit
approximately $1.93\times$ higher mean high-frequency spatial energy than
ABIDE volumes (Figure~\ref{fig:psd_comparison}). Because the datasets differ
in acquisition protocols, contrasts, populations, and pathology, this
comparison does not isolate disease-specific spectral content. Nevertheless,
in conjunction with the larger SL-MAE improvement observed on UCSF, the
spectral difference is consistent with the hypothesis that spectral pretraining
is more useful when high-frequency structural information contributes strongly
to the downstream discrimination problem.

The advantage of SL-MAE is substantially smaller on the UCSF multiclass task, where AUC increases only from 0.786 to 0.797 compared with the larger improvement observed for binary grading (0.770 to 0.833).  Multiclass classification requires distinguishing between multiple tumor subtypes that differ along axes beyond high-frequency texture alone, including global morphological patterns, tumor location, and tissue composition. As the task becomes more complex and less aligned with high-frequency spatial content specifically, the benefit of spectral pretraining diminishes. 

These findings suggest that spectral regularization is particularly beneficial for downstream tasks whose discriminative signal contains stronger high-frequency spatial structure, enabling the encoder to better capture fine-grained anatomical and textural variations during self-supervised pretraining.

Table~\ref{tab:mae_slmae_frozen} presents frozen evaluation results. SL-MAE maintains its advantage over plain MAE on ADNI NC 
vs.\ AD ($0.922 \pm 0.033$ vs.\ $0.914 \pm 0.025$) and on both UCSF 
tumor grading tasks (binary: 0.805 vs.\ 0.761; multiclass: 0.762 
vs.\ 0.755), showing that the effect of spectral pretraining is already evident when the pretrained encoder is kept fixed. 
MAE leads on ADNI NC vs.\ MCI ($0.757 \pm 0.047$ 
vs.\ $0.747 \pm 0.066$), NACC/SCAN NC vs.\ MCI (0.702 vs.\ 0.676), and 
ABIDE (0.616 vs.\ 0.610). The results suggest that spectral pretraining biases the learned representations toward higher-frequency spatial structure. This bias is beneficial when the downstream task itself contains discriminative high-frequency information, but may reduce the discriminability of frozen representations for tasks where such structure is weaker or less informative.
The strong frozen MAE result on ADNI NC vs.\ AD further illustrates that task-specific adaptation is not uniformly beneficial when the pretrained representation is already well aligned with the downstream discrimination problem.

Comparing Tables~\ref{tab:mae_slmae_unfrozen} and~\ref{tab:mae_slmae_frozen}, fine-tuning outperforms frozen evaluation with the SL-MAE encoder on ADNI NC vs.\ AD, NACC/SCAN NC vs.\ AD, and UCSF binary and multiclass tumor grading. This is expected, as fine-tuning allows the encoder to adapt its representations to the downstream task. However, task-specific adaptation is not uniformly beneficial for baseline MAE. Frozen MAE outperforms its fine-tuned counterpart on ADNI NC vs.\ AD (0.914 vs.\ 0.866), NACC/SCAN NC vs.\ MCI (0.702 vs.\ 0.677), and ABIDE (0.616 vs.\ 0.580), while remaining competitive on ADNI NC vs.\ MCI.

We hypothesize that this behavior reflects the interaction between downstream
task difficulty, dataset shift, and representation adaptation during fine-tuning.
For challenging tasks such as NC vs.\ MCI and ABIDE, pretrained representations
may already contain robust and generalizable anatomical information that cannot
be reliably improved through limited downstream supervision. In such settings,
fine-tuning may overfit acquisition-specific variation or dataset-specific
confounds, whereas frozen evaluation acts as an implicit regularizer that
preserves the broader anatomical structure learned during
pretraining \cite{Kumar2022,RepSim2025}.

Similarly, the NC vs.\ MCI task involves comparatively subtle and heterogeneous
disease-related changes, making representation preservation potentially more
advantageous than aggressive task-specific adaptation.

For SL-MAE, fine-tuning consistently outperforms frozen evaluation across all
tasks except ABIDE, where the frozen encoder yields a slightly higher AUC
(0.610 vs.\ 0.597). This exception may reflect the substantial demographic
and acquisition differences between ABIDE, whose cohort is predominantly
adolescent \cite{DiMartino2014}, and the heterogeneous MRI populations used
during pretraining.


\begin{table*}[!htbp]
\centering
\caption{%
  Fine-tuned evaluation comparing MAE and SL-MAE pretraining strategies
  across all downstream datasets and tasks.
  For ADNI, fold-wise and mean$\pm$std results are reported from 5-fold cross-validation.
  For NACC/SCAN, UCSF, and ABIDE, results are from a single train/test split.
  \textbf{Bold} indicates the better result per metric within each task.
  MAE\,=\,plain masked autoencoder pretraining;
  SL-MAE\,=\,MAE with spectral loss pretraining.%
}
\label{tab:mae_slmae_unfrozen}

\fontsize{10}{12}\selectfont

\setlength{\tabcolsep}{4pt}
\renewcommand{\arraystretch}{1.08}

\begin{tabular}{@{}llccc ccc@{}}
\toprule
& & \multicolumn{3}{c}{\textbf{MAE}}
& \multicolumn{3}{c}{\textbf{SL-MAE}} \\
\cmidrule(lr){3-5}\cmidrule(lr){6-8}

\textbf{Data} & \textbf{Task} &
  \textbf{Acc.\,(\%)} & \textbf{F1} & \textbf{AUC} &
  \textbf{Acc.\,(\%)} & \textbf{F1} & \textbf{AUC} \\
\midrule

\multicolumn{8}{@{}l}{\textit{ADNI\;---\;NC vs.\ AD}} \\[1pt]

& Fold\,0
  & 77.330 & 0.760 & 0.869
  & 79.849 & 0.784 & 0.899 \\

& Fold\,1
  & 62.987 & 0.609 & 0.616
  & 87.273 & 0.870 & 0.944 \\

& Fold\,2
  & 88.039 & 0.880 & 0.946
  & 90.999 & 0.910 & 0.965 \\

& Fold\,3
  & 84.299 & 0.839 & 0.940
  & 89.060 & 0.891 & 0.952 \\

& Fold\,4
  & 88.305 & 0.883 & 0.961
  & 89.976 & 0.900 & 0.964 \\

\cmidrule(lr){2-8}

& \textbf{Mean\,$\pm$\,Std}
  & $80.592 \pm 10.025$
  & $0.794 \pm 0.102$
  & $0.866 \pm 0.130$
  & $87.431 \pm 4.085$
  & $0.871 \pm 0.043$
  & $\mathbf{0.945 \pm 0.024}$ \\

\midrule

\multicolumn{8}{@{}l}{\textit{ADNI\;---\;NC vs.\ MCI}} \\[1pt]

& Fold\,0
  & 71.953 & 0.697 & 0.734
  & 71.166 & 0.700 & 0.712 \\

& Fold\,1
  & 75.530 & 0.727 & 0.789
  & 70.297 & 0.692 & 0.725 \\

& Fold\,2
  & 80.576 & 0.804 & 0.870
  & 81.727 & 0.815 & 0.843 \\

& Fold\,3
  & 80.148 & 0.795 & 0.846
  & 84.148 & 0.831 & 0.862 \\

& Fold\,4
  & 63.791 & 0.621 & 0.659
  & 72.702 & 0.710 & 0.731 \\

\cmidrule(lr){2-8}

& \textbf{Mean\,$\pm$\,Std}
  & $74.400 \pm 6.697$
  & $0.729 \pm 0.068$
  & $\mathbf{0.780 \pm 0.079}$
  & $75.208 \pm 5.688$
  & $0.750 \pm 0.060$
  & $0.774 \pm 0.063$ \\

\midrule

\multicolumn{8}{@{}l}{\textit{NACC/SCAN}} \\[1pt]

& NC vs.\ AD
  & 85.56 & 0.703 & 0.761
  & 87.97 & 0.738 & \textbf{0.841} \\

& NC vs.\ MCI
  & 72.17 & 0.615 & 0.677
  & 73.76 & 0.635 & \textbf{0.686} \\

\midrule

\multicolumn{8}{@{}l}{\textit{UCSF}} \\[1pt]

& 2 vs.\ 3--4
  & 67.11 & 0.669 & 0.770
  & 80.75 & 0.807 & \textbf{0.833} \\

& Multiclass
  & 59.05 & 0.548 & 0.786
  & 61.07 & 0.600 & \textbf{0.797} \\

\midrule

\multicolumn{8}{@{}l}{\textit{ABIDE}} \\[1pt]

& ASD vs.\ NC
  & 55.09 & 0.585 & 0.580
  & 55.56 & 0.550 & \textbf{0.597} \\

\bottomrule
\end{tabular}

\end{table*}


\begin{table*}[!htbp]
\centering
\caption{%
  Frozen evaluation comparing MAE and SL-MAE pretraining strategies
  across all downstream datasets and tasks.
  For ADNI, fold-wise and mean$\pm$std results are reported from 5-fold cross-validation.
  For NACC/SCAN, UCSF, and ABIDE, results are from a single train/test split.
  \textbf{Bold} indicates the better AUC per task.
  MAE\,=\,plain masked autoencoder pretraining;
  SL-MAE\,=\,MAE with spectral loss pretraining.%
}
\label{tab:mae_slmae_frozen}

\fontsize{10}{12}\selectfont

\setlength{\tabcolsep}{4pt}
\renewcommand{\arraystretch}{1.08}

\begin{tabular}{@{}llccc ccc@{}}
\toprule
& & \multicolumn{3}{c}{\textbf{MAE}}
& \multicolumn{3}{c}{\textbf{SL-MAE}} \\
\cmidrule(lr){3-5}\cmidrule(lr){6-8}

\textbf{Data} & \textbf{Task} &
  \textbf{Acc.\,(\%)} & \textbf{F1} & \textbf{AUC} &
  \textbf{Acc.\,(\%)} & \textbf{F1} & \textbf{AUC} \\
\midrule

\multicolumn{8}{@{}l}{\textit{ADNI\;---\;NC vs.\ AD}} \\[1pt]

& Fold\,0
  & 79.850 & 0.793 & 0.876
  & 80.230 & 0.797 & 0.881 \\

& Fold\,1
  & 87.660 & 0.874 & 0.929
  & 89.740 & 0.895 & 0.953 \\

& Fold\,2
  & 89.150 & 0.891 & 0.932
  & 87.300 & 0.873 & 0.926 \\

& Fold\,3
  & 84.300 & 0.843 & 0.897
  & 81.850 & 0.818 & 0.892 \\

& Fold\,4
  & 86.750 & 0.866 & 0.936
  & 89.860 & 0.899 & 0.956 \\

\cmidrule(lr){2-8}

& \textbf{Mean\,$\pm$\,Std}
  & $85.542 \pm 3.484$
  & $0.854 \pm 0.036$
  & $0.914 \pm 0.025$
  & $85.796 \pm 4.354$
  & $0.856 \pm 0.044$
  & $\mathbf{0.922 \pm 0.033}$ \\

\midrule

\multicolumn{8}{@{}l}{\textit{ADNI\;---\;NC vs.\ MCI}} \\[1pt]

& Fold\,0
  & 63.430 & 0.632 & 0.680
  & 61.884 & 0.584 & 0.642 \\

& Fold\,1
  & 70.440 & 0.698 & 0.765
  & 76.667 & 0.747 & 0.758 \\

& Fold\,2
  & 74.530 & 0.745 & 0.815
  & 78.125 & 0.765 & 0.809 \\

& Fold\,3
  & 79.850 & 0.789 & 0.789
  & 80.959 & 0.789 & 0.813 \\

& Fold\,4
  & 69.170 & 0.681 & 0.737
  & 66.481 & 0.630 & 0.713 \\

\cmidrule(lr){2-8}

& \textbf{Mean\,$\pm$\,Std}
  & $71.484 \pm 5.972$
  & $0.709 \pm 0.057$
  & $\mathbf{0.757 \pm 0.047}$
  & $72.823 \pm 7.677$
  & $0.703 \pm 0.082$
  & $0.747 \pm 0.066$ \\

\midrule

\multicolumn{8}{@{}l}{\textit{NACC/SCAN}} \\[1pt]

& NC vs.\ AD
  & 83.81 & 0.617 & 0.725
  & 82.09 & 0.651 & \textbf{0.742} \\

& NC vs.\ MCI
  & 81.66 & 0.536 & \textbf{0.702}
  & 74.86 & 0.569 & 0.676 \\

\midrule

\multicolumn{8}{@{}l}{\textit{UCSF}} \\[1pt]

& 2 vs.\ 3--4
  & 65.71 & 0.624 & 0.761
  & 73.26 & 0.731 & \textbf{0.805} \\

& Multiclass
  & 59.56 & 0.574 & 0.755
  & 58.64 & 0.562 & \textbf{0.762} \\

\midrule

\multicolumn{8}{@{}l}{\textit{ABIDE}} \\[1pt]

& ASD vs.\ NC
  & 60.71 & 0.600 & \textbf{0.616}
  & 57.41 & 0.559 & 0.610 \\

\bottomrule
\end{tabular}

\end{table*}

\subsection{Latent Predictive Pretraining}

Table~\ref{tab:jepa_vcrjepa_unfrozen} presents the fine-tuned evaluation results. VCR-JEPA consistently outperforms plain JEPA on both ADNI classification tasks, achieving mean AUCs of 0.835 vs.\ 0.759 for NC vs.\ AD and 0.736 vs.\ 0.724 for NC vs.\ MCI. For tumor grading on UCSF, VCR-JEPA improves AUC from 0.655 to 0.686 on the binary 2 vs.\ 3--4 task; while the multiclass task yields nearly identical AUCs (JEPA: 0.675, VCR-JEPA: 0.672), indicating little difference between the two objectives after fine-tuning in this setting. On NACC/SCAN NC vs.\ AD, VCR-JEPA again leads (0.678 vs.\ 0.663), whereas on NACC/SCAN NC vs.\ MCI, plain JEPA achieves a higher AUC (0.660 vs.\ 0.554). For ASD detection on ABIDE, VCR-JEPA provides a modest improvement (0.583 vs.\ 0.569).

VCR encourages sufficient variance across feature dimensions while reducing cross-dimensional covariance, thereby discouraging redundant use of the latent space, \cite{drozdov2024videorepresentationlearningjointembedding,Bardes2022,Zbontar2021}.
Our results suggest that the benefit of VCR depends strongly on the structure and heterogeneity of the downstream discriminative signal.

In Alzheimer's disease, structural MRI abnormalities involve multiple
anatomical systems, including medial temporal atrophy, cortical thinning, and
ventricular enlargement\cite{Frisoni2010,Planche2022}. Similarly, glioma
grading depends on spatially heterogeneous imaging characteristics associated
with necrosis, cellular heterogeneity, and infiltrative growth across distinct
tumor regions\cite{Rose2009,Gillies2016}. The comparatively stronger
VCR-JEPA performance on these tasks is therefore consistent with the
hypothesis that reducing redundancy across latent dimensions is useful when
downstream discrimination draws on multiple complementary sources of
structural variation.

In contrast, MCI-related structural changes are comparatively subtle and
heterogeneous, with prominent alterations often reported in medial temporal
structures such as the hippocampus and entorhinal cortex
\cite{Planche2022,Pennanen2004}. ASD-related structural MRI findings are
also generally weak and heterogeneous across individuals and studies, with
between-subject variability often exceeding case--control differences
\cite{vanRooij2018,Haar2016}. In such settings, the additional
variance--covariance constraint may provide less consistent benefit because
the downstream signal is weakly expressed or less reliably separable from background anatomical variation.


\begin{table*}[!htbp]
\centering
\caption{%
  Fine-tuned evaluation comparing JEPA and VCR-JEPA pretraining strategies
  across all downstream datasets and tasks.
  For ADNI, fold-wise and mean$\pm$std results are reported from 5-fold cross-validation.
  For NACC/SCAN, UCSF, and ABIDE, results are from a single train/test split.
  \textbf{Bold} indicates the better AUC per task.
  JEPA\,=\,plain JEPA pretraining;
  VCR-JEPA\,=\,JEPA with variance-covariance regularization pretraining.%
}
\label{tab:jepa_vcrjepa_unfrozen}

\fontsize{10}{12}\selectfont

\setlength{\tabcolsep}{4pt}
\renewcommand{\arraystretch}{1.08}

\begin{tabular}{@{}llccc ccc@{}}
\toprule
& & \multicolumn{3}{c}{\textbf{JEPA}}
& \multicolumn{3}{c}{\textbf{VCR-JEPA}} \\
\cmidrule(lr){3-5}\cmidrule(lr){6-8}

\textbf{Data} & \textbf{Task} &
  \textbf{Acc.\,(\%)} & \textbf{F1} & \textbf{AUC} &
  \textbf{Acc.\,(\%)} & \textbf{F1} & \textbf{AUC} \\
\midrule

\multicolumn{8}{@{}l}{\textit{ADNI\;---\;NC vs.\ AD}} \\[1pt]

& Fold\,0
  & 64.330 & 0.615 & 0.682
  & 71.350 & 0.706 & 0.787 \\

& Fold\,1
  & 68.750 & 0.670 & 0.736
  & 71.940 & 0.694 & 0.801 \\

& Fold\,2
  & 80.350 & 0.796 & 0.873
  & 80.340 & 0.792 & 0.869 \\

& Fold\,3
  & 68.960 & 0.656 & 0.781
  & 77.480 & 0.748 & 0.842 \\

& Fold\,4
  & 67.850 & 0.664 & 0.725
  & 79.009 & 0.793 & 0.876 \\

\cmidrule(lr){2-8}

& \textbf{Mean\,$\pm$\,Std}
  & $70.048 \pm 5.413$
  & $0.680 \pm 0.061$
  & $0.759 \pm 0.065$
  & $76.024 \pm 3.693$
  & $0.747 \pm 0.042$
  & $\mathbf{0.835 \pm 0.036}$ \\

\midrule

\multicolumn{8}{@{}l}{\textit{ADNI\;---\;NC vs.\ MCI}} \\[1pt]

& Fold\,0
  & 62.350 & 0.591 & 0.630
  & 68.210 & 0.646 & 0.688 \\

& Fold\,1
  & 72.920 & 0.707 & 0.708
  & 68.290 & 0.645 & 0.703 \\

& Fold\,2
  & 78.940 & 0.776 & 0.807
  & 78.690 & 0.775 & 0.820 \\

& Fold\,3
  & 75.820 & 0.731 & 0.795
  & 78.630 & 0.752 & 0.815 \\

& Fold\,4
  & 66.620 & 0.637 & 0.679
  & 65.600 & 0.617 & 0.653 \\

\cmidrule(lr){2-8}

& \textbf{Mean\,$\pm$\,Std}
  & $71.330 \pm 6.200$
  & $0.688 \pm 0.069$
  & $0.724 \pm 0.068$
  & $71.880 \pm 5.810$
  & $0.687 \pm 0.063$
  & $\mathbf{0.736 \pm 0.069}$ \\

\midrule

\multicolumn{8}{@{}l}{\textit{NACC/SCAN}} \\[1pt]

& NC vs.\ AD
  & 79.850 & 0.563 & 0.663
  & 82.800 & 0.596 & \textbf{0.678} \\

& NC vs.\ MCI
  & 75.740 & 0.546 & \textbf{0.660}
  & 73.470 & 0.539 & 0.554 \\

\midrule

\multicolumn{8}{@{}l}{\textit{UCSF}} \\[1pt]

& 2 vs.\ 3--4
  & 60.240 & 0.593 & 0.655
  & 65.710 & 0.641 & \textbf{0.686} \\

& Multiclass
  & 50.840 & 0.484 & \textbf{0.675}
  & 51.820 & 0.512 & 0.672 \\

\midrule

\multicolumn{8}{@{}l}{\textit{ABIDE}} \\[1pt]

& ASD vs.\ NC
  & 52.680 & 0.518 & 0.569
  & 55.360 & 0.544 & \textbf{0.583} \\

\bottomrule
\end{tabular}

\end{table*}


\begin{table*}[!htbp]
\centering
\caption{%
  Frozen (attentive probing) evaluation comparing JEPA and VCR-JEPA pretraining strategies
  across all downstream datasets and tasks.
  For ADNI, fold-wise and mean$\pm$std results are reported from 5-fold cross-validation.
  For NACC/SCAN, UCSF, and ABIDE, results are from a single train/test split.
  \textbf{Bold} indicates the better AUC per task.
  JEPA\,=\,plain JEPA pretraining;
  VCR-JEPA\,=\,JEPA with variance-covariance regularization pretraining.%
}
\label{tab:jepa_vcrjepa_frozen}

\fontsize{10}{12}\selectfont

\setlength{\tabcolsep}{4pt}
\renewcommand{\arraystretch}{1.08}

\begin{tabular}{@{}llccc ccc@{}}
\toprule
& & \multicolumn{3}{c}{\textbf{JEPA}}
& \multicolumn{3}{c}{\textbf{VCR-JEPA}} \\
\cmidrule(lr){3-5}\cmidrule(lr){6-8}

\textbf{Data} & \textbf{Task} &
\textbf{Acc.\,(\%)} & \textbf{F1} & \textbf{AUC} &
\textbf{Acc.\,(\%)} & \textbf{F1} & \textbf{AUC} \\
\midrule

\multicolumn{8}{@{}l}{\textit{ADNI\;---\;NC vs.\ AD}} \\[1pt]

& Fold\,0
  & 56.730 & 0.547 & 0.596
  & 61.800 & 0.596 & 0.655 \\

& Fold\,1
  & 61.220 & 0.589 & 0.647
  & 60.590 & 0.579 & 0.655 \\

& Fold\,2
  & 65.970 & 0.635 & 0.733
  & 76.000 & 0.747 & 0.807 \\

& Fold\,3
  & 63.760 & 0.606 & 0.698
  & 67.770 & 0.647 & 0.738 \\

& Fold\,4
  & 65.450 & 0.628 & 0.728
  & 63.640 & 0.621 & 0.692 \\

\cmidrule(lr){2-8}

& \textbf{Mean\,$\pm$\,Std}
  & $62.626 \pm 3.540$
  & $0.601 \pm 0.033$
  & $0.680 \pm 0.057$
  & $65.960 \pm 5.760$
  & $0.638 \pm 0.059$
  & $\mathbf{0.709 \pm 0.058}$ \\

\midrule

\multicolumn{8}{@{}l}{\textit{ADNI\;---\;NC vs.\ MCI}} \\[1pt]

& Fold\,0
  & 63.330 & 0.604 & 0.649
  & 64.890 & 0.615 & 0.656 \\

& Fold\,1
  & 69.820 & 0.681 & 0.736
  & 62.040 & 0.380 & 0.540 \\

& Fold\,2
  & 73.540 & 0.719 & 0.768
  & 76.560 & 0.751 & 0.823 \\

& Fold\,3
  & 72.140 & 0.688 & 0.774
  & 77.040 & 0.747 & 0.804 \\

& Fold\,4
  & 64.350 & 0.621 & 0.652
  & 64.350 & 0.611 & 0.649 \\

\cmidrule(lr){2-8}

& \textbf{Mean\,$\pm$\,Std}
  & $68.636 \pm 4.230$
  & $0.663 \pm 0.044$
  & $\mathbf{0.716 \pm 0.057}$
  & $68.976 \pm 6.470$
  & $0.621 \pm 0.138$
  & $0.694 \pm 0.106$ \\

\midrule

\multicolumn{8}{@{}l}{\textit{NACC/SCAN}} \\[1pt]

& NC vs.\ AD
  & 82.570 & 0.543 & 0.672
  & 83.075 & 0.653 & \textbf{0.719} \\

& NC vs.\ MCI
  & 72.870 & 0.547 & \textbf{0.612}
  & 70.230 & 0.540 & 0.569 \\

\midrule

\multicolumn{8}{@{}l}{\textit{UCSF}} \\[1pt]

& 2 vs.\ 3--4
  & 56.940 & 0.555 & 0.590
  & 61.720 & 0.603 & \textbf{0.673} \\

& Multiclass
  & 43.770 & 0.415 & 0.599
  & 45.570 & 0.436 & \textbf{0.640} \\

\midrule

\multicolumn{8}{@{}l}{\textit{ABIDE}} \\[1pt]

& ASD vs.\ NC
  & 58.040 & 0.570 & 0.531
  & 56.250 & 0.553 & \textbf{0.572} \\

\bottomrule
\end{tabular}

\end{table*}

Table~\ref{tab:jepa_vcrjepa_frozen} presents the frozen 
results, evaluating the quality of pretrained representations without 
any encoder fine-tuning. Under this evaluation protocol, 
VCR-JEPA again outperforms JEPA on NC vs.\ AD classification, 
achieving a mean AUC of $0.709 \pm 0.058$ compared to $0.680 \pm 0.057$, 
indicating that VCR improves the discriminative utility of the pretrained representations even when the encoder is kept fixed.
For NC vs.\ MCI, this pattern reverses. JEPA achieves a higher mean AUC ($0.716 \pm 0.057$ vs.\ $0.694 \pm 0.106$) with notably lower variance. 
VCR-JEPA also exhibits greater fold-to-fold variability (standard deviation 0.106 vs. 0.057), indicating that its benefit is less consistent across ADNI NC-vs-MCI splits.

In contrast, VCR-JEPA shows clear advantages on the UCSF tumor grading
tasks under frozen evaluation, improving both binary (AUC 0.673 vs.\ 0.590)
and multiclass (AUC 0.640 vs.\ 0.599) classification. These results provide
further support for the hypothesis that decorrelated latent representations
are beneficial when downstream discrimination depends on multiple
heterogeneous structural characteristics, as in glioma grading, even without
encoder fine-tuning. On ABIDE, VCR-JEPA achieves a marginal improvement
(AUC 0.572 vs.\ 0.531).

Overall, the frozen results reinforce the task-dependent effect of VCR:
VCR-JEPA outperforms baseline JEPA on five of seven dataset--task settings,
with its clearest gains occurring for AD classification and tumor grading.
The weaker or negative effects on NC vs.\ MCI indicate that broader,
less-redundant latent representations are not uniformly advantageous across
downstream problems.


\subsection{Comparison with Related Work}

Direct comparison across self-supervised medical imaging studies remains challenging due to differences in train/test splits, preprocessing pipelines, and evaluation protocols. To improve reproducibility and facilitate future
benchmarking, we publicly release all subject-level partitions used for
self-supervised pretraining and downstream evaluation in our GitHub repository.

We compare our framework with BrainFound, a recent self-supervised foundation
model for structural brain MRI based on DINOv2-style self-distillation
\cite{Mazher2025,Oquab2023}. Although the two studies employ different
experimental protocols, BrainFound provides one of the closest available
reference points because it similarly targets general-purpose brain MRI
representation learning.

Table~\ref{tab:comparison_brainfound} summarizes performance on overlapping
downstream tasks. BrainFound employs a ViT-Large backbone initialized from
natural-image pretraining and uses multiple MRI contrasts for several downstream
tasks, whereas our models use a ViT-Base backbone pretrained from scratch and
process individual MRI contrasts independently. ADNI provides the most directly
comparable setting because both studies evaluate T1-weighted MRI. On ADNI
NC vs.\ AD classification, SL-MAE achieves a mean AUC of
$0.945 \pm 0.024$, compared with $0.810$ reported by BrainFound. On ADNI
NC vs.\ MCI, MAE achieves 0.780 compared with 0.766. On UCSF tumor grading,
SL-MAE achieves an AUC of 0.833 compared with BrainFound's 0.780, despite
using individual single-contrast inputs rather than the multi-contrast input
available to BrainFound for this task. On NACC/SCAN NC vs.\ AD, BrainFound
reports a higher AUC (0.883 vs.\ 0.841), illustrating that the relative
ordering is not uniform across datasets.

Overall, our models achieve competitive reported performance across the
overlapping tasks, with higher AUCs in three of the four comparisons shown in
Table~\ref{tab:comparison_brainfound}. These results are obtained with a smaller
ViT-Base backbone, without natural-image initialization, and using individual
MRI contrasts rather than simultaneous multi-contrast inputs. Because the two
studies differ in data partitions, pretraining data composition, model scale,
and evaluation protocols, these comparisons should be interpreted as
cross-study reference points rather than controlled head-to-head benchmarks.

\begin{table}[!htbp]
\centering
\caption{Comparison with BrainFound~\cite{Mazher2025} on overlapping
downstream tasks. Direct numerical comparison is approximate because of
differences in data splits, preprocessing, pretraining data, and evaluation
protocols across studies. BrainFound uses multiple MRI contrasts (T1, T2,
FLAIR) and is pretrained on 10,000 volumes across 12 datasets; our models
process individual MRI contrasts and are pretrained on 58,781 volumes across
seven datasets.}
\label{tab:comparison_brainfound}
\setlength{\tabcolsep}{4pt}
\begin{tabular}{lcc}
\toprule
\textbf{Task} & \textbf{BrainFound} & \textbf{Ours (best)} \\
\midrule
ADNI NC vs.\ AD              & 0.810          & $\mathbf{0.945 \pm 0.024}$ (SL-MAE) \\
ADNI NC vs.\ MCI             & 0.766 & $\textbf{0.780}$ (MAE)    \\
NACC/SCAN NC vs.\ AD      & \textbf{0.883} & $0.841$ (SL-MAE)           \\
UCSF Tumor Grading           & 0.780          & $\mathbf{0.833}$ (SL-MAE)  \\
\bottomrule
\end{tabular}
\end{table}

\subsection{t-SNE Visualization and Clustering Metrics}

To further characterize class structure in the learned representation spaces,
we visualize encoder embeddings using t-distributed Stochastic Neighbor
Embedding (t-SNE), which projects high-dimensional representations into two
dimensions while approximately preserving local neighborhood structure
\cite{vanDerMaaten2008}. We extract 768-dimensional embeddings from the
held-out ADNI test set ($N=284$) and examine pretrained and fine-tuned variants
of MAE, SL-MAE, JEPA, and VCR-JEPA. Class separability is quantified in
the original embedding space using the Silhouette Score computed with the
ground-truth diagnostic labels. To additionally assess unsupervised cluster
agreement with the diagnostic classes, we apply $k$-means clustering with the
number of clusters set equal to the number of classes ($k=2$ for NC vs.\ AD)
and compute Normalized Mutual Information (NMI) between the resulting cluster
assignments and the ground-truth labels. $k$-means is run with a fixed random
seed and $n_{\mathrm{init}}=10$. The same metrics are also computed in the
two-dimensional t-SNE space for descriptive comparison. The resulting
visualizations are shown in Fig.~\ref{fig:tsne_comparison}.

Table~\ref{tab:clustering} summarizes the separability metrics across encoder
variants. In the original embedding space, the auxiliary-objective variants
yield numerically higher Silhouette Scores and NMI than their corresponding
baselines in both pretrained and fine-tuned settings. Before fine-tuning,
however, all values remain close to zero, indicating little explicit
disease-class separation in the pretrained representation spaces. After
fine-tuning, class structure becomes substantially more pronounced. SL-MAE
achieves higher Silhouette Score (0.9218 vs.\ 0.8127) and NMI
(0.8977 vs.\ 0.8387) than MAE, while VCR-JEPA similarly exceeds JEPA
(Silhouette: 0.8123 vs.\ 0.7865; NMI: 0.6919 vs.\ 0.6523). Definitions and implementation details for the t-SNE visualization and
representation-separability metrics are provided in Supplementary
Section~S2.4.

In the t-SNE projection, SL-MAE also shows greater class separation than MAE
(Silhouette: 0.0197 vs.\ $-$0.0015 pretrained; 0.7513 vs.\ 0.7379
fine-tuned). For VCR-JEPA, the advantage is retained after fine-tuning
(Silhouette: 0.6871 vs.\ 0.6313; NMI: 0.6919 vs.\ 0.6782), whereas pretrained
JEPA yields a slightly higher t-SNE Silhouette Score than pretrained VCR-JEPA
(0.0251 vs.\ 0.0162). We attribute this discrepancy to the known tendency of t-SNE to emphasize local neighborhood structure while distorting aspects of the global geometry of the embedding space. Because VCR explicitly encourages feature variance while reducing redundancy across latent dimensions, its representational
effects may be reflected more faithfully in the original high-dimensional
embedding space than in the two-dimensional t-SNE projection.

\begin{table}[!htbp]
\centering
\caption{Clustering quality metrics for all encoders computed on RAW
embedding space (768\,D) and t-SNE 2-D projection space (NC vs.\ AD,
$N{=}284$). $\uparrow$: higher is better.
\textbf{Bold}: best result per column within each training stage.
$^\dagger$: overall best across all fine-tuned encoders.}
\label{tab:clustering}
\setlength{\tabcolsep}{4pt}
\begin{tabular}{llcccc}
\toprule
& &
\multicolumn{2}{c}{\textbf{RAW (768\,D)}} &
\multicolumn{2}{c}{\textbf{t-SNE 2-D}} \\
\cmidrule(lr){3-4}\cmidrule(lr){5-6}
\textbf{Stage} & \textbf{Encoder} &
\textbf{Sil.\,$\uparrow$} &
\textbf{NMI\,$\uparrow$} &
\textbf{Sil.\,$\uparrow$} &
\textbf{NMI\,$\uparrow$} \\
\midrule
\multirow{4}{*}{\rotatebox[origin=c]{90}{\textit{Pretrained}}}
  & MAE     & 0.0076 & 0.0013 & -0.0015 & 0.0018 \\
  & SL-MAE  & \textbf{0.0105} & \textbf{0.0035} & 0.0197 & 0.0006 \\
  & JEPA    & 0.0040 & 0.0000 & \textbf{0.0251} & \textbf{0.0032} \\
  & VCR-JEPA& 0.0052 & 0.0002 & 0.0162 & 0.0001 \\
\midrule
\multirow{4}{*}{\rotatebox[origin=c]{90}{\textit{Finetuned}}}
  & MAE     & 0.8127 & 0.8387 & 0.7379 & 0.8364 \\
  & SL-MAE  & $\mathbf{0.9218}^\dagger$ & $\mathbf{0.8977}^\dagger$ & $\mathbf{0.7513}^\dagger$ & $\mathbf{0.8977}^\dagger$ \\
  & JEPA    & 0.7865 & 0.6523 & 0.6313 & 0.6782 \\
  & VCR-JEPA& 0.8123 & 0.6919 & 0.6871 & 0.6919 \\
\bottomrule
\end{tabular}
\end{table}

\begin{figure}[pos=htbp]
\centering

\setlength{\tabcolsep}{2pt}
\renewcommand{\arraystretch}{0.5}

\begin{tabular}{cccc}

{\small\textbf{MAE}} &
{\small\textbf{SL-MAE}} &
{\small\textbf{JEPA}} &
{\small\textbf{VCR-JEPA}} \\[3pt]

\includegraphics[width=0.245\linewidth]
{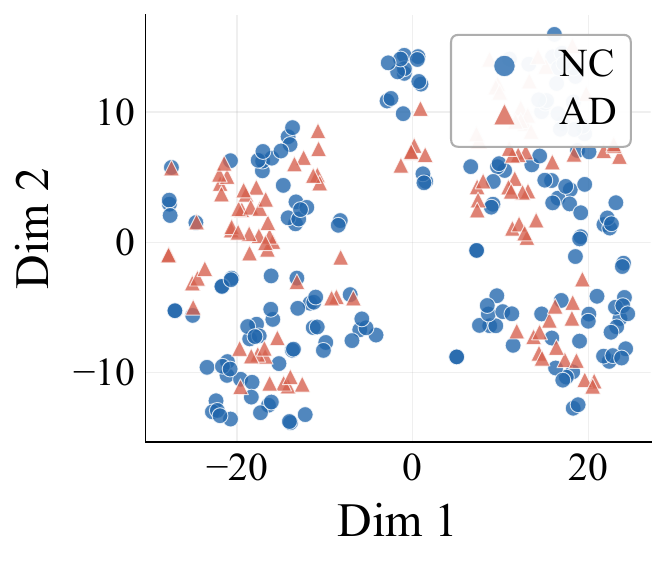} &
\includegraphics[width=0.245\linewidth]
{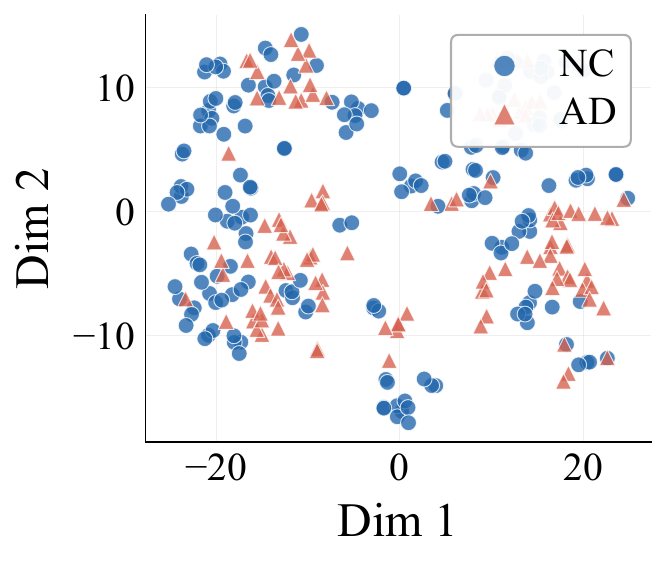} &
\includegraphics[width=0.245\linewidth]
{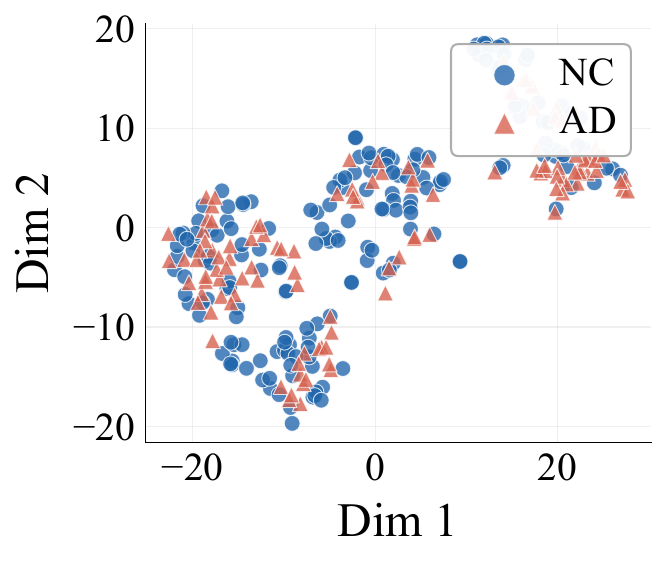} &
\includegraphics[width=0.245\linewidth]
{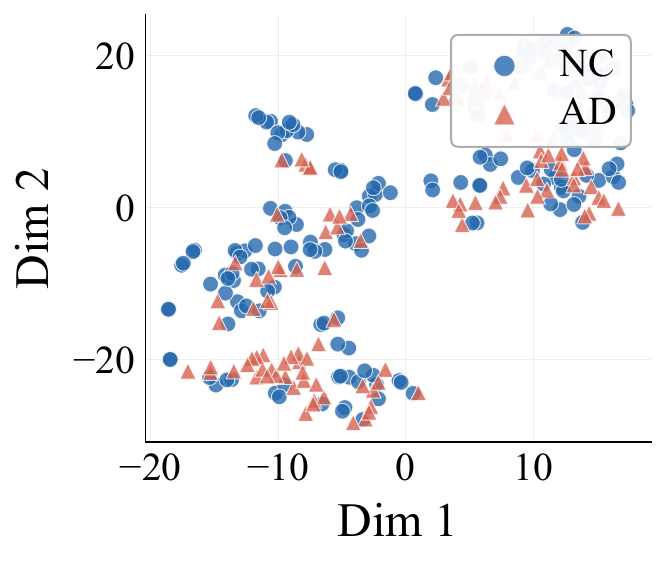} \\[-1pt]

\multicolumn{4}{c}{
    {\small (a)--(d)~~Pretrained encoders}
} \\[5pt]

\includegraphics[width=0.245\linewidth]
{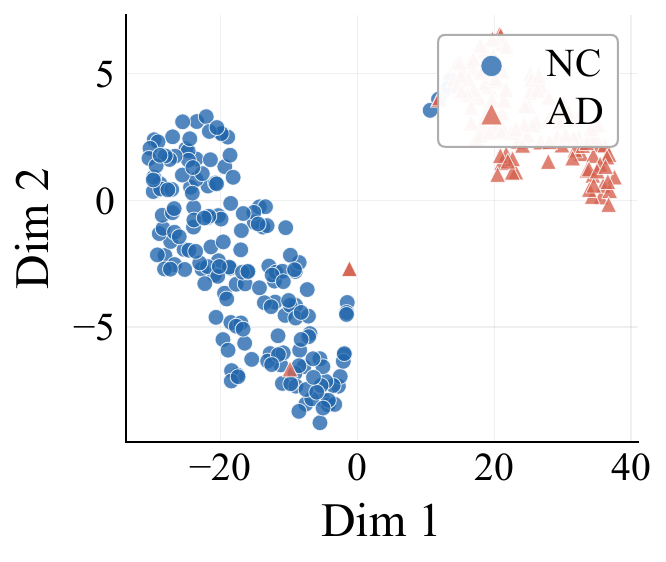} &
\includegraphics[width=0.245\linewidth]
{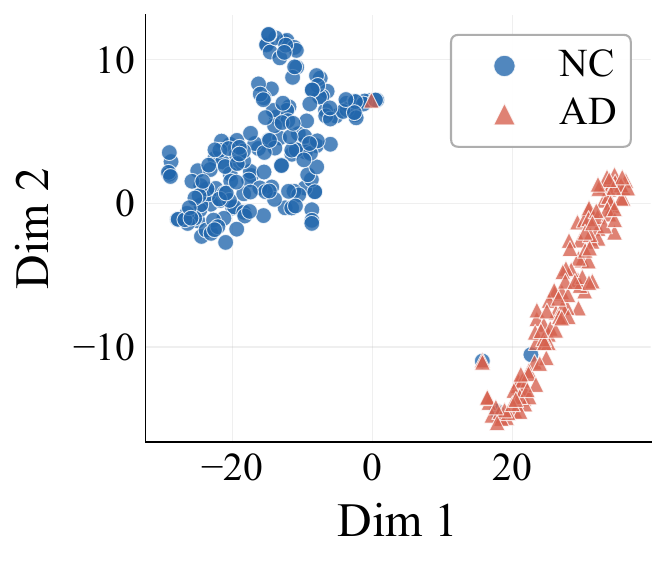} &
\includegraphics[width=0.245\linewidth]
{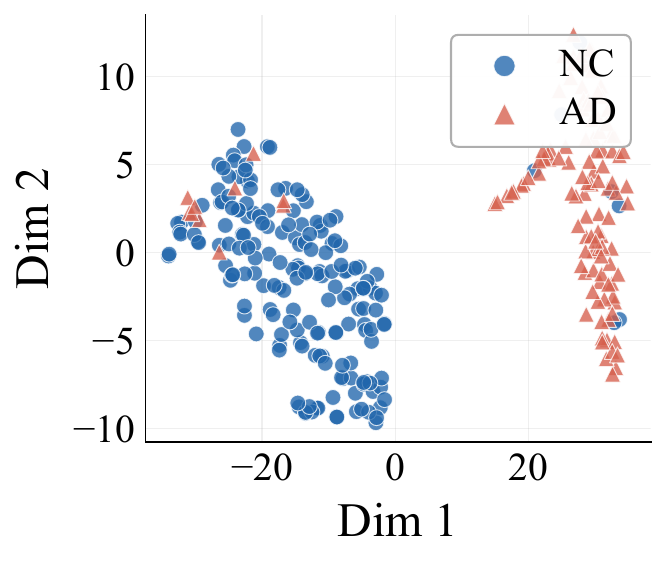} &
\includegraphics[width=0.245\linewidth]
{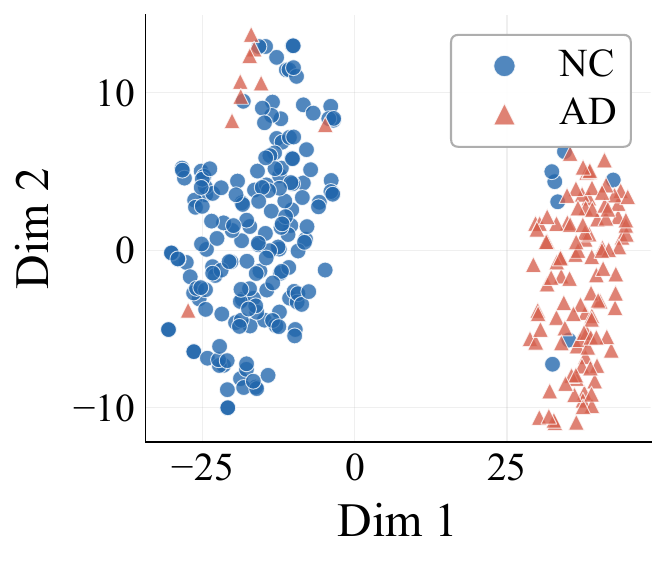} \\[-1pt]

\multicolumn{4}{c}{
    {\small (e)--(h)~~Fine-tuned encoders}
} \\

\end{tabular}

\caption{
t-SNE visualizations of encoder embeddings on the held-out test
set ($N{=}284$), projected from 768-D to 2-D.
\textbf{Columns} (left to right): MAE, SL-MAE, JEPA, and VCR-JEPA.
\textbf{Rows}: pretrained~(a--d) and fine-tuned~(e--h) encoders.
Blue circles~(\textcolor{ncblue}{$\bullet$}) denote Normal Control~(NC),
and red triangles~(\textcolor{adorange}{$\blacktriangle$}) denote
Alzheimer's Disease~(AD).
Pretrained encoders~(a--d) exhibit heavily intermixed NC/AD
distributions, reflecting the label-agnostic nature of
self-supervised pretraining. Following fine-tuning~(e--h),
MAE and SL-MAE produce visibly more compact and separated clusters,
while VCR-JEPA achieves competitive separation; see
Table~\ref{tab:clustering} for quantitative clustering metrics.
}

\label{fig:tsne_comparison}

\end{figure}
Across these analyses, the strongest class separation emerges after
fine-tuning, with SL-MAE and VCR-JEPA exhibiting greater separability than
their respective baselines in the original embedding space. These findings
complement the downstream classification results and indicate that the
auxiliary objectives influence how disease-related class structure emerges
during task-specific adaptation.

\subsection{Representational Similarity Analysis via Centered Kernel Alignment}

To characterize how self-supervised objective design and downstream adaptation shape representation geometry, we analyze four encoder variants—MAE, SL-MAE, JEPA, and VCR-JEPA—at their pretrained and task-fine-tuned checkpoints using Centered Kernel Alignment (CKA) \cite{kornblith2019similarity}. The analysis is performed on held-out ADNI cohorts for both NC vs.\ AD and NC vs.\ MCI, with identical subjects used across encoders within each comparison.

For each encoder, we extracted mean-pooled embeddings from fixed evaluation cohorts comprising $N=300$ test subjects for ADNI NC-vs-AD, $N=300$ test subjects for ADNI NC-vs-MCI, and $N=763$ test samples for the UCSF analyses. For each analysis, all encoders were evaluated on the same fixed set of test subjects, with identical subject ordering maintained across encoders to ensure row-wise correspondence of the representation matrices used for CKA. Given two embedding matrices $X \in \mathbb{R}^{n \times d_1}$ and $Y \in \mathbb{R}^{n \times d_2}$ from a pair of encoders, CKA operates not on the raw features directly but on the $n \times n$ Gram matrices

\begin{equation}
K^{\text{lin}} = XX^\top, \qquad
K^{\text{rbf}}_{ij} = \exp\!\left(-\frac{\lVert x_i - x_j \rVert^2}{2\sigma^2}\right),
\label{eq:gram}
\end{equation}

(linear and RBF kernels, respectively), which encode the pairwise similarity structure among subjects under each representation. For the RBF kernel, the bandwidth is set to $\sigma=0.4$ times the median pairwise Euclidean distance between representations.

To reduce finite-sample bias, we compute CKA using the unbiased HSIC estimator. Before computing HSIC, the diagonal entries of the Gram matrices are set to zero. For two such Gram matrices $K$ and $L$, the unbiased estimator is

\begin{equation}
\begin{aligned}
\mathrm{HSIC}_1(K,L)
=
\frac{1}{n(n-3)}
\Bigg[
&\operatorname{tr}(KL)
+
\frac{
(\mathbf{1}^{\top}K\mathbf{1})
(\mathbf{1}^{\top}L\mathbf{1})
}{
(n-1)(n-2)
}
\\
&-
\frac{2}{n-2}
\mathbf{1}^{\top}KL\mathbf{1}
\Bigg].
\end{aligned}
\label{eq:hsic-unbiased}
\end{equation}

The alignment score is then computed as

\begin{equation}
\mathrm{CKA}(X,Y)
=
\frac{\mathrm{HSIC}_1(K_X,K_Y)}
{\sqrt{
\mathrm{HSIC}_1(K_X,K_X)
\mathrm{HSIC}_1(K_Y,K_Y)
}},
\label{eq:cka}
\end{equation}

where larger values indicate greater representational alignment. Because we use an unbiased finite-sample HSIC estimator, values close to zero may occasionally be slightly negative. CKA is invariant to orthogonal transformations and isotropic scaling of either representation and is well defined even when $d_1 \neq d_2$.

We use linear CKA as the primary similarity measure throughout our analyses. For pretrained--fine-tuned and inter-objective comparisons, we additionally report RBF-kernel CKA as a robustness check to assess whether the observed similarity patterns persist beyond linear relationships between representations. The corresponding RBF-kernel results are provided in the Supplementary Material. In our models, the representation dimensions are equal, $d_1=d_2$.

To localize where in network depth the pretrained--fine-tuned divergence originates, we extended this analysis to individual transformer blocks. For each of the four matched pretrained/fine-tuned pairs (one per pretraining objective), we hooked every block of the ViT-Base backbone (12 blocks total), and mean-pooled over tokens to obtain one embedding per subject per layer. We report two complementary views of this layer-resolved similarity: a \emph{depth profile} comparing matched layer indices (pretrained layer $i$ vs.\ fine-tuned layer $i$) across depth for each pretraining objective, and a full \emph{layer $\times$ layer} grid for two focal pairs (JEPA and MAE), which additionally reveals whether alignment occurs at shifted, non-matching depths rather than only at matched indices. The first transformer block was excluded from the reported layer-wise profiles due to its heightened sensitivity to intermediate-feature normalization and the resulting instability in similarity estimates. Accordingly, we report the remaining 11 post-initial transformer depths consistently across all encoders.


\begin{figure}[pos=htbp]
    \centering
    \begin{subfigure}[b]{0.48\textwidth}
        \centering
        \includegraphics[width=0.8\textwidth]{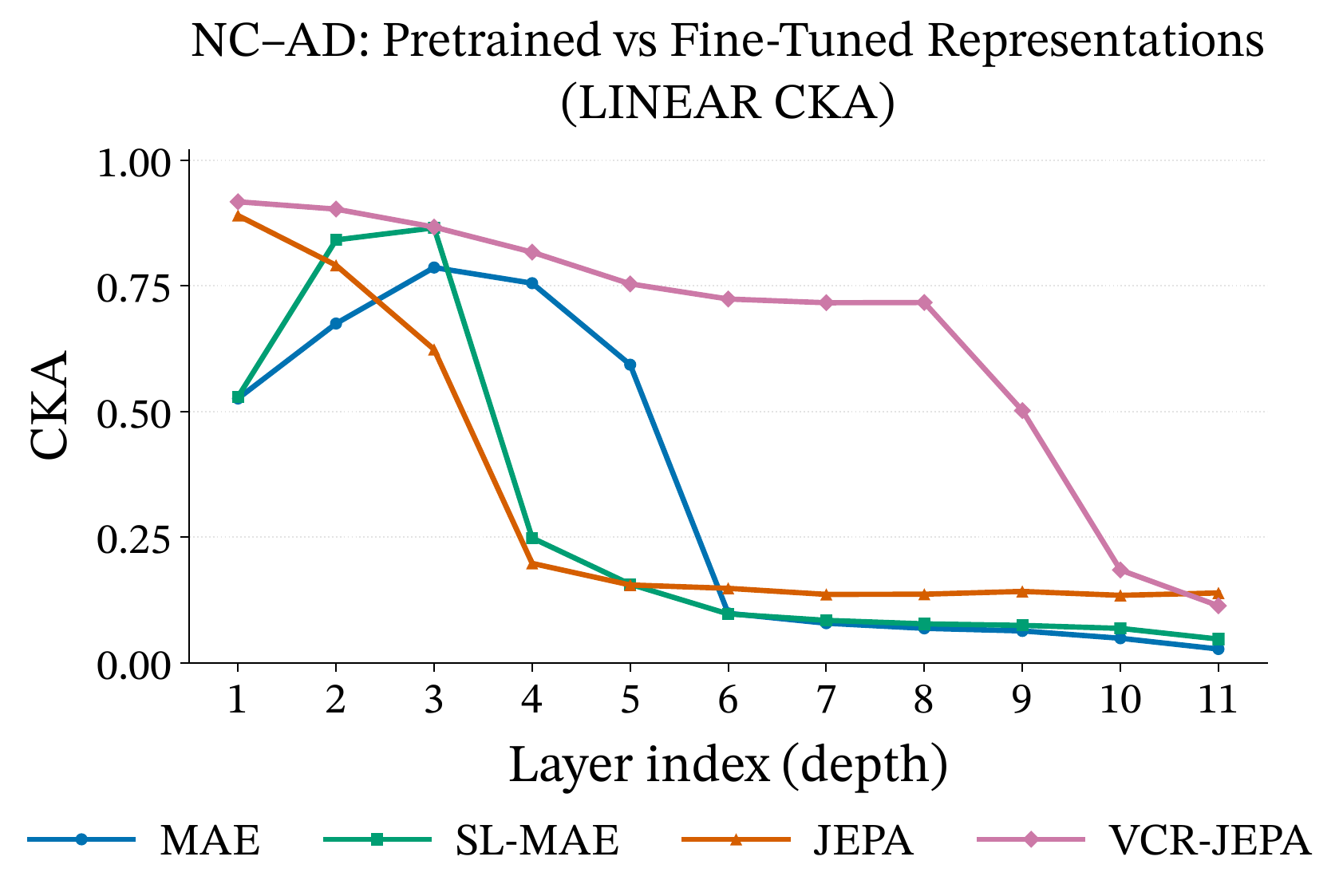}
        \caption{Linear CKA (AD)}
        \label{fig:depth-profile-matched-linear}
    \end{subfigure}
    \hfill
    \begin{subfigure}[b]{0.48\textwidth}
        \centering
        \includegraphics[width=0.8\textwidth]{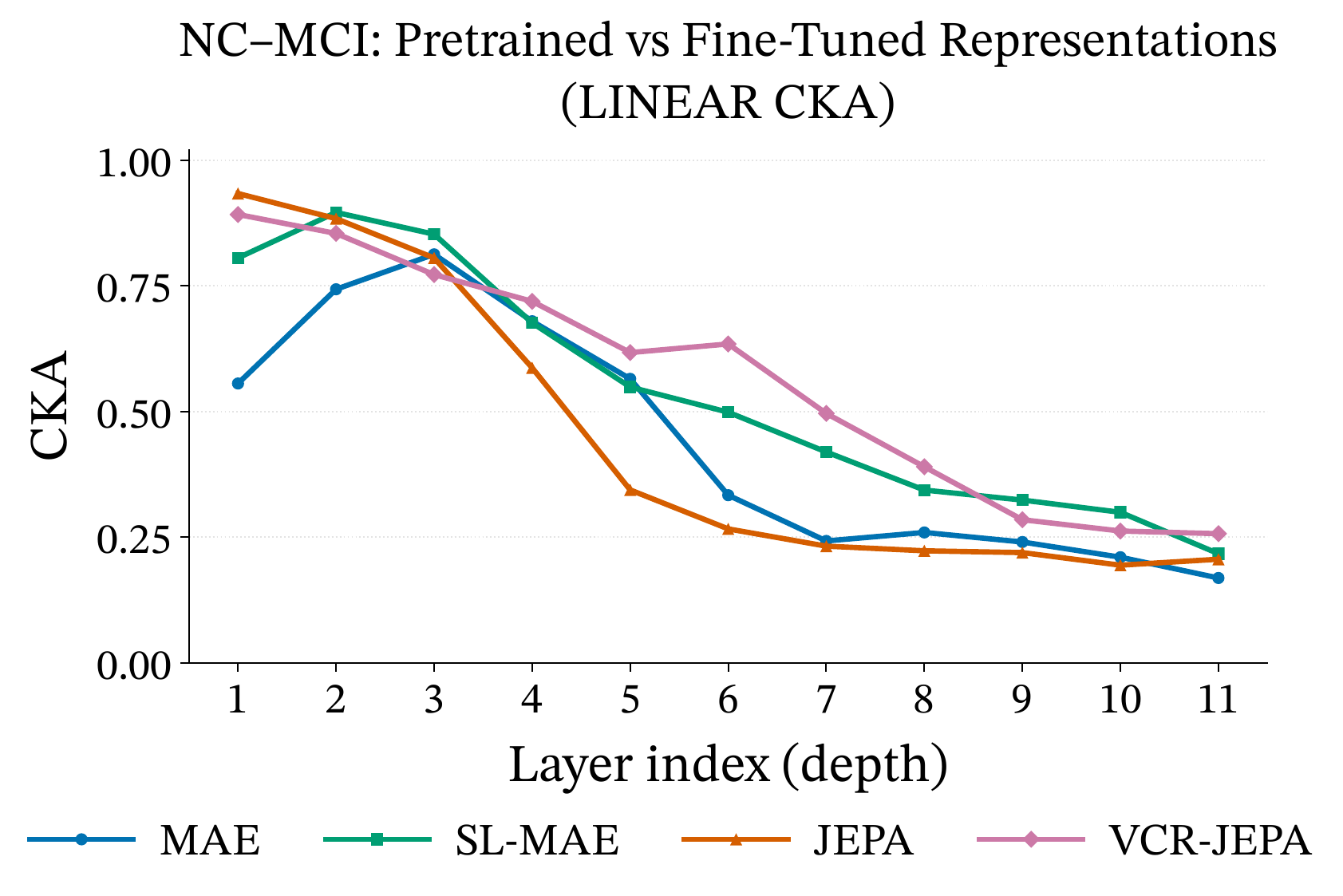}
        \caption{Linear CKA (MCI)}
        \label{fig:depth-profile-matched-linear-mci}
    \end{subfigure}

    \vspace{0.6em}

\caption{Layer-wise CKA between each encoder and its own fine-tuned
counterpart (MAE, SL-MAE, JEPA, and VCR-JEPA) using the linear kernel,
on ADNI NC-vs-AD (a) and NC-vs-MCI (b).
Corresponding RBF-kernel results are provided in Supplementary Fig.~S1.
}
    \label{fig:depth-profile-matched}
\end{figure}

Figure~\ref{fig:depth-profile-matched} compares each pretrained encoder with its task-fine-tuned counterpart at matched network depths. Across both tasks, pretrained-to-fine-tuned similarity generally decreases with depth, but the onset, magnitude, and progression of this divergence depend jointly on the pretraining objective and downstream task. VCR-JEPA retains substantial pretrained alignment deeper into the encoder than the other variants, particularly under NC vs.\ AD fine-tuning. The task contrast is also pronounced: NC vs.\ AD induces a sharper transition toward low similarity in later layers, whereas NC vs.\ MCI exhibits a more gradual decline and retains greater pretrained alignment across depth.



\newcolumntype{R}{>{\centering\arraybackslash}m{1.2em}}
\newcolumntype{P}{>{\centering\arraybackslash}m{0.26\textwidth}}
\newcommand{\rlab}[1]{\rotatebox[origin=c]{90}{\small #1}}
\newcommand{\pan}[1]{\includegraphics[width=\linewidth]{#1}}

\begin{figure}[pos=p]
  \centering

  \includegraphics[width=0.90\textwidth]
  {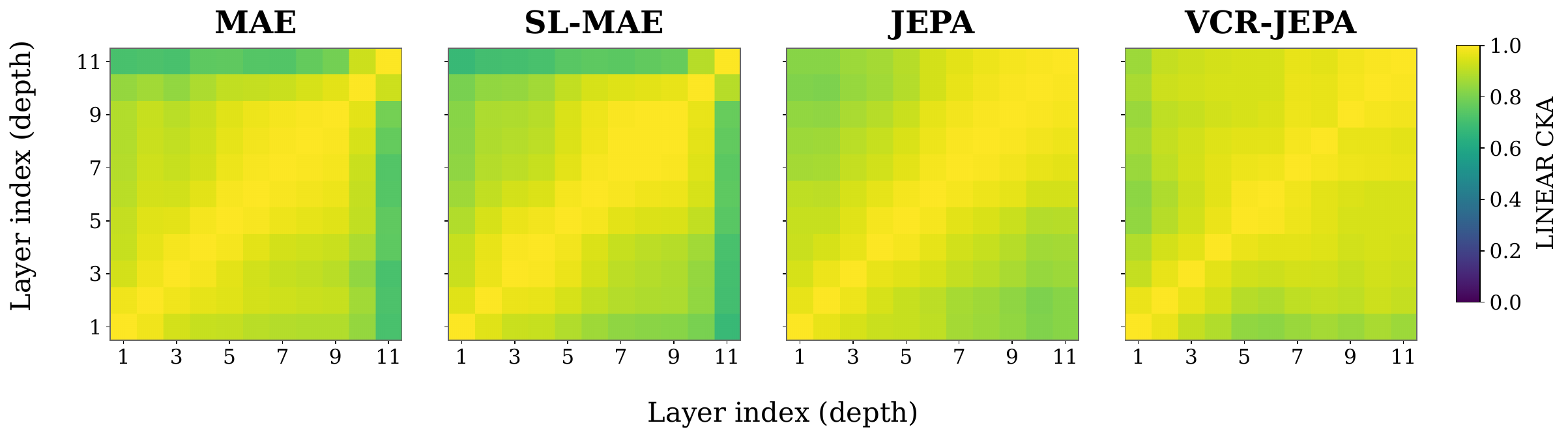}

  {\footnotesize (a) Pretraining Self-CKA maps - NC vs.\ AD\par}

  \vspace{5pt}

  \includegraphics[width=0.90\textwidth]
  {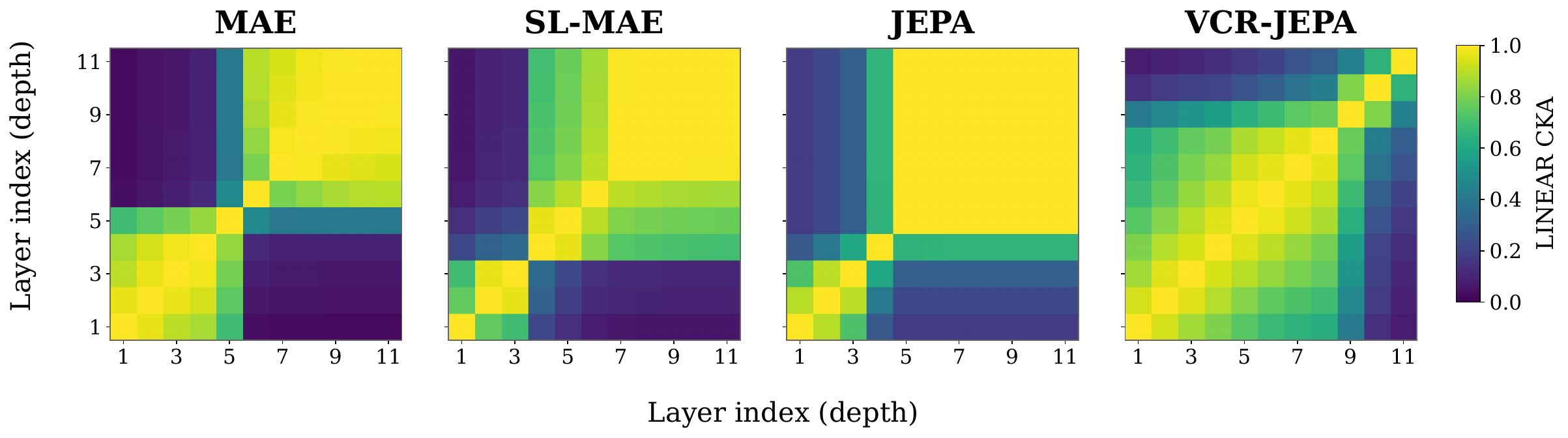}

  {\footnotesize (b) Finetuning Self-CKA maps - NC vs.\ AD\par}

  \vspace{5pt}

  \includegraphics[width=0.90\textwidth]
  {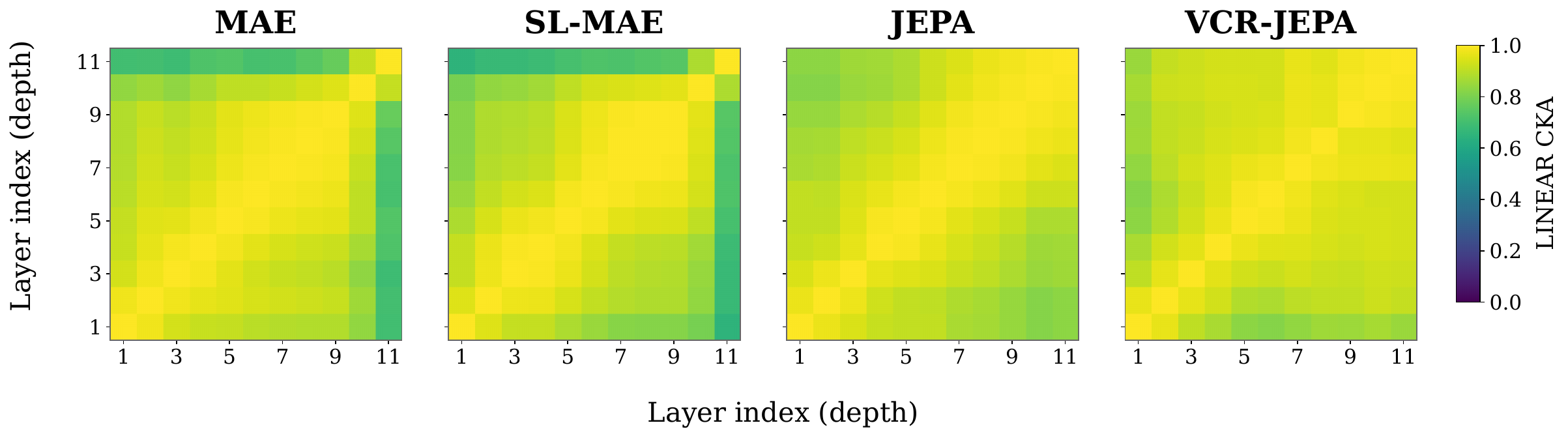}

  {\footnotesize (c) Pretraining Self-CKA maps - NC vs.\ MCI\par}

  \vspace{5pt}

  \includegraphics[width=0.90\textwidth]
  {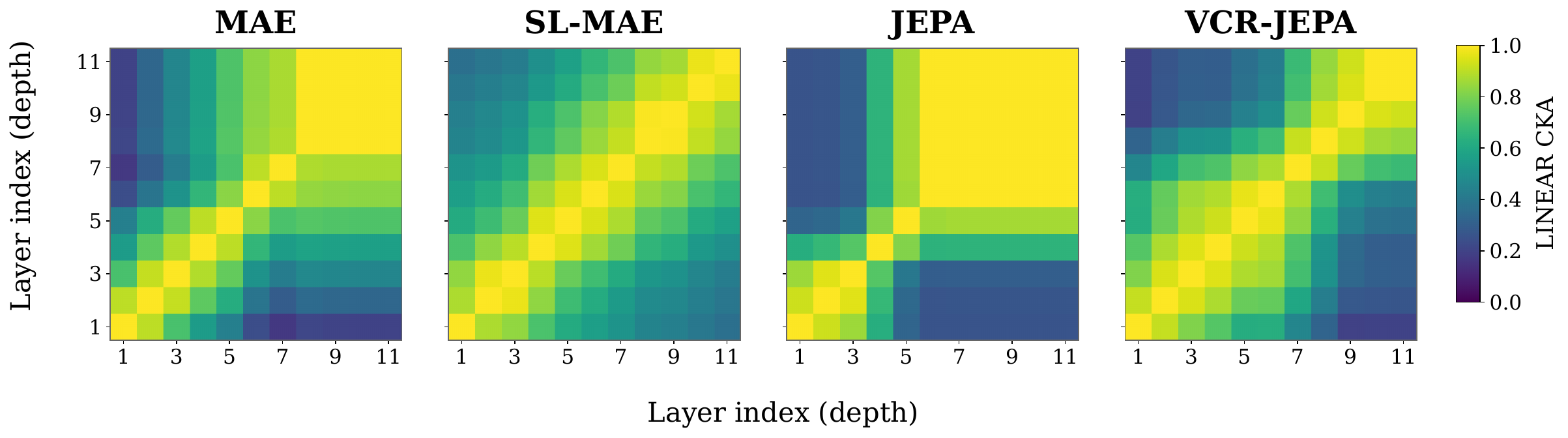}

  {\footnotesize (d) Finetuning Self-CKA maps - NC vs.\ MCI\par}

  \caption{Layer-by-layer linear self-CKA of each encoder before and after fine-tuning on ADNI NC-vs-AD and NC-vs-MCI. Pretrained encoders exhibit broadly high off-diagonal similarity across depth, whereas fine-tuning produces stronger depth differentiation and block-like organization. This transition is sharper for NC-vs-AD than for NC-vs-MCI, while VCR-JEPA exhibits a comparatively gradual fine-tuned organization across layers. The pretrained checkpoints are task-independent; differences between pretrained rows arise from the evaluation cohort. All panels share a common $0$--$1$ colour scale.}
  \label{fig:selfcka}
\end{figure}

Figure~\ref{fig:selfcka} provides a complementary view of the adaptation process by examining the internal layer-wise organization of each encoder under pretraining and fine-tuning for the NC vs.\ AD and NC vs.\ MCI tasks. During pretraining, all four models exhibit broadly high self-CKA similarity across intermediate layers, indicating relatively limited depth differentiation. Fine-tuning substantially alters this organization, with distinct blocks or transitions emerging across depth. This reorganization is sharper under NC vs.\ AD fine-tuning than under NC vs.\ MCI. VCR-JEPA differs from baseline JEPA in exhibiting a smoother, more distributed transition across layers rather than an abrupt separation between early and late representational regimes.

\begin{figure}[pos=H]
    \centering
    \begin{subfigure}[b]{0.235\textwidth}
        \centering
        \includegraphics[width=\textwidth]{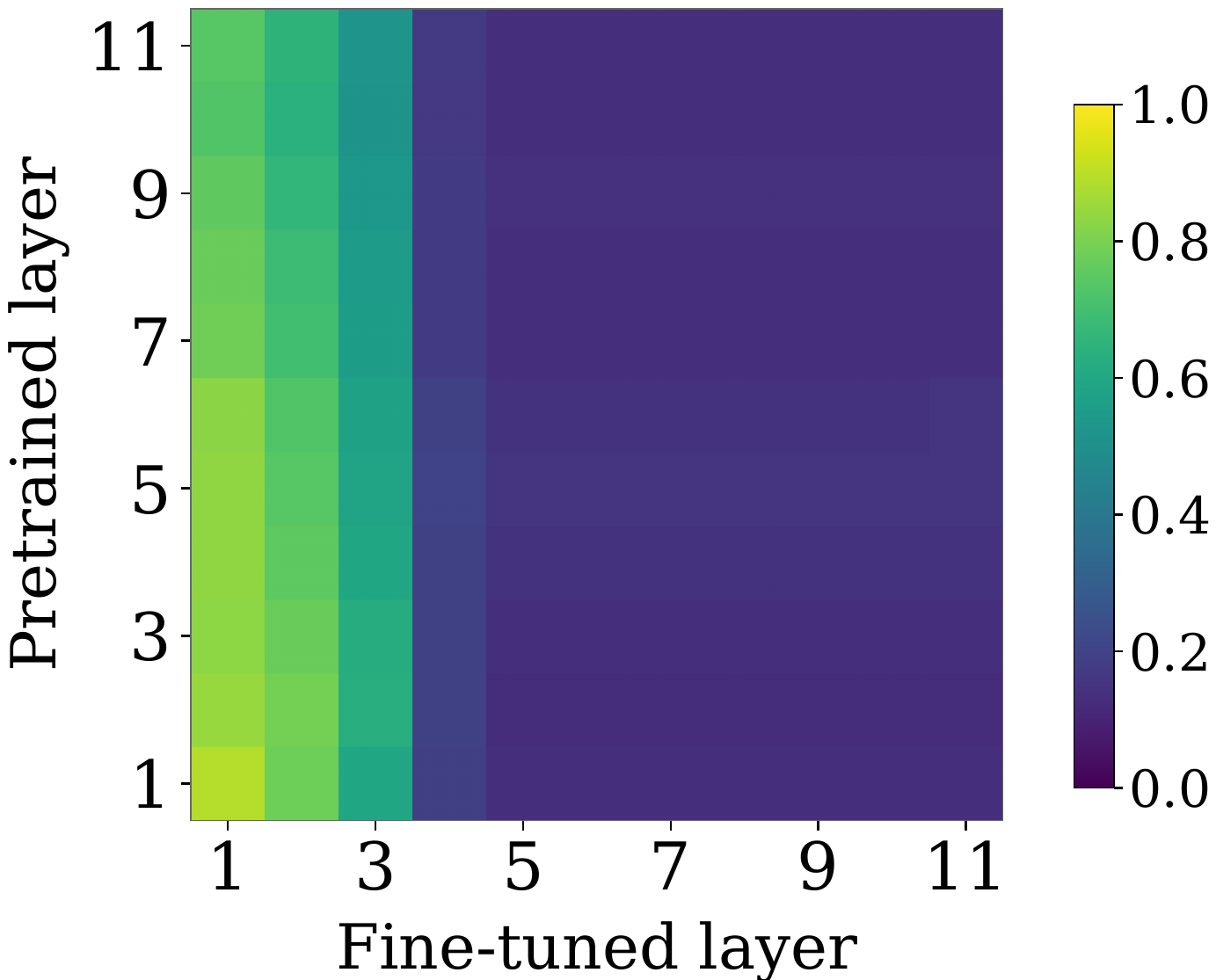}
        \caption{JEPA -- Linear (AD)}
        \label{fig:layer-heatmap-jepa-linear}
    \end{subfigure}
    \hfill
    \begin{subfigure}[b]{0.235\textwidth}
        \centering
        \includegraphics[width=\textwidth]{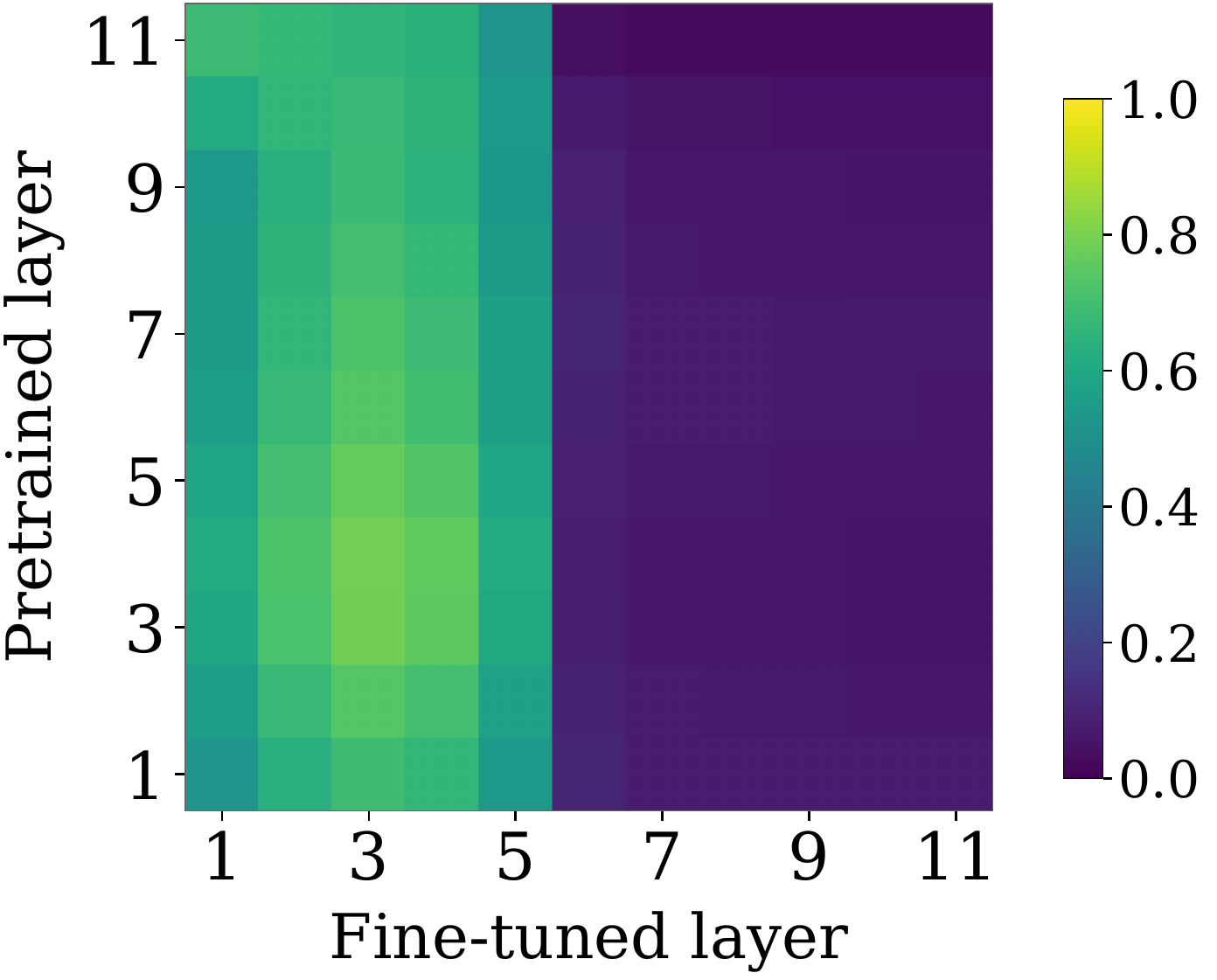}
        \caption{MAE -- Linear (AD)}
        \label{fig:layer-heatmap-mae-linear}
    \end{subfigure}
    \hfill
    \begin{subfigure}[b]{0.235\textwidth}
        \centering
        \includegraphics[width=\textwidth]{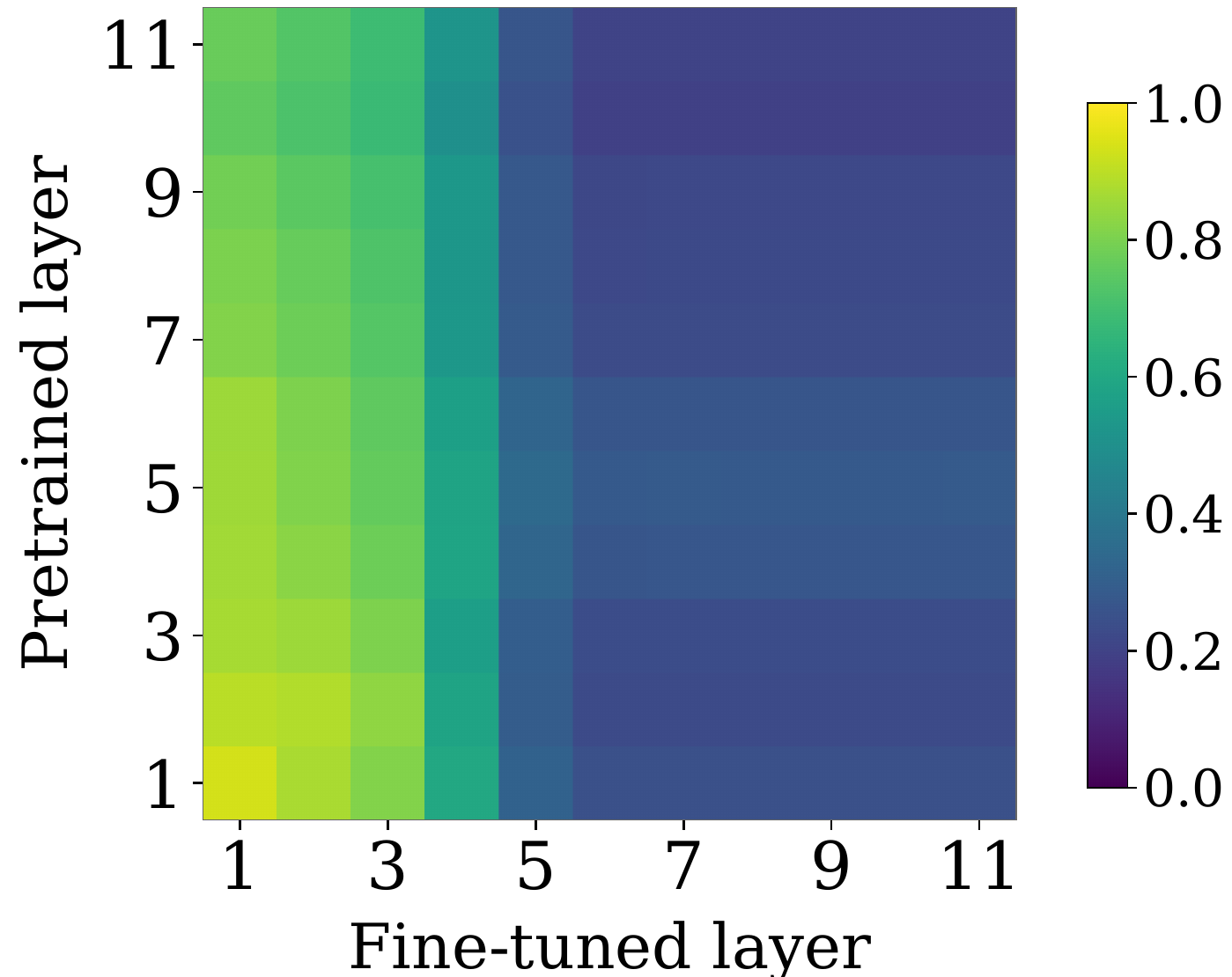}
        \caption{JEPA -- Linear (MCI)}
        \label{fig:layer-heatmap-jepa-linear-mci}
    \end{subfigure}
    \hfill
    \begin{subfigure}[b]{0.235\textwidth}
        \centering
        \includegraphics[width=\textwidth]{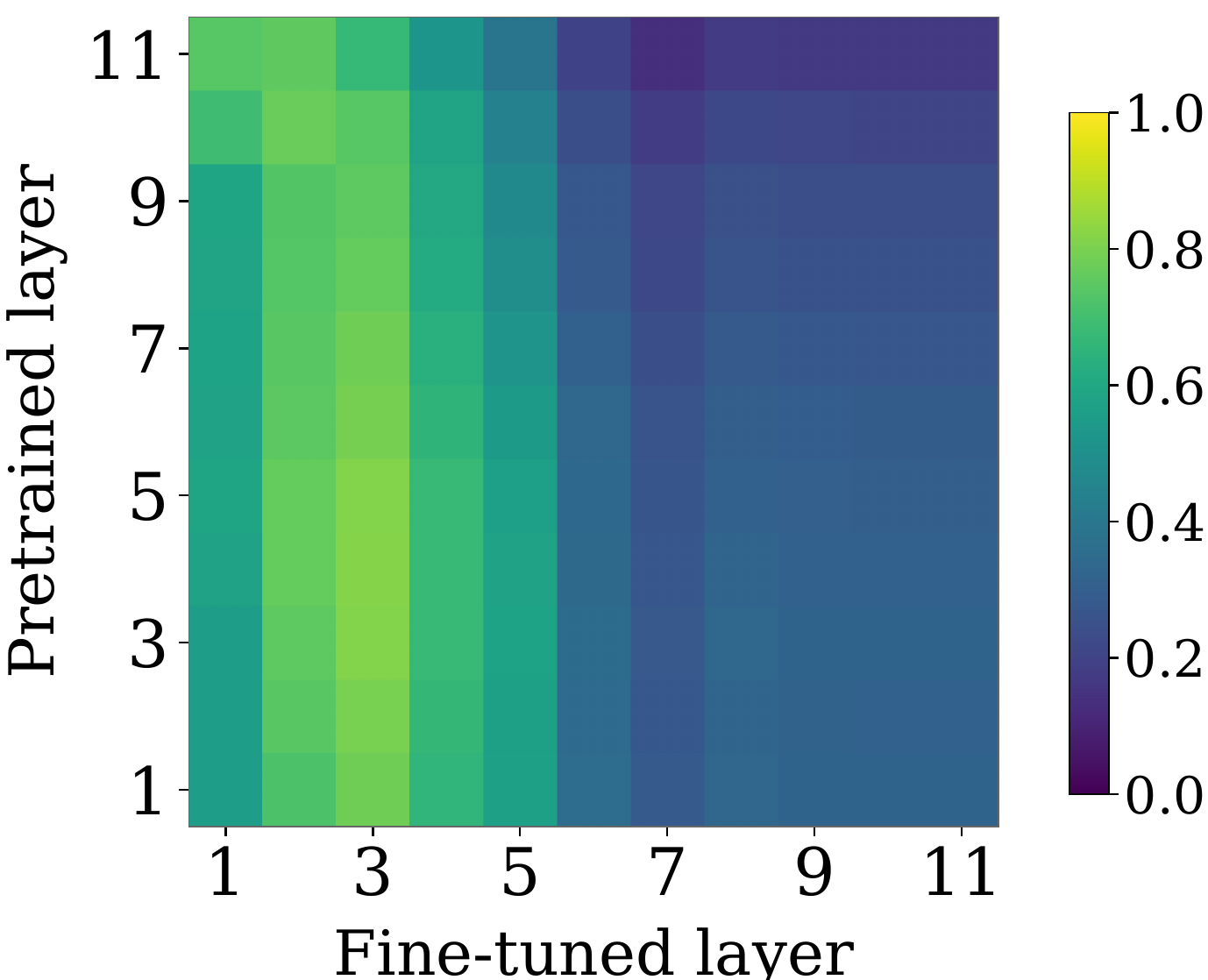}
        \caption{MAE -- Linear (MCI)}
        \label{fig:layer-heatmap-mae-linear-mci}
    \end{subfigure}

    \vspace{0.6em}

\caption{Layer-wise CKA between pretrained (y-axis) and fine-tuned (x-axis)
encoder representations for JEPA and MAE using the linear kernel on ADNI
NC-vs-AD (left) and NC-vs-MCI (right). Corresponding RBF-kernel results
are provided in Supplementary Fig.~S2.
}
    \label{fig:pre-fine-layer-heat}
\end{figure}

Figure~\ref{fig:pre-fine-layer-heat} illustrates the layer-wise CKA similarity between pretrained and fine-tuned encoders for both the NC vs.\ AD and NC vs.\ MCI tasks. In the NC vs.\ AD heatmaps, later fine-tuned layers exhibit substantially lower similarity to the pretrained representation hierarchy than those observed for NC vs.\ MCI. Consistent with the self-CKA analysis in Figure~\ref{fig:selfcka}, the full layer-by-layer maps reveal a depth-dependent transition in which fine-tuned representations progressively depart from the pretrained geometry. The depth at which this transition occurs varies with the pretraining objective, while the magnitude and progression of the representational change also depend on the downstream task. In particular, NC vs.\ AD produces a substantially sharper loss of pretrained alignment than NC vs.\ MCI, consistent with the matched-layer profiles in Figure~\ref{fig:depth-profile-matched}. This pattern is consistent with the more subtle and heterogeneous discriminative structure of NC vs.\ MCI discussed in Sections 6.1 and 6.2: the weaker task-specific signal may require less departure from the pretrained geometry. We therefore interpret the smaller CKA displacement for NC vs.\ MCI as reduced task-specific restructuring, while noting that CKA characterizes the extent of representational adaptation rather than its causal origin.


\begin{figure}[pos=htbp]
    \centering

    \begin{tabular}{@{}cccc@{}}
        \includegraphics[width=0.23\textwidth]{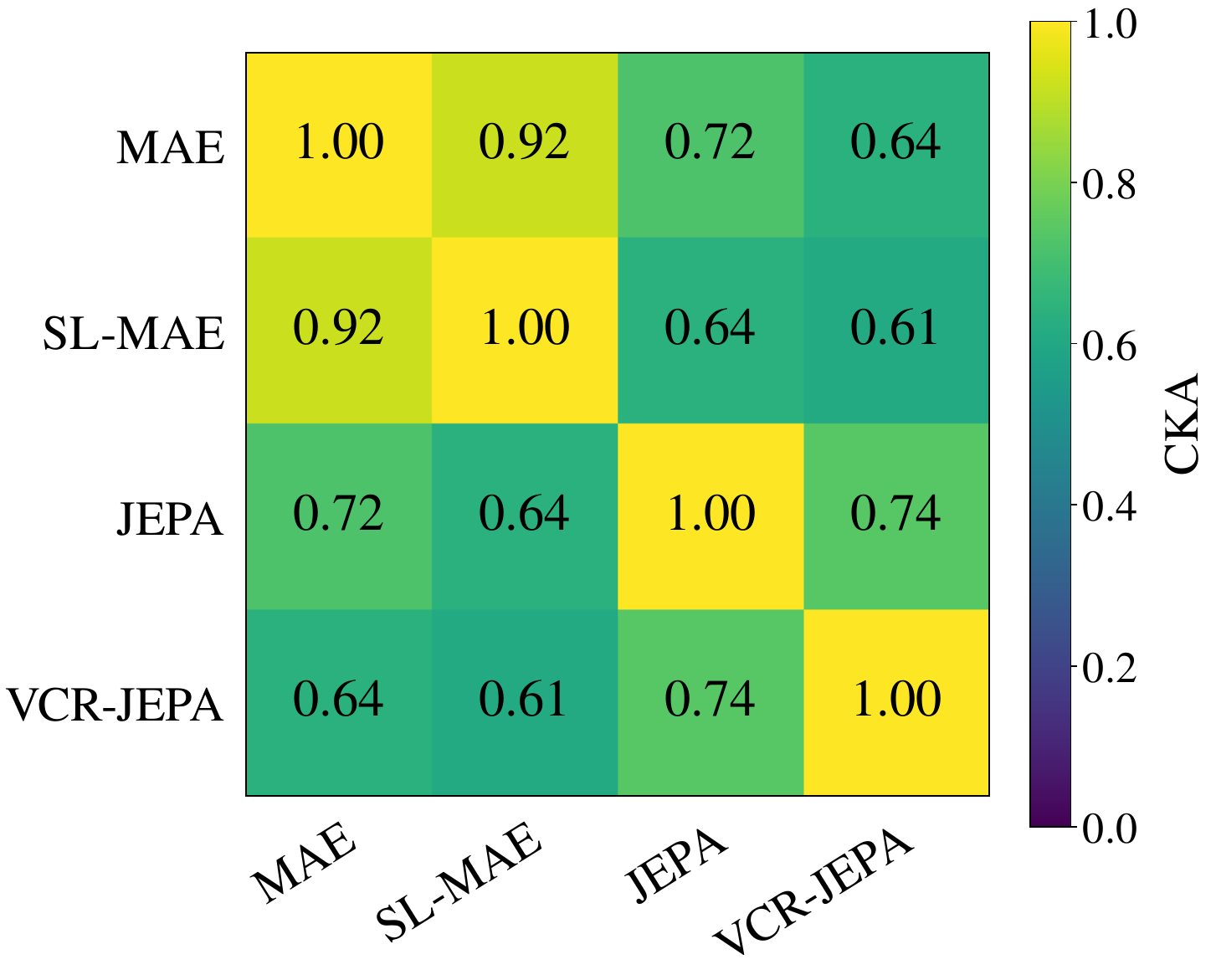} &
        \includegraphics[width=0.23\textwidth]{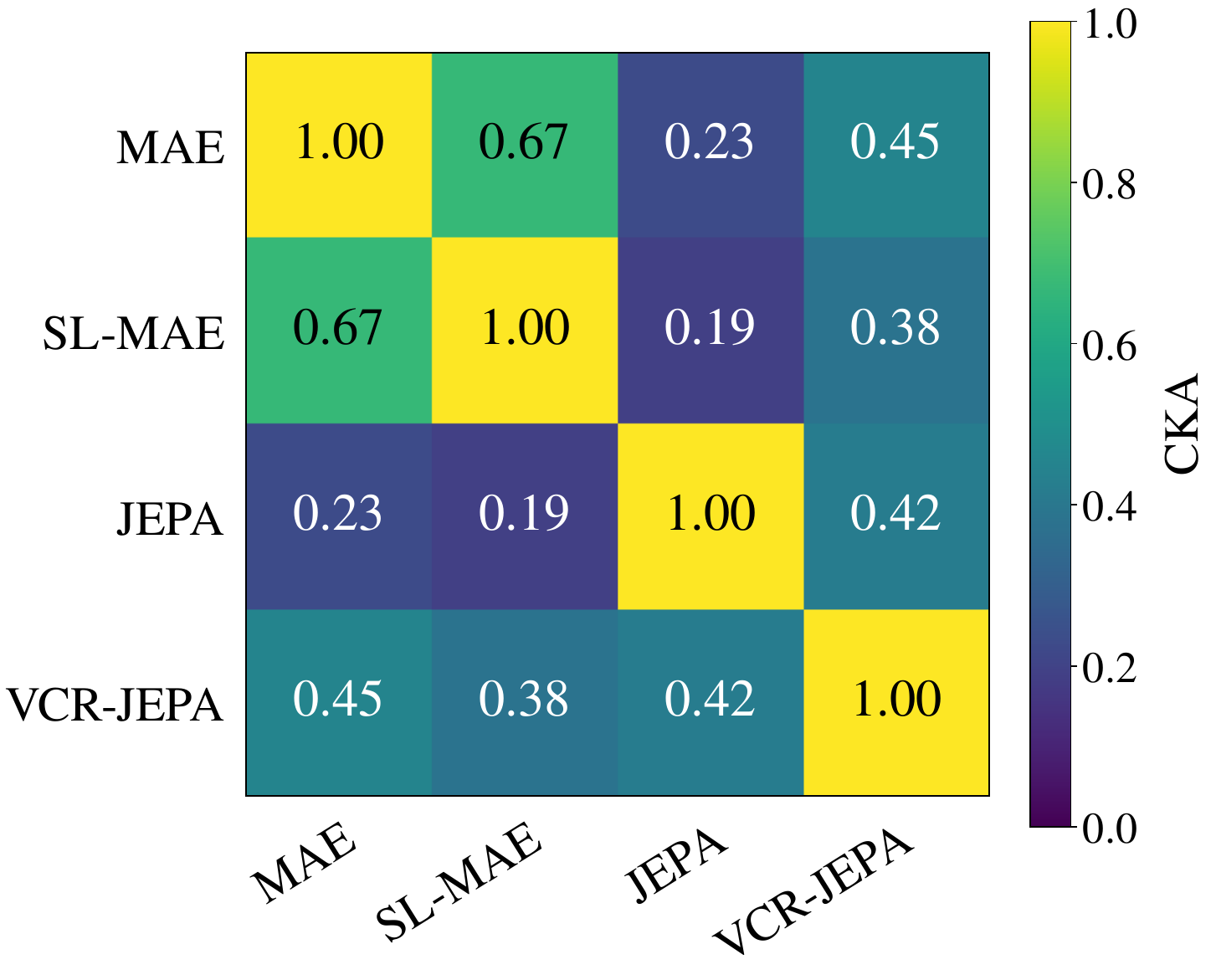} &
        \includegraphics[width=0.23\textwidth]{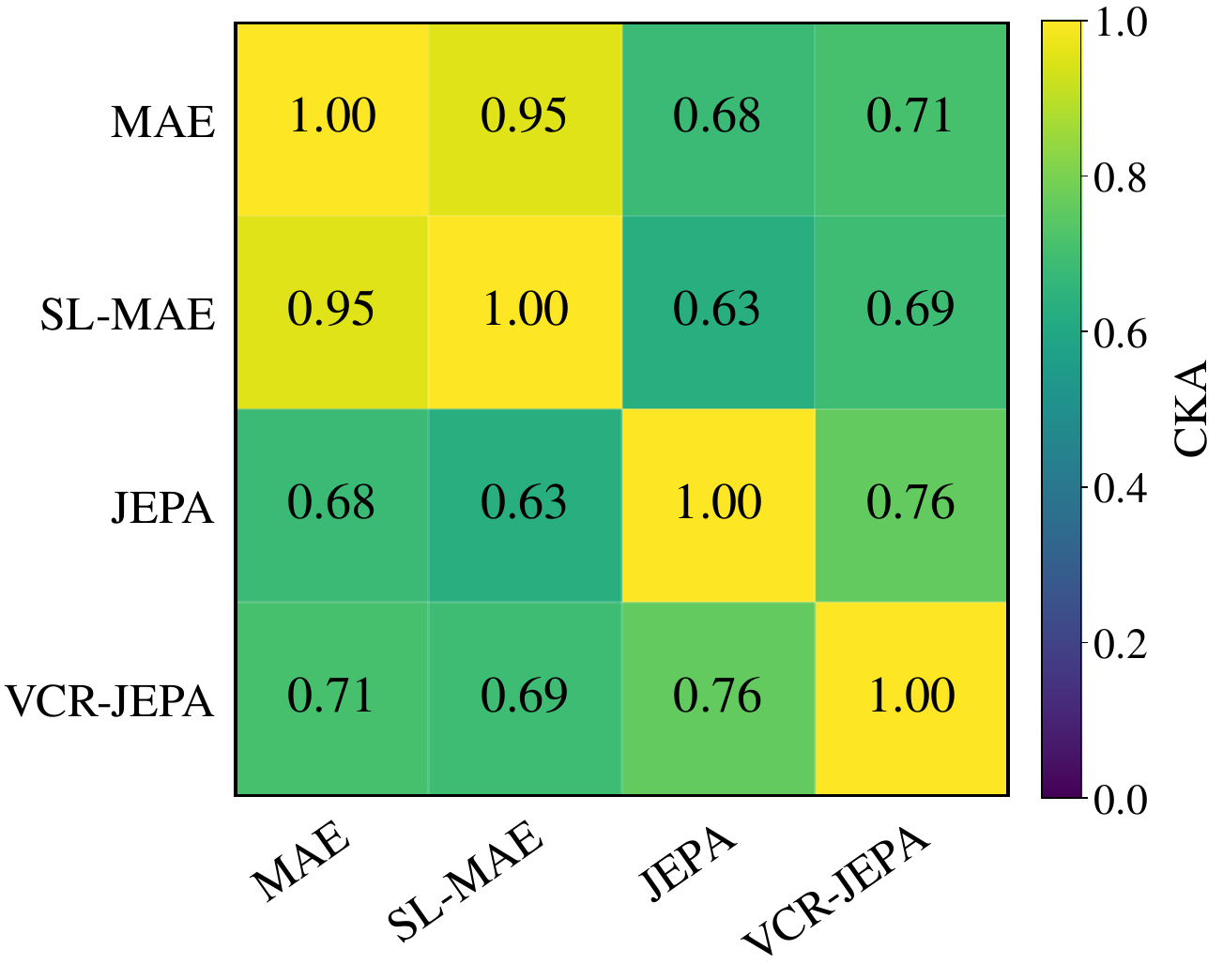} &
        \includegraphics[width=0.23\textwidth]{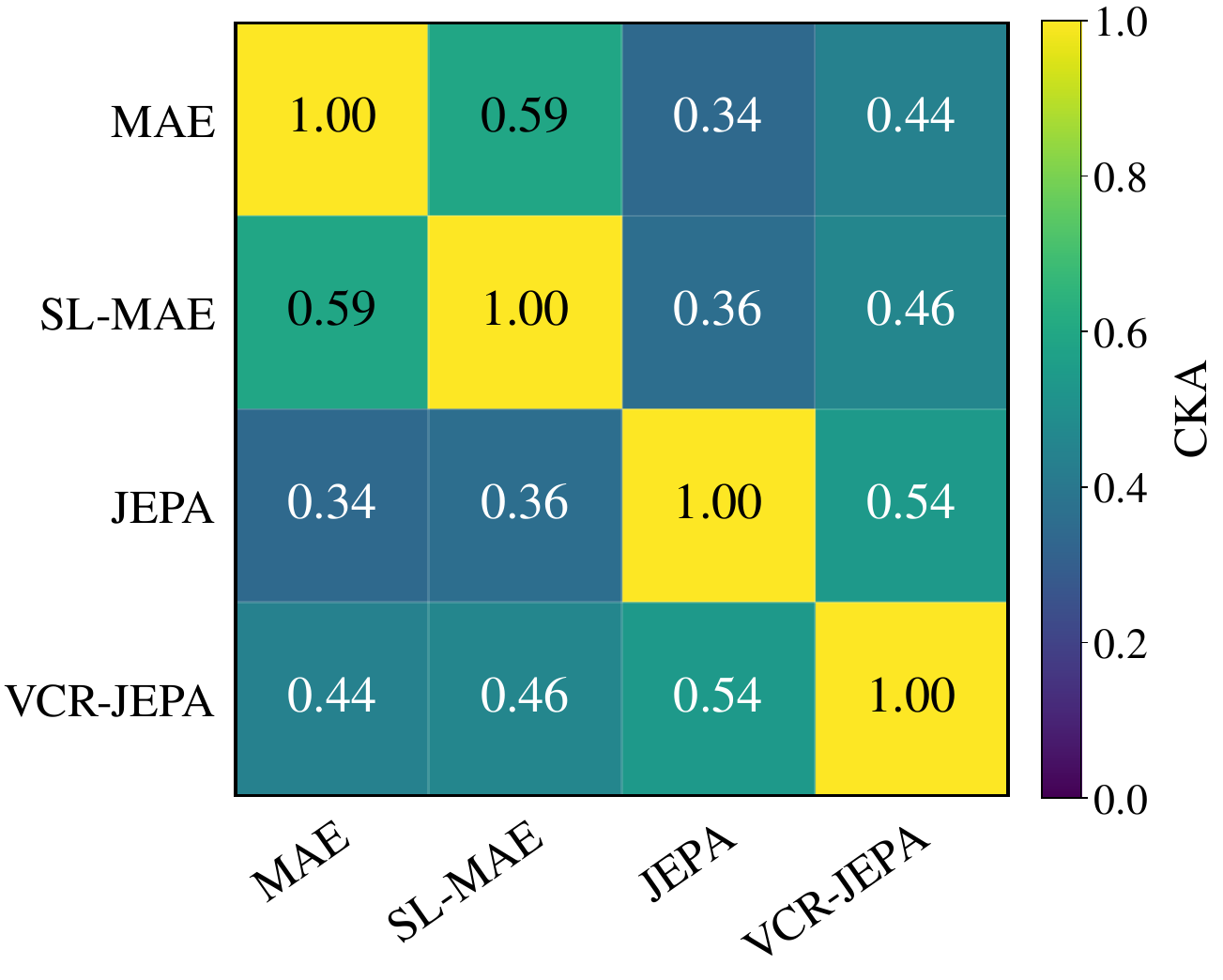}
        \\[3pt]

        (a) Pretrained, NC vs.\ AD. &
        (b) Fine-tuned, NC vs.\ AD. &
        (c) Pretrained, NC vs.\ MCI. &
        (d) Fine-tuned, NC vs.\ MCI.
    \end{tabular}

    \caption{\textbf{Pairwise CKA similarity for ADNI NC vs.\ AD and ADNI NC vs.\ MCI.}
    Pairwise similarity among MAE, SL-MAE, JEPA, and VCR-JEPA
    representations before and after fine-tuning, computed using the linear
    kernel. Within-family similarity is generally higher than cross-family
    similarity, while fine-tuning reduces representational agreement between
    objectives. Corresponding RBF-kernel results are provided in
    Supplementary Fig.~S3.}
    \label{fig:cka-pooled-ad}
\end{figure}


Figure~\ref{fig:cka-pooled-ad} reports pairwise CKA similarities computed
from the final-layer representations of the four pretrained and fine-tuned
encoders using the linear kernel. The corresponding RBF-kernel results are
provided in Supplementary Fig.~S3. At the pretrained stage, within-family similarity (MAE vs.\ SL-MAE; JEPA vs.\ VCR-JEPA) consistently exceeds cross-family similarity, indicating that reconstruction-based and latent-predictive objectives establish systematically different representational geometries, while their corresponding auxiliary objectives preserve the broader geometry of the underlying SSL paradigm. This organization is observed on both the NC vs.\ AD and NC vs.\ MCI evaluation cohorts, supporting its interpretation as an objective-dependent property of the pretrained representations rather than a cohort-specific effect. Following fine-tuning, pairwise representational similarities generally decrease, particularly for NC vs.\ AD, while the within-family versus cross-family organization remains evident. Thus, downstream supervision substantially reshapes the learned representations but does not erase objective-dependent geometric differences. Moreover, the greater divergence among fine-tuned encoders for NC vs.\ AD than for NC vs.\ MCI provides further evidence that the geometry reached through downstream adaptation depends on the target task as well as on the pretraining objective.

\begin{figure}[pos=H]
    \centering
    \begin{subfigure}[b]{0.48\textwidth}
        \centering
        \includegraphics[width=\textwidth]{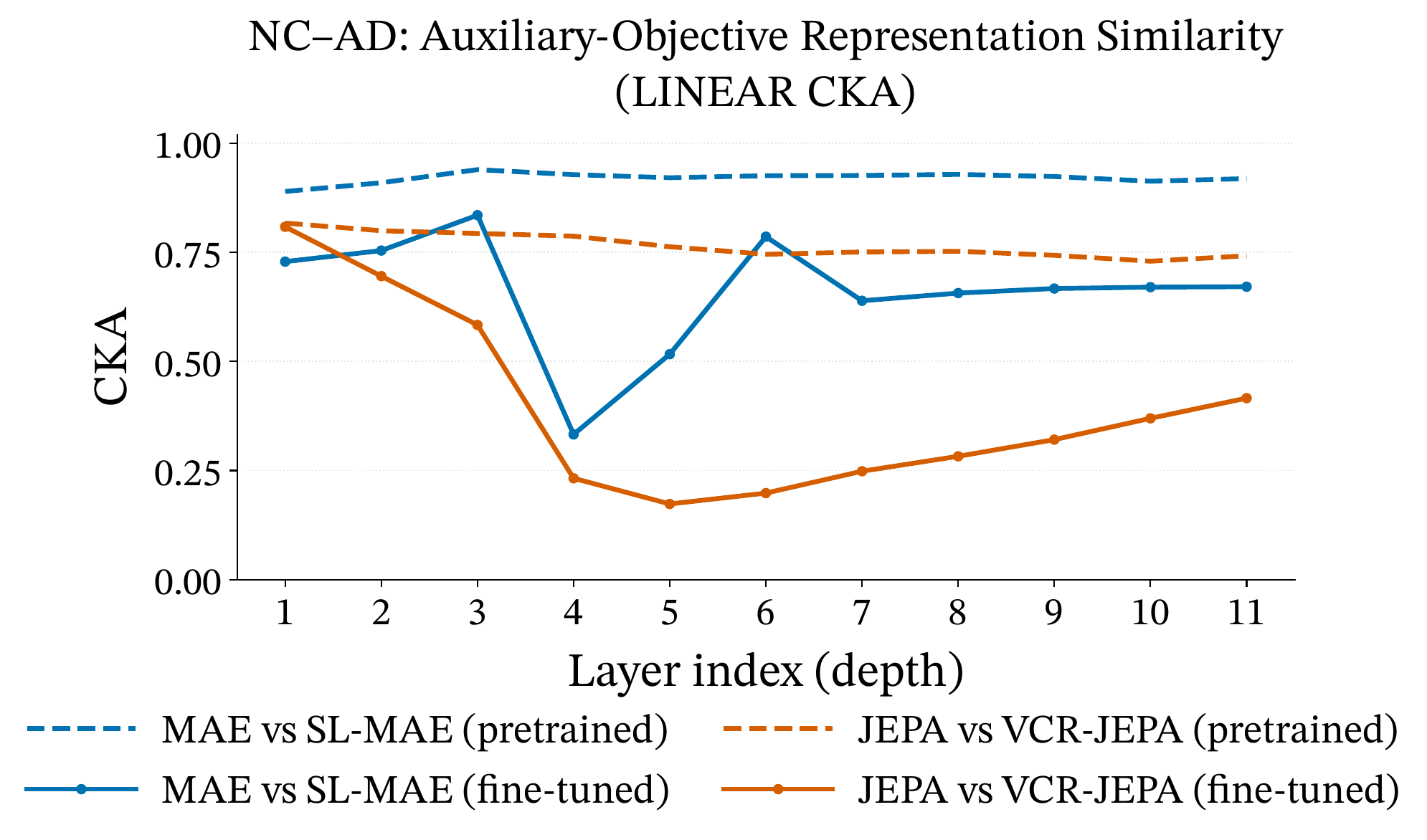}
        \caption{Linear CKA (AD)}
        \label{fig:depth-profile-aux-linear}
    \end{subfigure}
    \hfill
    \begin{subfigure}[b]{0.48\textwidth}
        \centering
        \includegraphics[width=\textwidth]{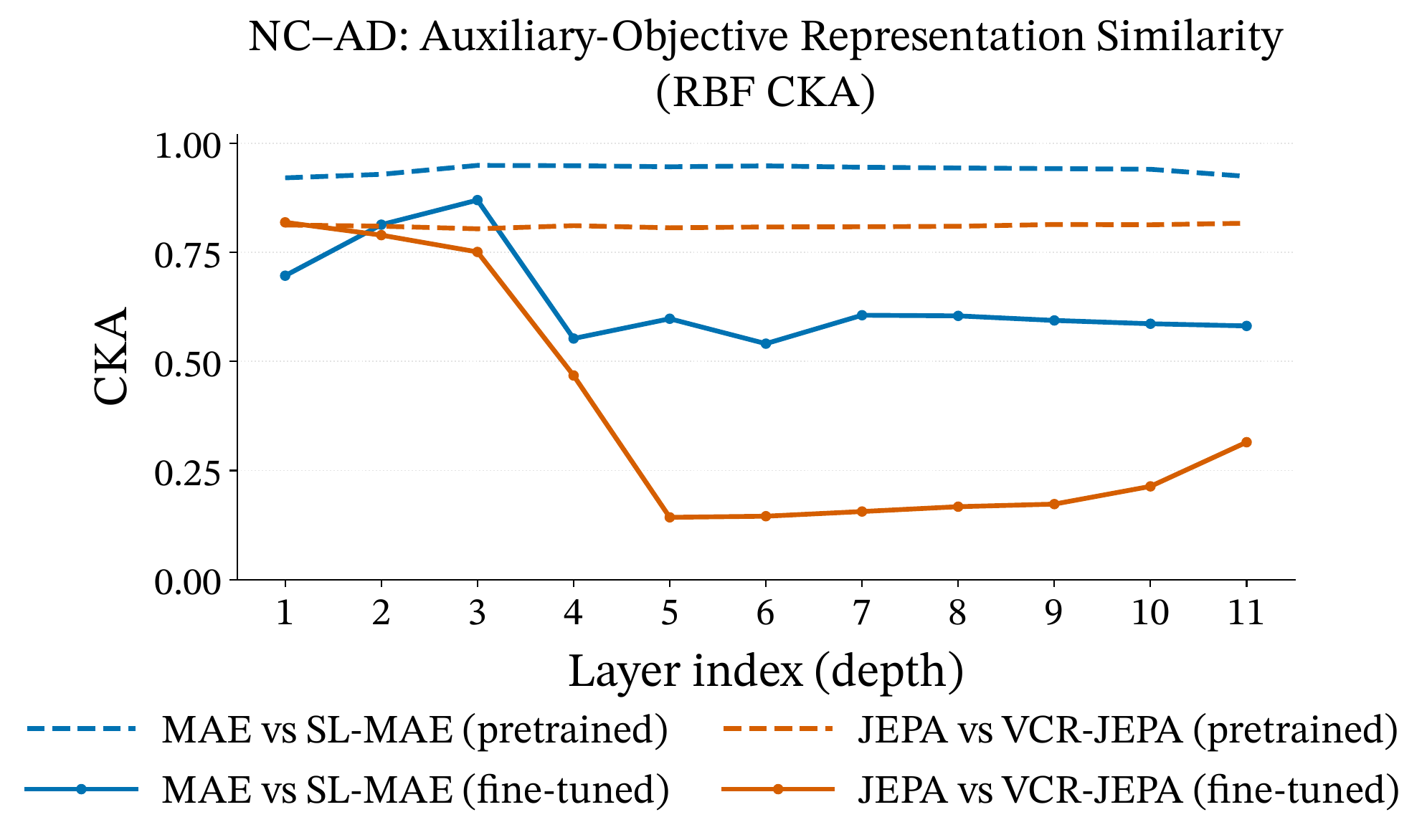}
        \caption{RBF CKA (AD)}
        \label{fig:depth-profile-aux-rbf}
    \end{subfigure}

    \vspace{0.6em}

    \begin{subfigure}[b]{0.48\textwidth}
        \centering
        \includegraphics[width=\textwidth]{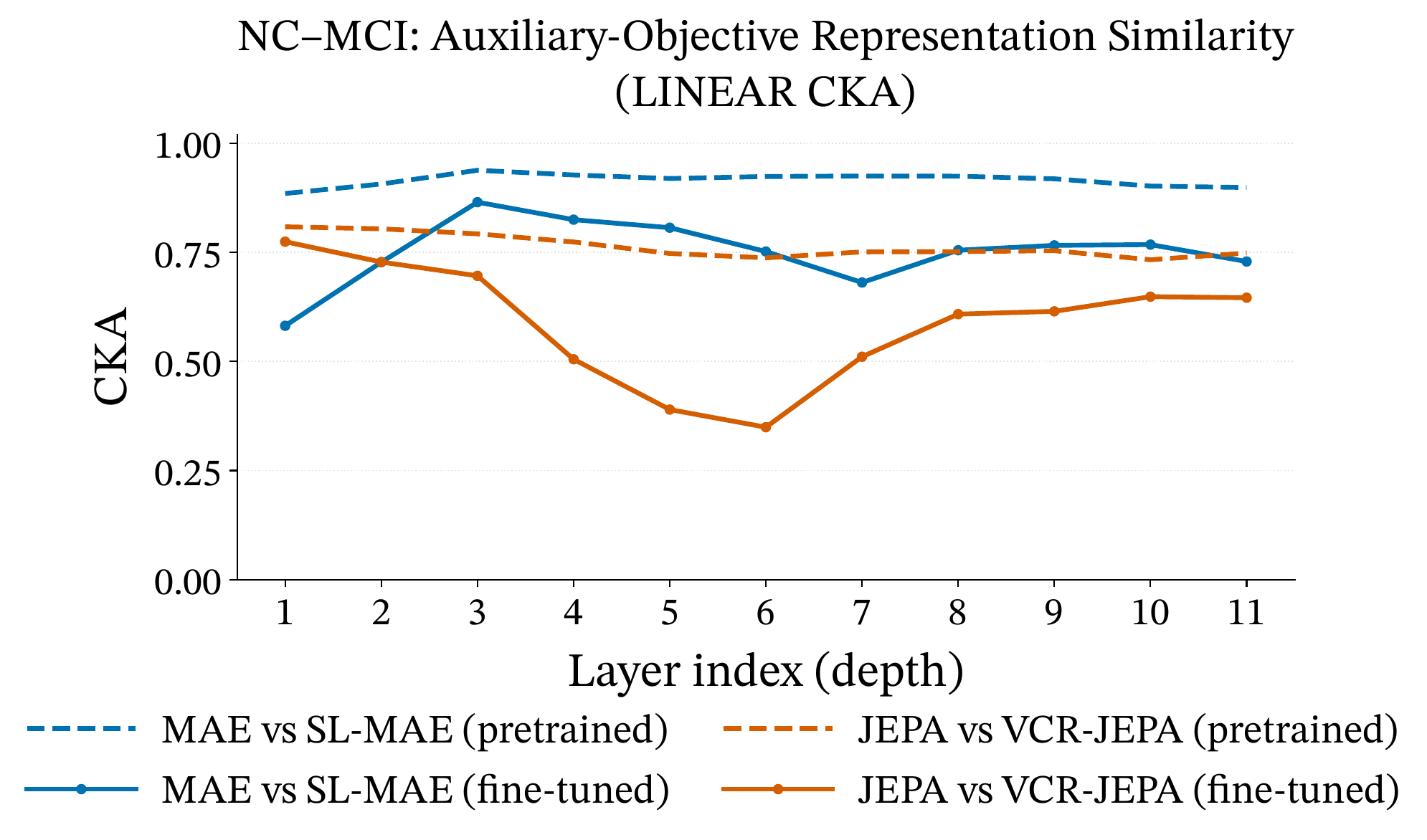}
        \caption{Linear CKA (MCI)}
        \label{fig:depth-profile-aux-linear-mci}
    \end{subfigure}
    \hfill
    \begin{subfigure}[b]{0.48\textwidth}
        \centering
        \includegraphics[width=\textwidth]{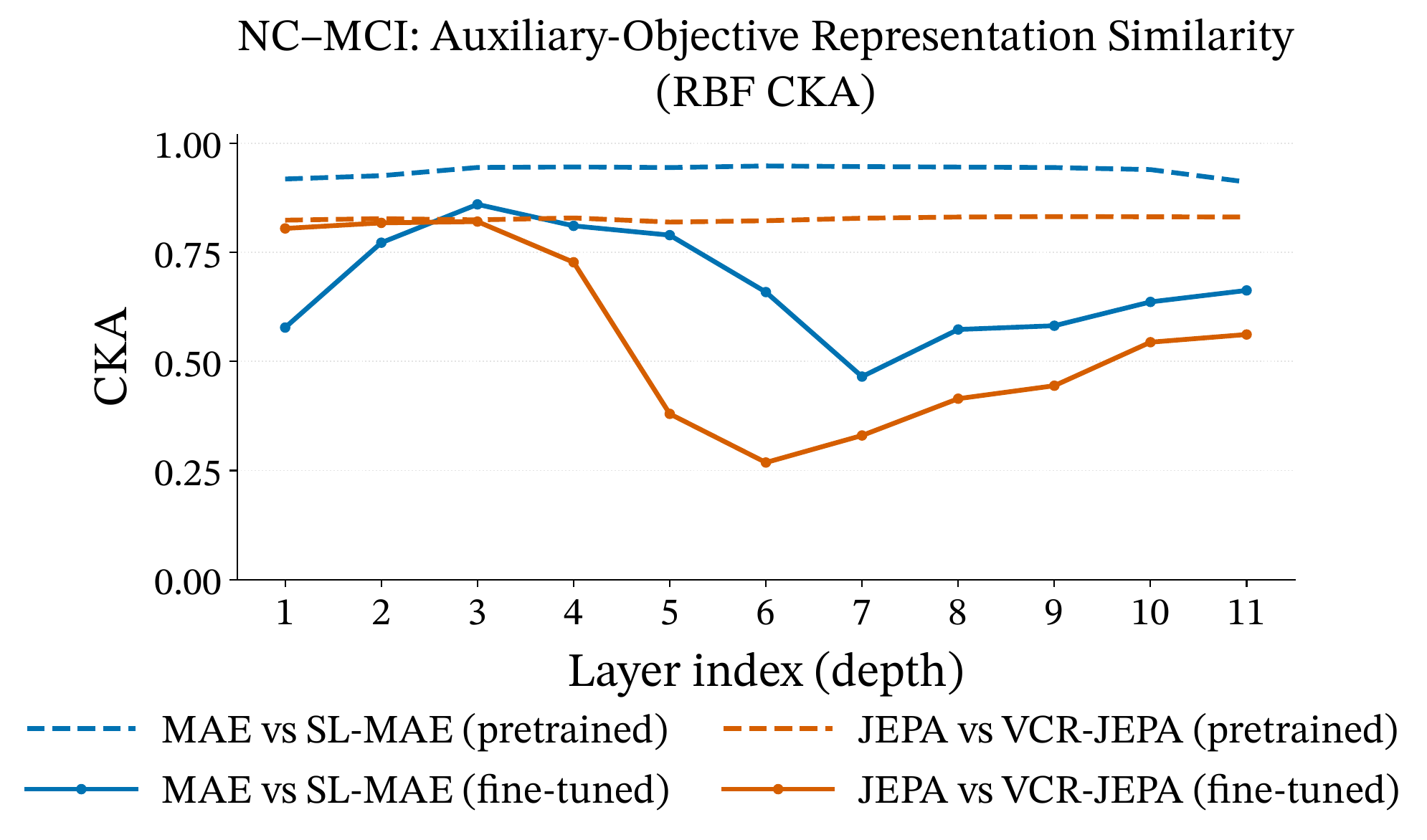}
        \caption{RBF CKA (MCI)}
        \label{fig:depth-profile-aux-rbf-mci}
    \end{subfigure}

    \caption{Layer-wise CKA between baseline and auxiliary-objective encoders (MAE vs.\ MAE+SL; JEPA vs.\ JEPA-VCR) before and after fine-tuning, using linear (left) and RBF (right) kernels on ADNI NC-vs-AD (top) and NC-vs-MCI (bottom). Pretrained pairs remain similar across depth, while fine-tuned pairs show a mid-layer similarity drop followed by partial or full recovery.
    }
    \label{fig:depth-profile-auxiliary}
\end{figure}

Figure~\ref{fig:depth-profile-auxiliary} isolates the geometric footprint of each auxiliary objective by comparing it with its corresponding baseline. During pretraining, both baseline–auxiliary pairs remain highly and relatively uniformly aligned across depth, indicating that the auxiliary objectives modestly reshape rather than replace the geometry established by the underlying SSL paradigm. SL-MAE remains particularly close to MAE, whereas VCR-JEPA exhibits a larger but still stable deviation from JEPA. Following fine-tuning, these initially modest differences become strongly depth-dependent, with pronounced intermediate-layer divergence followed by partial recovery toward later layers. The magnitude and trajectory of this divergence also differ between NC vs.\ AD and NC vs.\ MCI, indicating that auxiliary objectives modify how pretrained representations respond to task-specific supervision rather than imposing a uniform representational shift.

\begin{figure}[pos=H]
    \centering

    \begin{subfigure}[b]{0.49\textwidth}
        \centering
        \includegraphics[width=\textwidth]{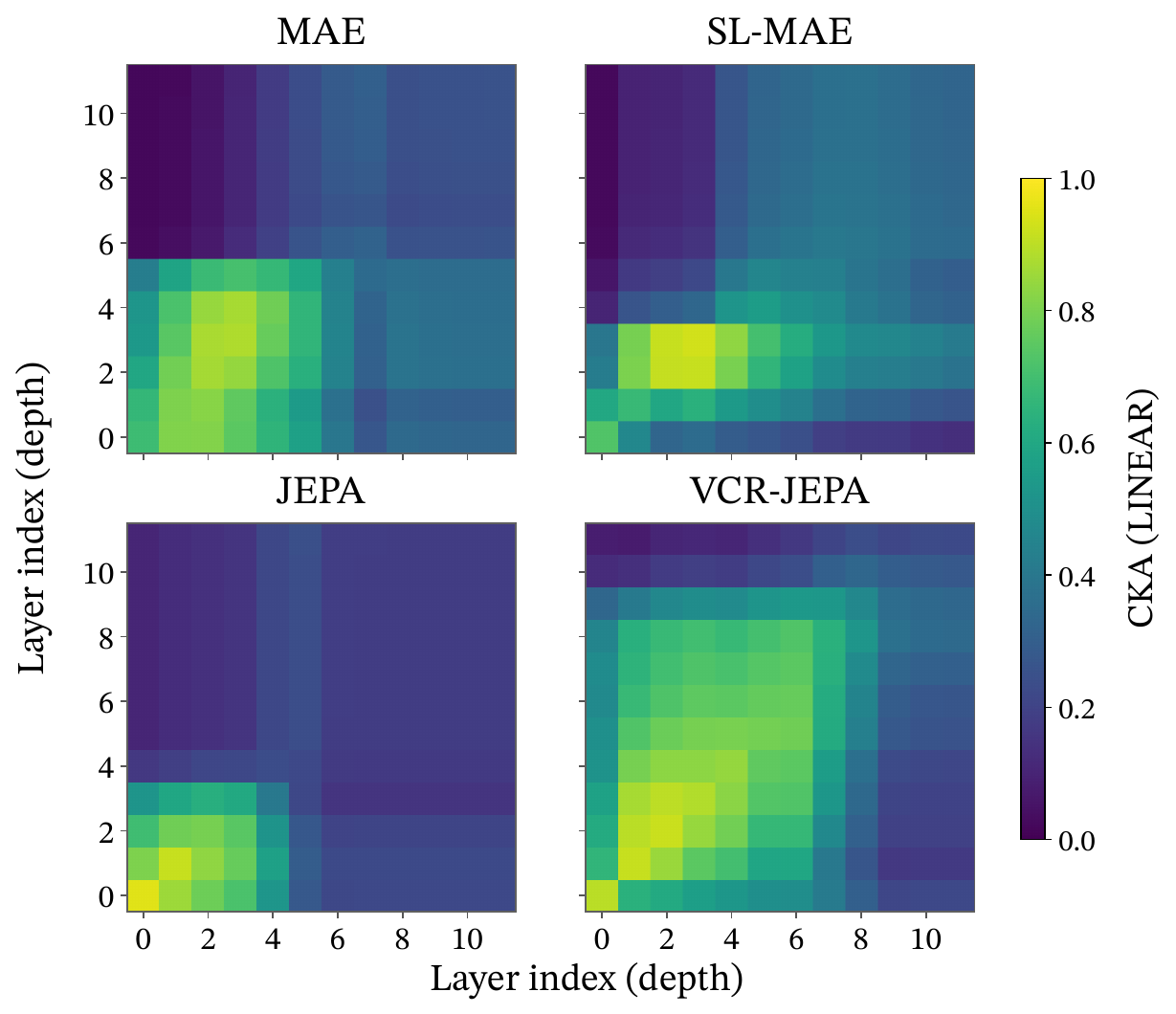}
        \caption{ADNI: NC vs.\ AD and NC vs.\ MCI}
        \label{fig:crosstask-adni}
    \end{subfigure}
    \hfill
    \begin{subfigure}[b]{0.49\textwidth}
        \centering
        \includegraphics[width=\textwidth]{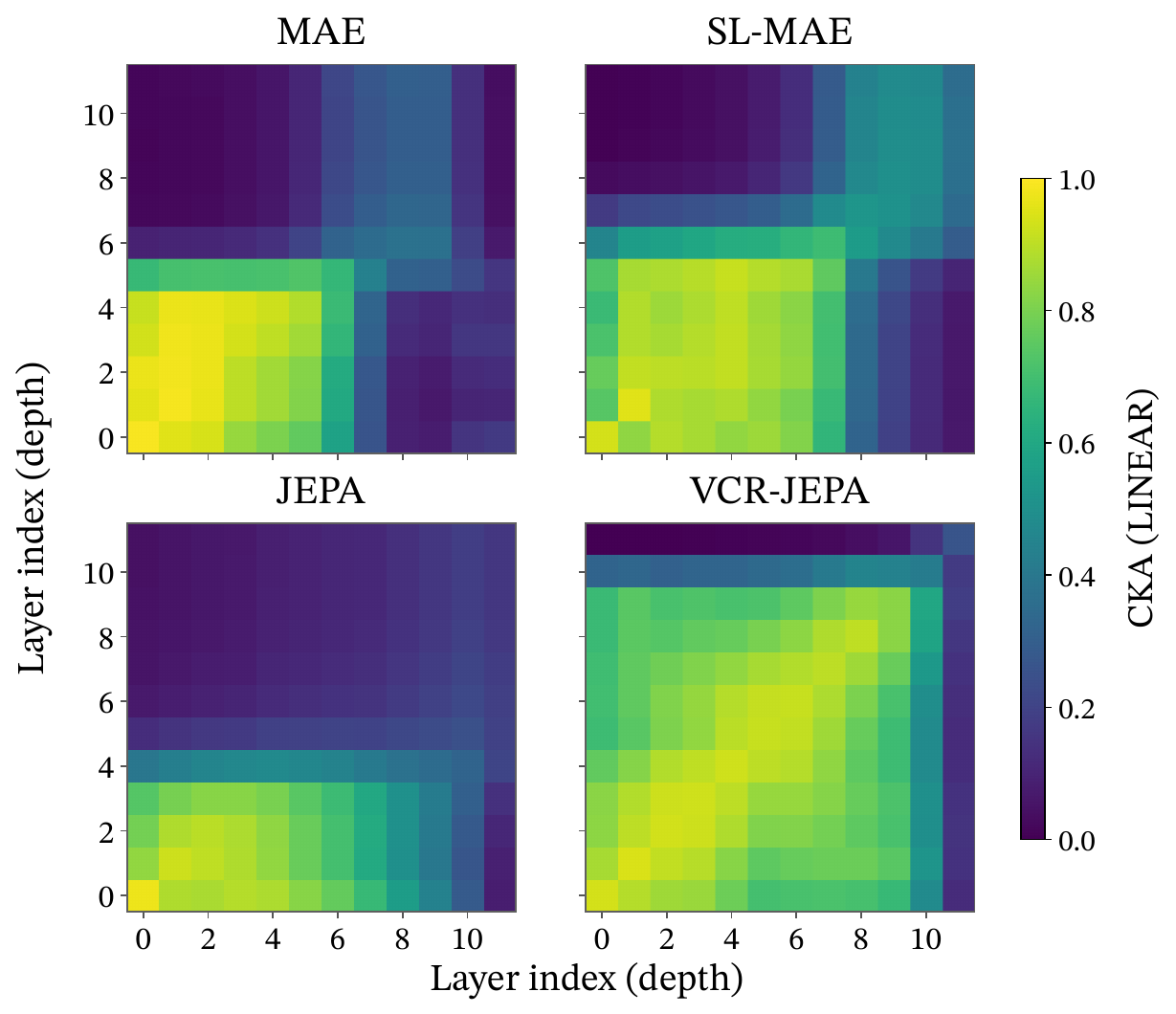}
        \caption{UCSF: binary and multiclass tumor grading}
        \label{fig:crosstask-ucsf}
    \end{subfigure}

    \caption{\textbf{Cross-task representational similarity.}
Layer-wise linear CKA between encoders fine-tuned from the same pretrained model on related tasks.
(a) ADNI: NC vs.\ AD (rows) and NC vs.\ MCI (columns).
(b) UCSF: binary (rows) and multiclass (columns) tumor grading.
Both checkpoints are evaluated on the same held-out scans in identical order.}
    \label{fig:crosstask}
\end{figure}

Finally, we examine how much representation geometry is shared when the same pretrained encoder is adapted to two related downstream tasks. Figure~\ref{fig:crosstask} presents this analysis for two independent task pairs: NC vs.\ AD and NC vs.\ MCI on ADNI (Fig.~\ref{fig:crosstask-adni}), and binary versus multiclass tumor grading on UCSF (Fig.~\ref{fig:crosstask-ucsf}). In both settings, VCR-JEPA exhibits the broadest cross-task alignment across early and intermediate layers, whereas baseline JEPA shows substantially more localized similarity. MAE and SL-MAE exhibit intermediate patterns of shared geometry. The recurrence of this pattern across two distinct downstream task pairs suggests that variance--covariance-regularized pretraining is associated with greater preservation of shared representation geometry during task-specific adaptation, while still permitting divergence in later layers. We interpret this greater cross-task CKA as evidence of representational reuse across related tasks rather than as a direct measure of downstream predictive performance.

The CKA analyses reveal complementary effects of objective design and downstream adaptation on representation geometry. The underlying SSL paradigm establishes characteristic objective-family geometries, while the auxiliary objectives introduce comparatively modest but measurable modifications during pretraining. Fine-tuning subsequently increases depth differentiation and reorganizes these representations in a manner that depends jointly on encoder depth, pretraining objective, and downstream task, with substantially sharper reorganization for NC vs.\ AD than for NC vs.\ MCI. Finally, cross-task CKA indicates that the extent to which a shared geometry is retained across related downstream adaptations is also objective-dependent, with VCR-JEPA exhibiting the broadest cross-task alignment in both the ADNI and UCSF comparisons.

\subsection{Shannon Effective Rank Analysis}

In addition to CKA, we characterize the dimensional structure of the learned
representations using Shannon effective rank following RankMe
\cite{garrido2023rankme}. Effective rank summarizes how broadly the
singular-value spectrum is distributed across latent directions: higher values
indicate that spectral mass is distributed across a larger number of effective
dimensions, whereas lower values indicate stronger concentration in fewer
dominant directions. Because effective rank is estimated from finite held-out
cohorts, we interpret its absolute magnitude cautiously and focus primarily on
relative model-wise and depth-wise patterns under matched evaluation conditions.

For a centered representation matrix $Z \in \mathbb{R}^{n\times d}$ with
singular values $\{\sigma_i\}$, we normalize the singular-value spectrum as
\begin{equation}
p_i = \frac{\sigma_i}{\sum_j \sigma_j},
\end{equation}
and compute the Shannon effective rank as
\begin{equation}
R_{\mathrm{eff}}
=
\exp\left(-\sum_i p_i \log p_i\right).
\end{equation}


\begin{figure}[pos=H]
    \centering

    \begin{subfigure}[t]{0.48\textwidth}
        \centering
        \includegraphics[width=\linewidth]{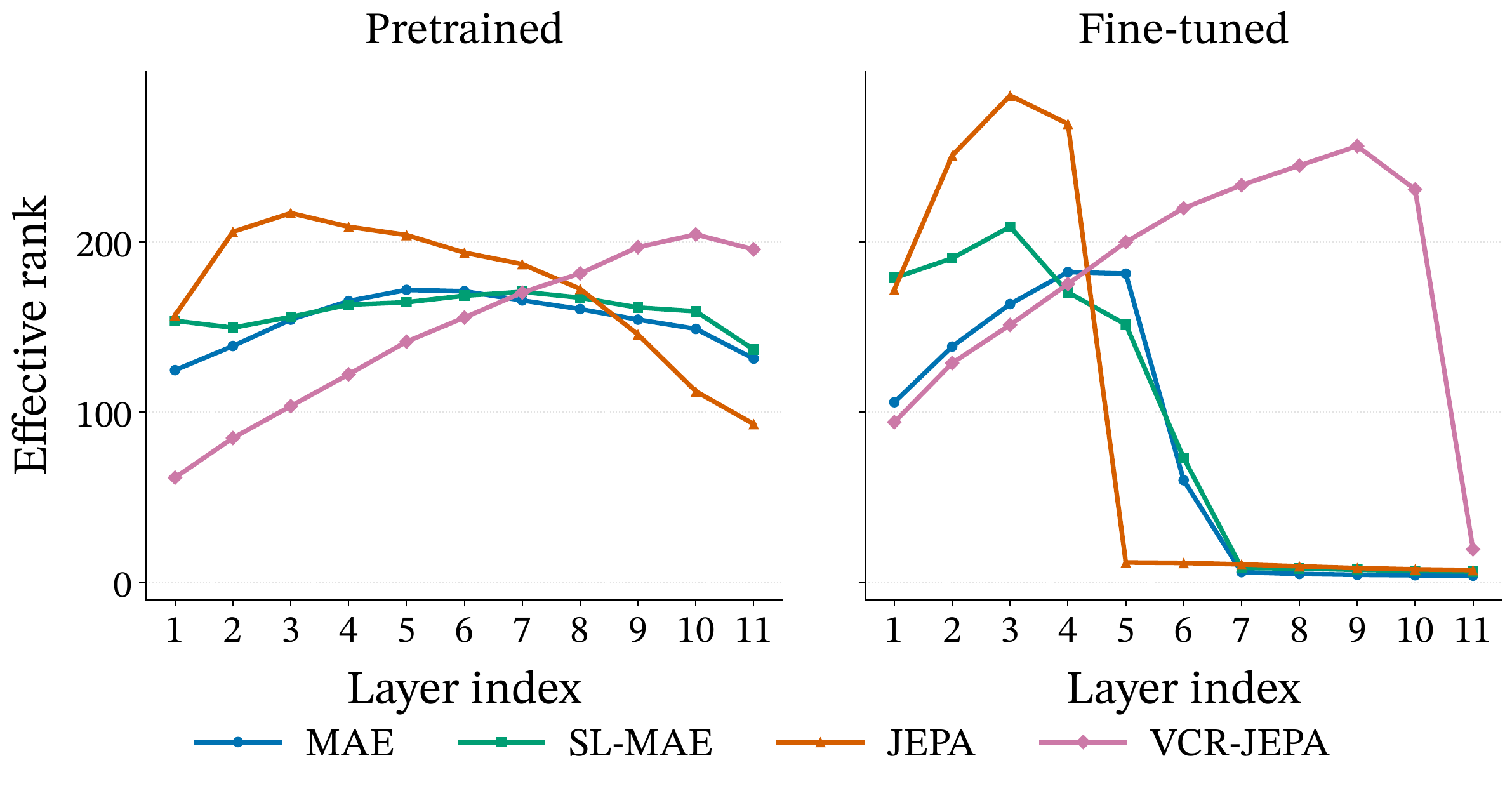}
        \caption{ADNI NC vs.\ AD}
        \label{fig:erank-layerwise-ad}
    \end{subfigure}
    \hfill
    \begin{subfigure}[t]{0.48\textwidth}
        \centering
        \includegraphics[width=\linewidth]{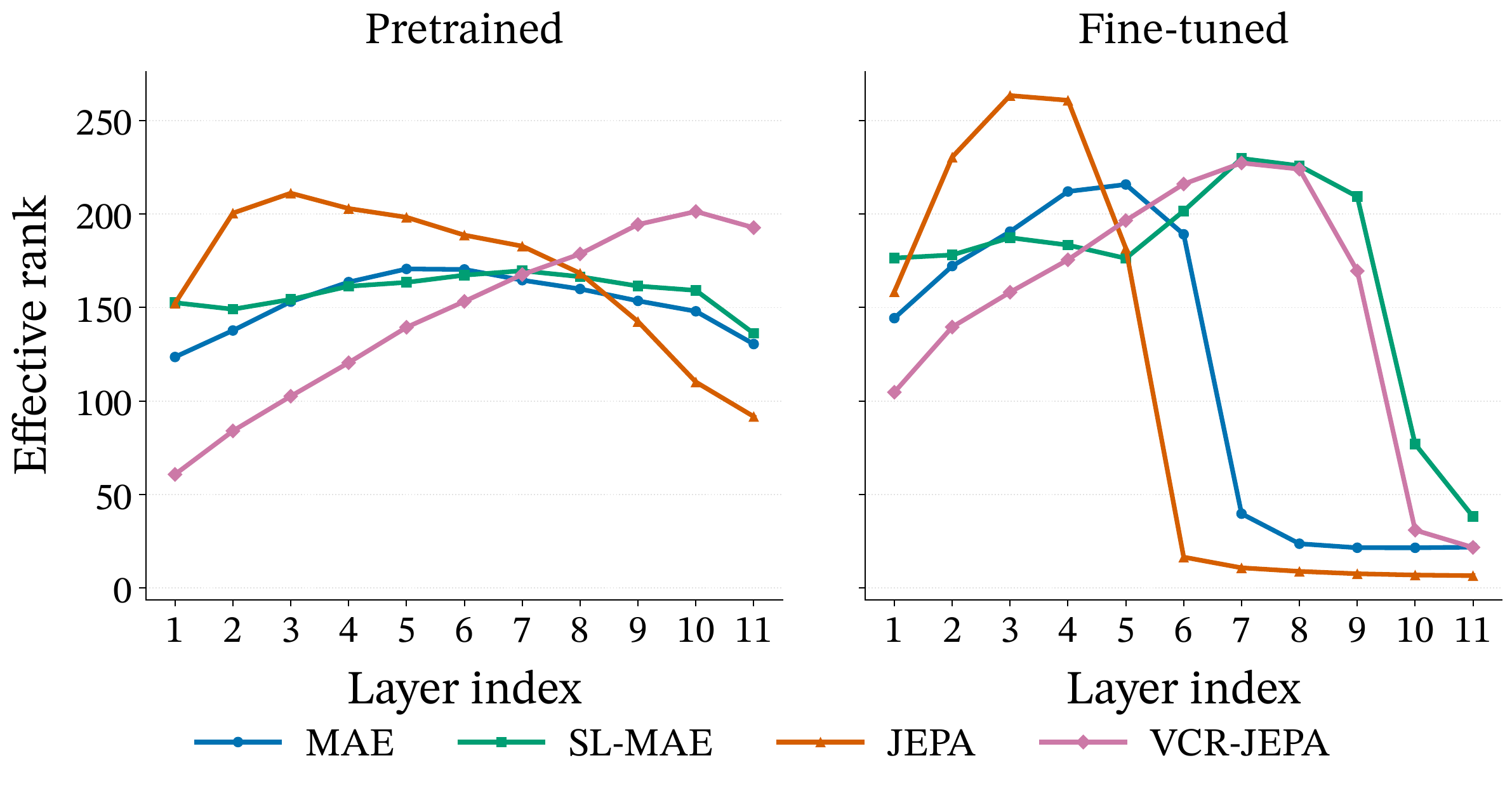}
        \caption{ADNI NC vs.\ MCI}
        \label{fig:erank-layerwise-mci}
    \end{subfigure}

\caption{
Layer-wise Shannon effective rank for ADNI NC-vs-AD (left) and NC-vs-MCI (right), showing pretrained and fine-tuned encoders. Pretraining yields consistent objective-dependent rank profiles, while fine-tuning introduces task-dependent changes.
}
    \label{fig:erank-layerwise}
\end{figure}

Figure~\ref{fig:erank-layerwise} shows the layer-wise effective-rank profiles of the four pretrained and fine-tuned encoders on NC vs.\ AD and NC vs.\ MCI. At the pretrained stage, the objective-dependent profiles are broadly consistent across the two evaluation cohorts. VCR-JEPA maintains the broadest effective-rank profile through most of the encoder depth, whereas plain JEPA reaches its highest rank earlier and subsequently decreases. MAE and SL-MAE exhibit more similar profiles, indicating a comparatively modest effect of spectral supervision on the dimensional structure established by MAE pretraining. These patterns complement the self-CKA results in Figure~\ref{fig:selfcka}: while pretrained encoders exhibit relatively limited differentiation in their layer-wise similarity structure, their effective dimensionality nevertheless evolves differently with depth depending on the pretraining objective.

Following fine-tuning, the effective-rank trajectories become markedly task-dependent. Under NC vs. AD fine-tuning, MAE, SL-MAE, and JEPA show substantial concentration of representation variance into fewer effective directions at intermediate or later depths, whereas VCR-JEPA retains a broader effective rank through most of the encoder. Under NC vs. MCI fine-tuning, rank reduction is more gradual, and SL-MAE also retains comparatively broad effective dimensionality into later layers. Thus, the influence of the auxiliary objectives on representational dimensionality is itself task-dependent: VCR-JEPA shows the clearest rank preservation under NC vs. AD adaptation, whereas spectral supervision also delays rank concentration under NC vs. MCI. 

The consistently broader rank profile of VCR-JEPA is compatible with the intended effect of variance--covariance regularization, which encourages variance across embedding dimensions while reducing feature redundancy. However, effective rank is not itself a direct measure of task-relevant information. Indeed, the ordering of effective-rank values in Figure~\ref{fig:erank-layerwise} does not match the ordering of downstream AUC scores reported in Sections 6.1 and 6.2. Higher effective dimensionality therefore should not be interpreted as inherently superior discriminability; rather, it characterizes how broadly representation variance is distributed across latent directions.

Together with the CKA analysis, these results provide complementary views of downstream adaptation. CKA shows that fine-tuning reorganizes representation geometry across depth, whereas effective rank shows that it also redistributes representation variance across latent directions. Both effects depend on the pretraining objective and the downstream task, indicating that task-specific adaptation changes not only the arrangement of learned representations but also their effective dimensional structure.

\subsection{Few-Shot Transfer Learning}

Self-supervised representation learning is particularly attractive in clinical imaging, where labeled data are often limited. To assess transfer under restricted supervision, we compare the auxiliary-objective variants SL-MAE and VCR-JEPA with a supervised ResNet50-3D trained from scratch using 16, 32, and 64 labeled samples per class.

We evaluate NC vs.\ AD and NC vs.\ MCI classification on ADNI using subsets that remain disjoint from the subjects used during self-supervised pretraining. For the few-shot experiments, $k\in\{16,32,64\}$ denotes the number of training subjects sampled per class, while the validation and held-out test sets are kept fixed across $k$. The SSL models are fine-tuned following the protocol described in Section~5.3, whereas ResNet50-3D is trained from scratch for 180 epochs.

\begin{figure}[pos=H]
    \centering
    \begin{subfigure}{.48\textwidth}
        \includegraphics[width=\linewidth]{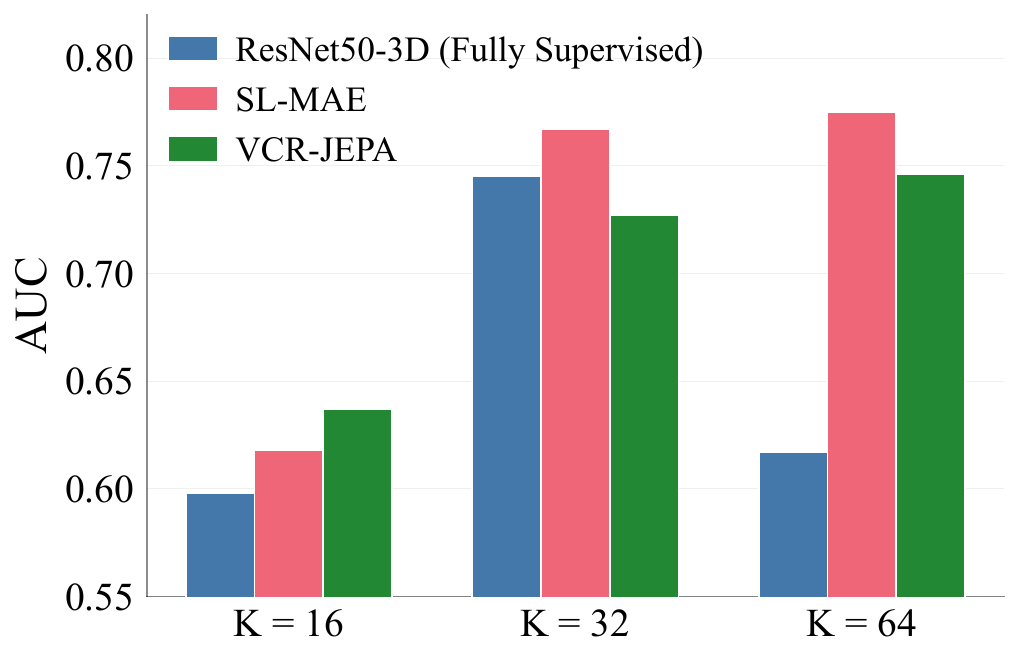}
        \caption{NC vs. AD}
        \label{fig:auc_ad}
    \end{subfigure}
    \hfill
    \begin{subfigure}{.48\textwidth}
        \includegraphics[width=\linewidth]{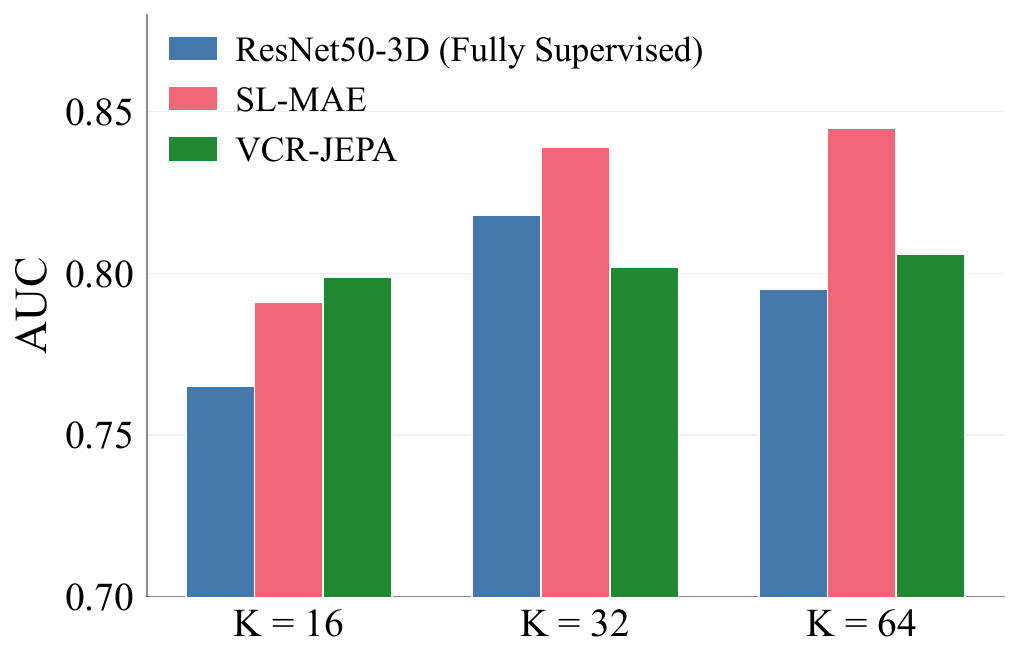}
        \caption{NC vs. MCI}
        \label{fig:auc_mci}
    \end{subfigure}
    \caption{$k$-shot classification performance (AUC) on ADNI. SL-MAE outperforms the supervised ResNet50-3D baseline across all \(k\) values for both NC vs.\ AD and NC vs.\ MCI, while VCR-JEPA outperforms the baseline at \(k=16\) and \(k=64\) and remains competitive at \(k=32\).}
    \label{fig:auc_comparison}
\end{figure}

Figure~\ref{fig:auc_comparison} summarizes few-shot AUC performance. SL-MAE outperforms the supervised ResNet50-3D baseline across all three labeled-data regimes for both NC vs.\ AD and NC vs.\ MCI. VCR-JEPA also exceeds the supervised baseline at \(k=16\) and \(k=64\) for both tasks and remains competitive at \(k=32\). These results indicate that the pretrained representations remain effective under limited downstream supervision. In particular, the strong few-shot performance of SL-MAE and VCR-JEPA suggests that representations shaped by spectral and variance--covariance regularization can support effective adaptation in label-constrained clinical imaging settings.

\section{Discussion}

In this work, we investigated how different self-supervised pretraining formulations generalize across multiple disease-detection tasks using 3D brain MRI. Specifically, we investigated two major paradigms for self-supervised learning in 3D brain MRI modeling: reconstruction-based learning through Masked Autoencoders (MAE) and predictive representation learning through Joint Embedding Predictive Architectures (JEPA). In addition to comparing these frameworks directly, we introduced and evaluated two auxiliary objectives tailored to each paradigm: a spectral-domain reconstruction loss for MAE and variance-covariance regularization (VCR) for JEPA. Across our controlled comparisons, MAE-based approaches generally achieved stronger downstream performance than JEPA-based approaches, particularly in disease-detection tasks involving subtle anatomical differences.

The central finding of this study is that neither spectral regularization nor variance--covariance regularization provides uniform benefit across downstream tasks. Instead, their effectiveness varies with the characteristics of the discriminative signal. Our results are consistent with spectral supervision being most useful when clinically relevant information includes pronounced high-frequency anatomical structure, and with VCR being most useful when discriminative information is distributed across multiple complementary feature dimensions. Tasks in which these characteristics are weaker exhibit correspondingly smaller gains. These findings suggest that auxiliary SSL objectives should not be viewed as universally beneficial design choices, but rather as mechanisms that impose distinct representational biases whose utility depends on the downstream task.

While much prior MRI SSL work has emphasized segmentation and dense prediction, our study focuses on disease-state classification across four clinical problem domains including Alzheimer's disease, mild cognitive impairment, glioma grading, and autism spectrum disorder, evaluated across seven dataset--task settings. These problems span focal, diffuse, and heterogeneous structural alterations and therefore provide complementary tests of representation transfer beyond segmentation-specific supervision. Our results show that self-supervised pretraining can produce transferable representations across these clinically distinct disease-detection settings, while also revealing substantial variation in the benefit of different pretraining objectives.

A common intuition in SSL is that reconstruction-based methods are particularly well suited to dense prediction because they preserve local image content, whereas latent-predictive and invariance-based methods may be better aligned with semantic classification. Our findings suggest that this distinction is too simple for structural MRI. Although MAE reconstructs masked patches in voxel space, successful recovery of missing 3D anatomy requires integration of structural context across the volume. Moreover, because the MAE decoder in our framework is intentionally lightweight, the encoder must retain substantial anatomical information that can subsequently support image-level disease classification. The strong downstream performance of MAE-based models therefore indicates that voxel-level reconstruction can yield highly transferable representations for structural MRI, rather than being useful primarily for dense prediction.

Our experiments further suggest that spectral-domain regularization is
particularly beneficial for downstream tasks in which discriminative information
is associated with pronounced high-frequency spatial structure. The largest gains
occur for Alzheimer's disease classification and tumor grading, where fine
anatomical boundaries, structural heterogeneity, and textural variation contribute
to the imaging phenotype. In contrast, gains are substantially smaller for NC
vs.\ MCI and ASD, where discriminative structural differences are subtler and
less strongly expressed in high-frequency spatial content. Together with the
frequency-domain analysis in Section~\ref{subsec:MAE-SL}, these results are
consistent with spectral supervision preferentially emphasizing fine-grained
structural information that becomes useful when such information aligns with the
downstream discrimination problem. This interpretation is also consistent with
prior MRI texture- and frequency-domain studies reporting more pronounced
structural and textural abnormalities in established Alzheimer's disease than
in MCI.

Similarly, the JEPA experiments suggest that VCR is most beneficial when
downstream discrimination can draw on multiple complementary sources of
structural variation. VCR-JEPA shows its clearest improvements over baseline
JEPA for AD classification and tumor grading, tasks in which abnormalities
involve multiple anatomical regions or heterogeneous pathological patterns,
whereas its benefit is smaller or absent in several MCI and ASD settings.
This pattern is consistent with the intended effect of variance--covariance
regularization: maintaining variance while reducing redundancy across latent
dimensions may be particularly useful when discriminative information is
distributed across several partially independent features.

These results suggest that SL and VCR introduce distinct representational biases rather than uniformly improving representation quality: spectral supervision favors fine-grained, frequency-sensitive structural information, whereas VCR favors broader and less redundant latent representations. Their downstream utility
therefore depends on how well these biases align with the statistical structure
of the target task.

The few-shot experiments provide complementary evidence that the learned
representations remain useful under limited downstream supervision. SL-MAE
outperforms the supervised ResNet50-3D baseline across all evaluated label
regimes, while VCR-JEPA also remains competitive or superior in most settings.
The strong few-shot performance of the two auxiliary-objective variants suggests
that representations shaped by spectral and variance--covariance regularization
can support effective adaptation in label-constrained clinical imaging settings,
where annotated data are often scarce.

The comparison between frozen and fine-tuned evaluation further shows that downstream adaptation is not uniformly beneficial. On several challenging or distribution-shifted tasks, including NC vs.\ MCI and ABIDE, frozen MAE representations match or exceed their fine-tuned counterparts. This suggests that when downstream supervision is limited or the disease signal is subtle and heterogeneous, preserving the broader anatomical representation learned during pretraining can be competitive with full task-specific adaptation. Conversely, stronger gains from fine-tuning are observed when the downstream signal provides sufficient task-specific structure to support effective specialization. The representation analyses below provide a geometric counterpart to this observation, showing that the extent of fine-tuning-induced reorganization itself varies substantially across tasks.

Layer-wise CKA between each pretrained encoder and its fine-tuned counterpart shows that downstream adaptation progressively alters pretrained representation geometry, with changes generally becoming more pronounced toward deeper layers. The trajectory of this change depends jointly on the pretraining objective and downstream task: objectives exhibit distinct adaptation profiles across depth, while the target task modulates both the magnitude and progression of representational divergence. Complementarily, self-CKA shows substantially greater similarity across layers in pretrained than fine-tuned encoders, indicating lower depth differentiation before downstream adaptation. Fine-tuning therefore modifies not only individual layer representations relative to pretraining but also the encoder's internal organization across depth.

The auxiliary-objective CKA analysis further shows that these changes are not distributed uniformly across depth. Baseline and auxiliary-objective variants remain relatively similar during pretraining, whereas fine-tuning amplifies their differences at intermediate layers before partial recovery toward later depths. VCR-JEPA additionally exhibits a smoother internal transition across depth than baseline JEPA, indicating a more gradual pattern of fine-tuned depth differentiation. At the final layer, pairwise similarities generally decrease after fine-tuning, while within-family similarity remains higher than cross-family similarity. Thus, downstream supervision substantially reshapes the representations but does not erase geometric differences associated with the pretraining objective.

Cross-task CKA provides a complementary view of representational reuse during downstream adaptation. Across both related task pairs: NC vs.\ AD versus NC vs.\ MCI on ADNI, and binary versus multiclass tumor grading on UCSF, VCR-JEPA exhibits the broadest cross-task alignment across early and intermediate layers. This recurring pattern suggests that variance--covariance-regularized pretraining is associated with greater preservation of common representation structure when the same pretrained encoder is adapted to related downstream tasks, while still permitting task-specific divergence in later layers. This geometric consistency is complemented by the transfer results: VCR-JEPA outperforms baseline JEPA on five of the seven tasks under frozen evaluation and on five of the seven tasks after full fine-tuning. The frozen results are particularly informative because they indicate that the benefits of VCR are already evident in the pretrained representation itself, even when downstream adaptation is restricted to a lightweight prediction head. Overall, VCR-JEPA tends to transfer more effectively than baseline JEPA, including when the encoder is frozen, and it also preserves broader cross-task representational alignment after downstream adaptation. Cross-task CKA should nevertheless not be interpreted as a direct measure of predictive performance.

The effective-rank analysis complements CKA by showing that fine-tuning changes not only relationships among representations across layers but also their effective dimensional structure. Pretrained encoders exhibit comparatively stable, objective-dependent rank profiles, whereas fine-tuning produces more heterogeneous trajectories whose shape depends on both the downstream task and pretraining objective. These patterns indicate that adaptation redistributes representation variance across latent directions rather than merely displacing an otherwise unchanged geometry. VCR-JEPA generally maintains broader effective dimensionality through a larger portion of the encoder, while SL-MAE also delays rank concentration under NC vs.\ MCI fine-tuning, further demonstrating that the dimensional effect of auxiliary objectives is task-dependent. Importantly, higher effective rank does not systematically correspond to better downstream discrimination, as the effective-rank ordering does not follow the AUC ordering. Effective dimensionality should therefore not be interpreted as a proxy for task-relevant information. Together with the CKA results, this indicates that fine-tuning acts along at least two complementary axes: it reorganizes relationships among representations and redistributes representation variance across latent directions.

Viewed together, the auxiliary objectives introduce distinct and interpretable representational biases rather than uniformly improving representation quality. Spectral supervision preferentially emphasizes frequency-sensitive anatomical information, whereas VCR encourages broader and less redundant latent representations. Their downstream utility therefore depends on how well these biases align with the statistical structure of the target clinical problem.


Recent large-scale JEPA variants such as Neuro-JEPA demonstrate that substantial transfer gains can be achieved through architectural and objective-level innovations including multimodal routing, multiscale masking, and foreground-aware supervision. Our findings should therefore not be interpreted as evidence of an inherent limitation of predictive learning. Rather, Neuro-JEPA and the present study provide complementary perspectives: Neuro-JEPA demonstrates the potential of carefully designed predictive architectures at scale, while our controlled experiments show that the benefit of particular objective choices can vary substantially with the statistical characteristics of the downstream task.
More broadly, our findings also complement DINO-style approaches such as BrainFound, which learn view-consistent representations through self-distillation rather than pixel-space reconstruction or latent-target prediction. These approaches impose a different form of invariance on the learned representation and therefore provide a third point of comparison with the reconstruction-based and predictive paradigms examined here. Our results suggest that such objective-level differences should be understood as distinct representational biases whose downstream utility depends on the characteristics of the target clinical task.

Unlike recent multimodal neuroimaging foundation models that simultaneously exploit complementary MRI contrasts, our framework operates on individual MRI contrasts and evaluates transfer across heterogeneous single-contrast datasets. The strong performance of reconstruction-based pretraining under these conditions suggests that voxel-level anatomical fidelity remains highly valuable when clinically relevant information is encoded in subtle structural variations. Importantly, these results are obtained without access to complementary multimodal information or dedicated mechanisms for fusing contrast-specific evidence. Introducing multimodal inputs would confound the present objective-level comparison with the additional design problem of cross-modal information fusion. Reconstruction-based pretraining therefore remained highly competitive despite operating with substantially less input information and without additional multimodal fusion machinery.

We additionally underline the importance of reproducibility in medical self-supervised representation learning research. Existing work often uses non-standardized subject splits and preprocessing pipelines, making direct comparison difficult even when the same datasets are employed. To facilitate future benchmarking and transparent evaluation, we publicly release all subject-level train, validation, and test splits used for both pretraining and downstream evaluation.

\paragraph{Limitations} Despite the breadth of the evaluation, several limitations remain. First, the present study focuses on disease-detection tasks in structural brain MRI and evaluates primarily ViT-Base-scale encoders. Whether the objective-dependent representational patterns identified here persist at substantially larger model and pretraining scales remains to be established. Second, ADNI is evaluated using stratified five-fold cross-validation, whereas the remaining downstream cohorts are evaluated using fixed subject-level partitions, reflecting differences in cohort structure and, in some cases, limited sample availability. Consequently, split-to-split variability is characterized more comprehensively for ADNI than for the other downstream datasets. Evaluation on larger external cohorts would allow more extensive repeated-split or cross-validation analyses in future work. Third, our controlled comparison focuses on reconstruction-based and latent-predictive SSL formulations and their corresponding auxiliary objectives. Architecture-matched comparisons with newer self-distillation and contrastive-style frameworks, including DINOv3, constitute an important direction for future work. Finally, the task-dependent biases observed for spectral and variance--covariance regularization should be tested across additional clinical tasks, imaging domains, and larger-scale pretraining settings. Extending these analyses to multimodal neuroimaging may also be informative, although such settings introduce the separate methodological problem of integrating information across imaging contrasts and are therefore beyond the objective-level comparison considered here.

\section{Conclusion}

In this paper, we presented a systematic comparison of reconstruction-based
and latent-predictive self-supervised pretraining for 3D structural brain MRI,
using MAE and JEPA as representative formulations. We further examined two
paradigm-specific auxiliary objectives: a novel spectral-domain reconstruction loss
for MAE and variance--covariance regularization for JEPA. Across seven
dataset--task settings spanning four clinical problem domains, our results show
that downstream transfer depends not only on the broad self-supervised
pretraining paradigm, but also on the representational bias introduced by the
specific objective. Importantly, these effects are not uniform across clinical
tasks.

Reconstruction-based MAE pretraining generally achieved stronger downstream
performance than baseline JEPA, demonstrating that voxel-space reconstruction
remains highly effective for learning transferable structural MRI
representations. Spectral regularization provided its clearest benefits for
tasks characterized by pronounced fine-grained or high-frequency anatomical
structure, particularly Alzheimer's disease classification and tumor grading,
while producing smaller gains for NC vs.\ MCI and ASD. Variance--covariance
regularization improved JEPA most clearly for tasks whose discriminative signal
appears to involve multiple complementary sources of structural variation, but
its benefit was likewise task-dependent. Few-shot and frozen-encoder
experiments further showed that these objective-dependent differences are
already present in the pretrained representations and can remain relevant when
downstream supervision or adaptation is limited.

The representation analyses provide a complementary view of these findings.
CKA shows that the underlying pretraining paradigm establishes characteristic
representation geometries that are subsequently reorganized during fine-tuning
in a depth-, objective-, and task-dependent manner. Cross-task analysis further
indicates that VCR-JEPA tends to preserve broader representational alignment
when adapted to related downstream tasks. Effective-rank analysis shows that
fine-tuning also redistributes representation variance across latent directions,
with auxiliary objectives producing distinct task-dependent dimensional
profiles. Importantly, effective-rank ordering does not follow downstream AUC
ordering, indicating that effective dimensionality and task-relevant
discriminability are separable properties. Thus, objectives that preserve or
distribute representation variance more broadly are not necessarily those that
perform best on every downstream task.

Overall, our findings suggest that auxiliary objectives in medical
self-supervised learning should be understood as mechanisms that impose
distinct representational biases rather than as universally beneficial
improvements. Their value depends on how well those biases align with the
statistical and anatomical structure of the downstream clinical problem.
Accordingly, evaluation of self-supervised representation learning should
consider not only aggregate benchmark performance, but also how objective
design affects transfer across qualitatively different clinical tasks.

To support reproducible comparison, we publicly release all subject-level
pretraining and downstream evaluation partitions used in this study. Future
work should examine whether the objective-dependent patterns identified here
persist at larger model and pretraining scales, across additional imaging
domains and clinical tasks, and under emerging self-supervised formulations
including newer JEPA and DINO variants.

\section*{Funding}
This work was supported in part by the NIH under Grant P41 EB017183 and was performed under the rubric of the Center for Advanced Imaging Innovation and Research (CAI2R), an NIBIB National Center for Biomedical Imaging and Bioengineering. Additional support was provided by the Scientific Research Projects (BAP) Unit of Istanbul Technical University under Grant No. FHD-2026-48279, which supported computational resources, including access to NVIDIA A100 GPUs on the ITU UHEM cluster.

 
\section*{Declaration of Competing Interests}
The authors declare that they have no competing interests.
 
\section*{Data Availability}

The imaging datasets used in this study are available from their respective data custodians subject to their individual access and usage requirements. The authors do not redistribute imaging data. The code and subject-level partitions used for self-supervised pretraining and downstream evaluation are publicly available at \url{https://github.com/ituvisionlab/mjepa}.
 
\section*{Acknowledgements}

We thank all institutions and research consortia that made their datasets publicly available, making this research possible.
Data collection and sharing for the Alzheimer's Disease Neuroimaging Initiative (ADNI) is funded by the National
Institute on Aging (National Institutes of Health Grant U19 AG024904). The grantee organization is the Northern
California Institute for Research and Education. In the past, ADNI has also received funding from the National
Institute of Biomedical Imaging and Bioengineering, the Canadian Institutes of Health Research, and private
sector contributions through the Foundation for the National Institutes of Health (FNIH) including generous
contributions from the following: AbbVie, Alzheimer’s Association; Alzheimer’s Drug Discovery Foundation;
Araclon Biotech; BioClinica, Inc.; Biogen; Bristol-Myers Squibb Company; CereSpir, Inc.; Cogstate; Eisai Inc.;
Elan Pharmaceuticals, Inc.; Eli Lilly and Company; EuroImmun; F. Hoffmann-La Roche Ltd and its affiliated
company Genentech, Inc.; Fujirebio; GE Healthcare; IXICO Ltd.; Janssen Alzheimer Immunotherapy Research \&
Development, LLC.; Johnson \& Johnson Pharmaceutical Research \& Development LLC.; Lumosity; Lundbeck;
Merck \& Co., Inc.; Meso Scale Diagnostics, LLC.; NeuroRx Research; Neurotrack Technologies; Novartis
Pharmaceuticals Corporation; Pfizer Inc.; Piramal Imaging; Servier; Takeda Pharmaceutical Company; and
Transition Therapeutics. 

The NACC database is funded by NIA/NIH Grant U24 AG072122. SCAN is a multi-institutional project that was funded as a U24 grant (AG067418) by the National Institute on Aging in May 2020. Data collected by SCAN and shared by NACC are contributed by the NIA-funded ADRCs as follows:
Arizona Alzheimer's Center - P30 AG072980 (PI: Eric Reiman, MD); R01 AG069453 (PI: Eric Reiman (contact), MD); P30 AG019610 (PI: Eric Reiman, MD); and the State of Arizona which provided additional funding supporting our center; Boston University - P30 AG013846 (PI Neil Kowall MD); Cleveland ADRC - P30 AG062428 (James Leverenz, MD); Cleveland Clinic, Las Vegas - P20AG068053; Columbia - P50 AG008702 (PI Scott Small MD); Duke/UNC ADRC - P30 AG072958; Emory University - P30AG066511 (PI Levey Allan, MD, PhD); Indiana University - R01 AG19771 (PI Andrew Saykin, PsyD); P30 AG10133 (PI Andrew Saykin, PsyD); P30 AG072976 (PI Andrew Saykin, PsyD); R01 AG061788 (PI Shannon Risacher, PhD); R01 AG053993 (PI Yu-Chien Wu, MD, PhD); U01 AG057195 (PI Liana Apostolova, MD); U19 AG063911 (PI Bradley Boeve, MD); and the Indiana University Department of Radiology and Imaging Sciences; Johns Hopkins - P30 AG066507 (PI Marilyn Albert, Phd.); Mayo Clinic - P50 AG016574 (PI Ronald Petersen MD PhD); Mount Sinai - P30 AG066514 (PI Mary Sano, PhD); R01 AG054110 (PI Trey Hedden, PhD); R01 AG053509 (PI Trey Hedden, PhD); New York University - P30AG052012-01S2 (PI Thomas Wisniewski, MD); R01AG056031 (PI Ricardo Osorio, MD); R01AG056531 (PIs Ricardo Osorio, MD; Girardin Jean-Louis, PhD); Northwestern University - P30 AG013854 (PI Robert Vassar PhD); R01 AG045571 (PI Emily Rogalski, PhD); R56 AG045571, (PI Emily Rogalski, PhD); R01 AG067781, (PI Emily Rogalski, PhD); U19 AG073153, (PI Emily Rogalski, PhD); R01 DC008552, (M.-Marsel Mesulam, MD); R01 AG077444, (PIs M.-Marsel Mesulam, MD, Emily Rogalski, PhD); R01 NS075075 (PI Emily Rogalski, PhD); R01 AG056258 (PI Emily Rogalski, PhD); Oregon Health and Science University - P30 AG008017 (PI Jeffrey Kaye MD); R56 AG074321 (PI Jeffrey Kaye, MD); Rush University - P30 AG010161 (PI David Bennett MD); Stanford - P30AG066515; P50 AG047366 (PI Victor Henderson MD MS); University of Alabama, Birmingham - P20; University of California, Davis - P30 AG10129 (PI Charles DeCarli, MD); P30 AG072972 (PI Charles DeCarli, MD); University of California, Irvine - P50 AG016573 (PI Frank LaFerla PhD); University of California, San Diego - P30AG062429 (PI James Brewer, MD, PhD); University of California, San Francisco - P30 AG062422 (Rabinovici, Gil D., MD); University of Kansas - P30 AG035982 (Russell Swerdlow, MD); University of Kentucky - P30 AG028283-15S1 (PIs Linda Van Eldik, PhD and Brian Gold, PhD); University of Michigan ADRC - P30AG053760 (PI Henry Paulson, MD, PhD) P30AG072931 (PI Henry Paulson, MD, PhD) Cure Alzheimer's Fund 200775 - (PI Henry Paulson, MD, PhD) U19 NS120384 (PI Charles DeCarli, MD, University of Michigan Site PI Henry Paulson, MD, PhD) R01 AG068338 (MPI Bruno Giordani, PhD, Carol Persad, PhD, Yi Murphey, PhD) S10OD026738-01 (PI Douglas Noll, PhD) R01 AG058724 (PI Benjamin Hampstead, PhD) R35 AG072262 (PI Benjamin Hampstead, PhD) W81XWH2110743 (PI Benjamin Hampstead, PhD) R01 AG073235 (PI Nancy Chiaravalloti, University of Michigan Site PI Benjamin Hampstead, PhD) 1I01RX001534 (PI Benjamin Hampstead, PhD) IRX001381 (PI Benjamin Hampstead, PhD); University of New Mexico - P20 AG068077 (Gary Rosenberg, MD); University of Pennsylvania - State of PA project 2019NF4100087335 (PI David Wolk, MD); Rooney Family Research Fund (PI David Wolk, MD); R01 AG055005 (PI David Wolk, MD); University of Pittsburgh - P50 AG005133 (PI Oscar Lopez MD); University of Southern California - P50 AG005142 (PI Helena Chui MD); University of Washington - P50 AG005136 (PI Thomas Grabowski MD); University of Wisconsin - P50 AG033514 (PI Sanjay Asthana MD FRCP); Vanderbilt University - P20 AG068082; Wake Forest - P30AG072947 (PI Suzanne Craft, PhD); Washington University, St. Louis - P01 AG03991 (PI John Morris MD); P01 AG026276 (PI John Morris MD); P20 MH071616 (PI Dan Marcus); P30 AG066444 (PI John Morris MD); P30 NS098577 (PI Dan Marcus); R01 AG021910 (PI Randy Buckner); R01 AG043434 (PI Catherine Roe); R01 EB009352 (PI Dan Marcus); UL1 TR000448 (PI Brad Evanoff); U24 RR021382 (PI Bruce Rosen); Avid Radiopharmaceuticals / Eli Lilly; Yale - P50 AG047270 (PI Stephen Strittmatter MD PhD); R01AG052560 (MPI: Christopher van Dyck, MD; Richard Carson, PhD); R01AG062276 (PI: Christopher van Dyck, MD); 1Florida - P30AG066506-03 (PI Glenn Smith, PhD); P50 AG047266 (PI Todd Golde MD PhD).

MRI data were obtained from the Standardized Centralized Alzheimer's and 
Related Dementias Neuroimaging (SCAN) Initiative, funded by NIA/NIH Grant 
U24 AG067418. SCAN is a collaboration between UC Berkeley, Mayo Clinic, 
University of Michigan, UC Davis, NACC, and LONI.

PPMI – a public-private partnership – is funded by the Michael J. Fox Foundation for Parkinson’s Research and funding partners, including 4D Pharma, Abbvie,
AcureX, Allergan, Amathus Therapeutics, Aligning Science Across Parkinson's, AskBio, Avid Radiopharmaceuticals, BIAL, BioArctic, Biogen, Biohaven,
BioLegend, BlueRock Therapeutics, Bristol-Myers Squibb, Calico Labs, Capsida Biotherapeutics, Celgene, Cerevel Therapeutics, Coave Therapeutics, DaCapo
Brainscience, Denali, Edmond J. Safra Foundation, Eli Lilly, Gain Therapeutics, GE HealthCare, Genentech, GSK, Golub Capital, Handl Therapeutics, Insitro,
Jazz Pharmaceuticals, Johnson \& Johnson Innovative Medicine, Lundbeck, Merck, Meso Scale Discovery, Mission Therapeutics, Neurocrine Biosciences,
Neuron23, Neuropore, Pfizer, Piramal, Prevail Therapeutics, Roche, Sanofi, Servier, Sun Pharma Advanced Research Company, Takeda, Teva, UCB, Vanqua Bio,
Verily, Voyager Therapeutics, the Weston Family Foundation and Yumanity Therapeutics.

We acknowledge the principal investigators of OASIS-3: T. Benzinger, D. Marcus, J. Morris; NIH P30
AG066444, P50 AG00561, P30 NS09857781, P01 AG026276, P01 AG003991,
R01 AG043434, UL1 TR000448, R01 EB009352. AV-45 doses were provided
by Avid Radiopharmaceuticals, a wholly owned subsidiary of Eli Lilly.

We gratefully acknowledge all institutions, consortia, and research teams whose publicly available datasets supported this work, including the Alzheimer's Disease Neuroimaging Initiative (ADNI), the National Alzheimer's Coordinating Center and Standardized Centralized Alzheimer's and Related Dementias Neuroimaging initiative (NACC/SCAN), the Parkinson's Progression Markers Initiative (PPMI), the Autism Brain Imaging Data Exchange (ABIDE), OASIS, IXI, BraTS, and UCSF-PDGM. 

G.U. acknowledges the support and research environment provided by New York University during her 2024--2025 sabbatical leave, during which foundational aspects of this work were developed.

We further thank all researchers, clinicians, coordinators, and study participants whose efforts made these publicly available datasets possible.

\section*{Declaration of generative AI and AI-assisted technologies in the manuscript preparation process}
During the preparation of this work, the authors used ChatGPT (OpenAI) in order to assist with grammar and language editing. After using this tool, the authors reviewed and edited the content as needed and take full responsibility for the content of the published article.






\printcredits

\bibliographystyle{model1-num-names}

\bibliography{cas-refs}

@misc{drozdov2024videorepresentationlearningjointembedding,
      title={Video Representation Learning with Joint-Embedding Predictive Architectures}, 
      author={Katrina Drozdov and Ravid Shwartz-Ziv and Yann LeCun},
      year={2024},
      howpublished={arXiv preprint},
      url={https://arxiv.org/abs/2412.10925}
}

@article{yu2026mrbc,
  author  = {Yu, Hu and Zhang, Yiyu and Li, Jingming and Song, Zhiwei and Lyu, Zhengyuan and Fu, Si and Yang, Libin and Ruan, Sen and Guo, Xiaojuan},
  title   = {Morphology-Aware Representational Brain Connectome ({MRBC}): A Deep Feature-Driven Engine with Robust Representation, Reproducibility, and Clinical Relevance},
  journal = {Neurocomputing},
  volume  = {694},
  pages   = {133938},
  year    = {2026},
  issn    = {0925-2312},
  doi     = {10.1016/j.neucom.2026.133938}
}

@article{fedata,
  author  = {Dai, Jian and Wu, Hao and Liu, Huan and Yu, Liheng and Hu, Xing and Liu, Xiao and Geng, Daoying},
  title   = {{FedATA}: Adaptive Attention Aggregation for Federated Self-Supervised Medical Image Segmentation},
  journal = {Neurocomputing},
  volume  = {613},
  pages   = {128691},
  year    = {2025},
  issn    = {0925-2312},
  doi     = {10.1016/j.neucom.2024.128691}
}

@misc{vjepa24,
  title={Revisiting feature prediction for learning visual representations from video},
  author={Bardes, Alain and Garrido, Quentin and Ponce, Jean and Chen, Xinlei and Rabbat, Michael and LeCun, Yann and Assran, Mahmoud and Ballas, Nicolas},
      year={2024},
      howpublished={arXiv preprint},
      url={https://arxiv.org/abs/2404.08471}
}

@inproceedings{wald2024spark3d,
  title={Revisiting MAE Pre-training for 3D Medical Image Segmentation},
  author={Wald, Tassilo and Ulrich, Constantin and Lukyanenko, Stanislav and Goncharov, Andrei and Paderno, Alberto and Miller, Maximilian and Maerkisch, Leander and Jaeger, Paul and Maier-Hein, Klaus},
  booktitle={Proceedings of the IEEE/CVF Conference on Computer Vision and Pattern Recognition (CVPR)},
  pages={5186--5196},
  year={2025},
  doi={10.1109/CVPR52734.2025.00489}
}

@misc{rajamohan2026ssl,
  title        = {Self-Supervised Learning for Knee Osteoarthritis: Diagnostic Limitations and Prognostic Value of Hospital Data},
  author       = {Rajamohan, Haresh Rengaraj and Chen, Yuxuan and Cho, Kyunghyun and Deniz, Cem M.},
  year         = {2026},
  howpublished = {arXiv preprint},
  doi          = {10.48550/arXiv.2603.24903}
}

@inproceedings{gao2026mass,
  title     = {Learning Generalizable 3D Medical Image Representations 
               from Mask-Guided Self-Supervision},
  author    = {Gao, Yunhe and Zhang, Yabin and Wang, Chong and Liu, Jiaming 
               and Varma, Maya and Delbrouck, Jean-Benoit and Chaudhari, 
               Akshay and Langlotz, Curtis},
  booktitle = {Proceedings of the IEEE/CVF Conference on Computer Vision 
               and Pattern Recognition (CVPR)},
  year      = {2026}
}

@misc{usjepa,
  title        = {{US-JEPA}: A Joint Embedding Predictive Architecture for Medical Ultrasound},
  author       = {Radhachandran, Ashwath and Ivezi{\'c}, Vedrana and Athreya, Shreeram and Anilkumar, Ronit and Arnold, Corey W. and Speier, William},
  year         = {2026},
  howpublished = {arXiv preprint},
  doi          = {10.48550/arXiv.2602.19322}
}

@inproceedings{he2022masked,
  title={Masked Autoencoders Are Scalable Vision Learners},
  author={He, Kaiming and Chen, Xinlei and Xie, Saining and Li, Yanghao and Doll{\'a}r, Piotr and Girshick, Ross},
  booktitle={Proceedings of the IEEE/CVF Conference on Computer Vision and Pattern Recognition (CVPR)},
  pages={16000--16009},
  year={2022},
  doi={10.1109/CVPR52688.2022.01553}
}

@inproceedings{assran2023ijepa,
  title={Self-Supervised Learning from Images with a Joint-Embedding Predictive Architecture},
  author={Assran, Mahmoud and Duval, Quentin and Misra, Ishan and Bojanowski, Piotr and Vincent, Pascal and Rabbat, Michael and LeCun, Yann and Ballas, Nicolas},
  booktitle={Proceedings of the IEEE/CVF Conference on Computer Vision and Pattern Recognition (CVPR)},
  pages={15619--15629},
  year={2023},
  doi={10.1109/CVPR52729.2023.01499}
}

@inproceedings{wald2025openmind,
  title={Abstract: {An} {OpenMind} for {3D} Medical Vision Self-supervised Learning},
  author={Ulrich, Constantin and Wald, Tassilo and Suprijadi, Jonathan and Ziegler, Sebastian and Nohel, Michal and Peretzke, Robin and K{\"o}hler, Gregor and Maier-Hein, Klaus},
  booktitle={Bildverarbeitung f{\"u}r die Medizin 2026},
  series={Informatik aktuell},
  pages={109},
  year={2026},
  publisher={Springer Vieweg, Wiesbaden},
  doi={10.1007/978-3-658-51100-5_22}
}

@INPROCEEDINGS{rui2025brainmvp,
  author={Rui, Shaohao and Chen, Lingzhi and Tang, Zhenyu and Wang, Lilong and Liu, Mianxin and Zhang, Shaoting and Wang, Xiaosong},
  booktitle={2025 IEEE/CVF Conference on Computer Vision and Pattern Recognition (CVPR)}, 
  title={Multi-modal Vision Pre-training for Medical Image Analysis}, 
  year={2025},
  volume={},
  number={},
  pages={5164-5174},
  doi={10.1109/CVPR52734.2025.00487}}

@article{zhu2025foundation,
  author    = {Zhu, Weicheng and Huang, Haoxu and Tang, Huanze and Musthyala, Rushabh and Yu, Boyang and Chen, Long and Vega, Emilio and O'Donnell, Thomas and Hayek, Reya and Kuohn, Lindsey and Dehkharghani, Seena and Frontera, Jennifer A. and Masurkar, Arjun V. and Melmed, Kara and Razavian, Narges},
  title     = {3{D} foundation model for generalizable disease detection in head computed tomography},
  journal   = {Nat. Biomed. Eng.},
  year      = {2026},
  month     = {April},
  doi       = {10.1038/s41551-026-01668-w},

  issn      = {2157-846X}
}

@Article{avesta2023,
AUTHOR = {Avesta, Arman and Hossain, Sajid and Lin, MingDe and Aboian, Mariam and Krumholz, Harlan M. and Aneja, Sanjay},
TITLE = {Comparing 3D, 2.5D, and 2D Approaches to Brain Image Auto-Segmentation},
JOURNAL = {Bioengineering},
VOLUME = {10},
YEAR = {2023},
NUMBER = {2},
pages = {181},
doi     = {10.3390/bioengineering10020181}
}

@Article{liu2022razavian,
AUTHOR       = {Liu, Sheng and Masurkar, Arjun V. and Rusinek, Henry and Chen, Jingyun and Ben, Zhang and Zhu, Weicheng and Fernandez-Grande, Carlos and Razavian, Narges},
TITLE        = {Generalizable deep learning model for early {Alzheimer’s} disease detection from structural {MRIs}},
JOURNAL      = {Sci. Rep.},
VOLUME       = {12},
YEAR         = {2022},
NUMBER       = {1},
pages = {17106},
DOI          = {10.1038/s41598-022-20674-x},
ISSN         = {2045-2322}
}

@inproceedings{chalcroft2025,
  author    = {Chalcroft, Liam and Crinion, Jenny and Price, Cathy J. and Ashburner, John},
  title     = {Unified {3D} {MRI} Representations via Sequence-Invariant Contrastive Learning},
  booktitle = {Simulation and Synthesis in Medical Imaging (SASHIMI 2025)},
  series    = {Lecture Notes in Computer Science},
  volume    = {16085},
  pages     = {63--74},
  year      = {2025},
  publisher = {Springer, Cham},
  doi       = {10.1007/978-3-032-05573-6_7}
}

@article{xu20253dino,
  author  = {Xu, Tony and Hosseini, Sepehr and Anderson, Chris and Rinaldi, Anthony and Krishnan, Rahul G. and Martel, Anne L. and Goubran, Maged},
  title   = {A generalizable {3D} framework and model for self-supervised learning in medical imaging},
  journal = {npj Digit. Med.},
  volume  = {8},
  number  = {1},
  pages   = {639},
  year    = {2025},
  doi     = {10.1038/s41746-025-02035-w}
}

@misc{mullerfranzes2024,
  author       = {M{\"u}ller-Franzes, Gustav and Khader, Firas and Siepmann, Robert and Han, Tianyu and Kather, Jakob Nikolas and Nebelung, Sven and Truhn, Daniel},
  title        = {Medical Slice Transformer: Improved Diagnosis and Explainability on {3D} Medical Images with {DINO}v2},
  year         = {2024},
  howpublished = {arXiv Preprint},
  doi          = {10.48550/arXiv.2411.15802}
}

@misc{zheng2025fair,
      title={Towards Fair Medical AI: Adversarial Debiasing of 3D CT Foundation Embeddings}, 
      author={Guangyao Zheng and Michael A. Jacobs and Vladimir Braverman and Vishwa S. Parekh},
      year={2025},
      howpublished={arXiv Preprint},
      doi = {10.48550/arXiv.2502.04386}

}

@ARTICLE{ye2025cads,
  author={Ye, Yiwen and Zhang, Jianpeng and Chen, Ziyang and Xia, Yong},
  journal={IEEE Trans. Med. Imaging}, 
  title={{CADS}: A Self-Supervised Learner via Cross-Modal Alignment and Deep Self-Distillation for CT Volume Segmentation}, 
  year={2025},
  volume={44},
  number={1},
  pages={118-129},
  doi={10.1109/TMI.2024.3431916}}

@article{jiang2024self,
author = {Jiang, Jue and Rangnekar, Aneesh and Veeraraghavan, Harini},
title = {Self-supervised learning improves robustness of deep learning lung tumor segmentation models to CT imaging differences},
journal = {Med. Phys.},
volume = {52},
number = {3},
pages = {1573-1588},
doi = {10.1002/mp.17541},
year = {2025}
}

@article{yu2024drasclr,
title = {{DrasCLR}: A self-supervised framework of learning disease-related and anatomy-specific representation for 3D lung CT images},
journal = {Med. Image Anal.},
volume = {92},
pages = {103062},
year = {2024},
issn = {1361-8415},
doi = {10.1016/j.media.2023.103062},
author = {Ke Yu and Li Sun and Junxiang Chen and Maxwell Reynolds and Tigmanshu Chaudhary and Kayhan Batmanghelich}
}

@inproceedings{taleb-neurips2020,
  author    = {Taleb, Aiham and Loetzsch, Winfried and Danz, Noel and Severin, Julius and Gaertner, Thomas and Bergner, Benjamin and Lippert, Christoph},
  title     = {{3D} Self-Supervised Methods for Medical Imaging},
  booktitle = {Advances in Neural Information Processing Systems},
  editor    = {Larochelle, H. and Ranzato, M. and Hadsell, R. and Balcan, M.F. and Lin, H.},
  volume    = {33},
  pages     = {18158--18172},
  publisher = {Curran Associates, Inc.},
  year      = {2020}
}

@article{adni_dataset,
  title={Alzheimer's Disease Neuroimaging Initiative ({ADNI}): Clinical characterization},
  author={Petersen, R C and Aisen, P S and Beckett, L A and Donohue, M C and Gamst, A C and Harvey, D J and Jack, C R and Jagust, W J and Shaw, L M and Toga, A W and Trojanowski, J Q and Weiner, M W},
  journal={Neurology},
  volume={74},
  number={3},
  pages={201--209},
  year={2010},
  publisher={Wolters Kluwer},
  doi={10.1212/WNL.0b013e3181cb3e25},
  pmid={20042704},
  pmcid={PMC2809036}
}

@article{ppmi_dataset,
  title={The parkinson’s progression markers initiative (ppmi)–establishing a pd biomarker cohort.},
  author={Marek, K and Chowdhury, S and Siderowf, SL and Coffey, CS and Caspell-Garcia, C and Simuni, T and Jennings, D and Tanner, CM and Trojanowski, JQ and others},
  journal={Ann. Clin. Transl. Neurol.},
  volume={5},
  number={12},
  pages={1460--1477},
  year={2018},
  doi = {10.1002/acn3.644}
}

@inproceedings{li-2024-isbi,
  author={Li, Haofeng and Ouyang, Yiming and Wan, Xiang},
  booktitle={2024 IEEE International Symposium on Biomedical Imaging (ISBI)}, 
  title={Self-Supervised Alignment Learning For Medical Image Segmentation}, 
  year={2024},
  volume={},
  number={},
  pages={1-5},
  doi={10.1109/ISBI56570.2024.10635540}}

@INPROCEEDINGS{tang2022swintransformers,
  author={Tang, Yucheng and Yang, Dong and Li, Wenqi and Roth, Holger R. and Landman, Bennett and Xu, Daguang and Nath, Vishwesh and Hatamizadeh, Ali},
  title={Self-Supervised Pre-Training of Swin Transformers for 3D Medical Image Analysis}, 
  booktitle = {Proceedings of the IEEE/CVF Conference on Computer Vision and Pattern Recognition (CVPR)},
  pages     = {20698--20708},
  year      = {2022},
  doi       = {10.1109/CVPR52688.2022.02007}
  
  }

@misc{dong2025mricoreFM,
      title={{MRI-CORE}: A Foundation Model for Magnetic Resonance Imaging}, 
      author={Haoyu Dong and Yuwen Chen and Hanxue Gu and Nicholas Konz and Yaqian Chen and Qihang Li and Maciej A. Mazurowski},
      year={2025},
      howpublished={arXiv Preprint},
      doi={10.48550/arXiv.2506.12186}
}

@article{deniz2025mrtransformer,
  author  = {Zhang, Chaojie and Chen, Shengjia and Cigdem, Ozkan and 
             Rajamohan, Haresh Rengaraj and Cho, Kyunghyun and 
             Kijowski, Richard and Deniz, Cem M.},
  title   = {{MR-Transformer}: A Vision Transformer-based Deep Learning 
             Model for Total Knee Replacement Prediction Using {MRI}},
  journal = {Radiol. Artif. Intell.},
  year    = {2025},
  volume = {7},
number = {5},
pages = {e240373},
  doi     = {10.1148/ryai.240373},
}

@article{vanDerMaaten2008,
  author  = {van der Maaten, Laurens and Hinton, Geoffrey},
  title   = {Visualizing Data using {t-SNE}},
  journal = {J. Mach. Learn. Res.},
  year    = {2008},
  volume  = {9},
  pages   = {2579--2605},
  url     = {http://jmlr.org/papers/v9/vandermaaten08a.html}
}

@inproceedings{Zhang2022GFNet,
author    = {Zhang, Shengjie and Chen, Xiang and Ren, Bohan
and Yang, Haibo and Yu, Ziqi and Zhang, Xiao-Yong
and Zhou, Yuan},
title     = {{3D} Global {Fourier} Network for {Alzheimer's}
Disease Diagnosis Using Structural {MRI}},
booktitle = {Medical Image Computing and Computer Assisted
Intervention -- {MICCAI} 2022},
pages     = {34--43},
year      = {2022},
publisher = {Springer Nature Switzerland},
doi       = {10.1007/978-3-031-16431-6_4}
}

@article{Leandrou2020,
author  = {Leandrou, Stephanos and Lamnisos, Demetris and Mamais,
Ioannis and Kyriacou, Panicos A. and Pattichis,
Constantinos S.},
title   = {Assessment of {Alzheimer's} Disease Based on Texture
Analysis of the Entorhinal Cortex},
journal = {Front. Aging Neurosci.},
year    = {2020},
volume  = {12},
pages   = {176},
doi     = {10.3389/fnagi.2020.00176}
}

@article{Wang2022,
author  = {Wang, Luoyu and Feng, Qi and Ge, Xiuhong and
Chen, Fenyang and Yu, Bo and Chen, Bing and
Liao, Zhengluan and Lin, Biying and Lv, Yating
and Ding, Zhongxiang},
title   = {Textural Features Reflecting Local Activity of the
Hippocampus Improve the Diagnosis of {Alzheimer's}
Disease and Amnestic Mild Cognitive Impairment:
A Radiomics Study Based on Functional Magnetic
Resonance Imaging},
journal = {Front. Neurosci.},
year    = {2022},
volume  = {16},
pages   = {970245},
doi     = {10.3389/fnins.2022.970245}
}

@article{Wearn2023,
author  = {Wearn, Alfie and Raket, Lars Lau and
Collins, D. Louis and Spreng, R. Nathan},
title   = {Longitudinal Changes in Hippocampal Texture
from Healthy Aging to {Alzheimer's} Disease},
journal = {Brain Commun.},
year    = {2023},
volume  = {5},
number  = {4},
pages   = {fcad195},
doi     = {10.1093/braincomms/fcad195}
}

@article{Lee2020,
author  = {Lee, Hyunna and Lee, Seokjun and Kim, Kun Woo},
title   = {Magnetic Resonance Imaging Texture Predicts
Progression to Dementia due to {Alzheimer}
Disease Earlier than Hippocampal Volume},
journal = {J. Psychiatry Neurosci.},
year    = {2020},
volume  = {45},
number  = {1},
pages   = {7--14},
doi     = {10.1503/jpn.180171}
}

@article{Chaddad2017,
  title={Multi-scale radiomic analysis of sub-cortical regions in 
         {MRI} related to autism, gender and age},
  author={Chaddad, Ahmad and Desrosiers, Christian and Toews, Matthew},
  journal={Sci. Rep.},
  volume={7},
  pages={45639},
  year={2017},
  publisher={Nature Publishing Group},
  doi={10.1038/srep45639}
}

@inproceedings{Kumar2022,
  title={Fine-Tuning can Distort Pretrained Features and 
         Underperform Out-of-Distribution},
  author={Kumar, Ananya and Raghunathan, Aditi and Jones, Robbie 
          and Ma, Tengyu and Liang, Percy},
  booktitle={International Conference on Learning Representations ({ICLR})},
  year={2022},
  doi={10.48550/arXiv.2202.10054}
}

@misc{RepSim2025,
  title={Keeping Representation Similarity in Finetuning for 
         Medical Image Analysis},
  author={Zu, Wenqiang and Xie, Shenghao and Chen, Hao and Liang, Yiming and Ma, Lei},
  year={2025},
  howpublished = {arXiv Preprint},
  doi = {
10.48550/arXiv.2503.07399},

}

@article{Bardes2022,
  author    = {Bardes, Adrien and Ponce, Jean and LeCun, Yann},
  title     = {{VICReg}: Variance-Invariance-Covariance Regularization
               for Self-Supervised Learning},
  journal   = {International Conference on Learning Representations},
  year      = {2022},
  url       = {https://arxiv.org/abs/2105.04906}
}

@inproceedings{Zbontar2021,
  author    = {Zbontar, Jure and Jing, Li and Misra, Ishan and
               LeCun, Yann and Deny, Stephane},
  title     = {{Barlow Twins}: Self-Supervised Learning via
               Redundancy Reduction},
  booktitle = {Proceedings of the 38th International Conference
               on Machine Learning},
  series    = {PMLR},
  volume    = {139},
  year      = {2021},
  pages     = {12310--12320},
  doi={10.48550/arXiv.2103.03230}
}

@article{Frisoni2010,
  author    = {Frisoni, Giovanni B. and Fox, Nick C. and
               Jack, Clifford R. and Scheltens, Philip and
               Thompson, Paul M.},
  title     = {The clinical use of structural {MRI} in
               {Alzheimer} disease},
  journal   = {Nat. Rev. Neurol.},
  volume    = {6},
  number    = {2},
  pages     = {67--77},
  year      = {2010},
  doi       = {10.1038/nrneurol.2009.215},
  pmcid     = {PMC2938772}
}

@article{Planche2022,
  author    = {Planche, Vincent and Manjon, Jos\'{e} V. and
               Mansencal, Boris and Lanuza, Enrique and
               Tourdias, Thomas and Catheline,
               Gwena\"{e}lle and Coup\'{e}, Pierrick},
  title     = {Structural progression of {Alzheimer's}
               disease over decades: the {MRI} staging
               scheme},
  journal   = {Brain Commun.},
  volume    = {4},
  number    = {3},
  pages     = {fcac109},
  year      = {2022},
  doi       = {10.1093/braincomms/fcac109},
  pmcid     = {PMC9113086}
}

@article{Rose2009,
  author    = {Rose, Chris J. and Mills, Samantha J. and
               O'Connor, James P. B. and Buonaccorsi, Giovanni A. and
               Roberts, Caleb and Watson, Yvonne and
               Cheung, Susan and Zhao, Sha and
               Whitcher, Brandon and Jackson, Alan and
               Parker, Geoffrey J. M.},
  title     = {Quantifying Spatial Heterogeneity in Dynamic
               Contrast-Enhanced {MRI} Parameter Maps},
  journal   = {Magn. Reson. Med.},
  volume    = {62},
  number    = {2},
  pages     = {488--499},
  year      = {2009},
  doi       = {10.1002/mrm.22003}
}

@article{Gillies2016,
  author    = {Gillies, Robert J. and Kinahan, Paul E. and
               Hricak, Hedvig},
  title     = {Radiomics: Images Are More than Pictures,
               They Are Data},
  journal   = {Radiology},
  volume    = {278},
  number    = {2},
  pages     = {563--577},
  year      = {2016},
  doi       = {10.1148/radiol.2015151169}
}

@article{Pennanen2004,
  author    = {Pennanen, Corina and Kivipelto, Miia and
               Tuomainen, Sari and Hartikainen, Paivi and
               Hanninen, Tuomo and Laakso, Mikko P. and
               Hallikainen, Merja and Vanhanen, Matti and
               Nissinen, Aulikki and Helkala, Eeva-Liisa and
               Vainio, Pauli and Vanninen, Ritva and
               Partanen, Kaarina and Soininen, Hilkka},
  title     = {Hippocampus and entorhinal cortex in mild
               cognitive impairment and early {Alzheimer's} disease},
  journal   = {Neurobiol. Aging},
  volume    = {25},
  number    = {3},
  pages     = {303--310},
  year      = {2004},
  doi       = {10.1016/S0197-4580(03)00084-8}
}

@inproceedings{chen2020simclr,
author = {Chen, Ting and Kornblith, Simon and Norouzi, Mohammad and Hinton, Geoffrey},
title = {A simple framework for contrastive learning of visual representations},
year = {2020},
publisher = {JMLR.org},
booktitle = {Proceedings of the 37th International Conference on Machine Learning},
articleno = {149},
numpages = {11},
series = {ICML'20},
doi = {10.5555/3524938.3525087}
}

@article{vanRooij2018,
  author    = {van Rooij, Daan and Anagnostou, Evdokia and Arango, Celso and Auzias, Guillaume and Behrmann, Marlene and Busatto, Geraldo F. and Calderoni, Sara and Daly, Eileen and Deruelle, Christine and Di Martino, Adriana and Dinstein, Ilan and Duran, Fabio Luis Souza and Durston, Sarah and Ecker, Christine and Fair, Damien and Fedor, Jennifer and Fitzgerald, Jackie and Freitag, Christine M. and Gallagher, Louise and Gori, Ilaria and Haar, Shlomi and Hoekstra, Liesbeth and Jahanshad, Neda and Jalbrzikowski, Maria and Janssen, Joost and Lerch, Jason and Luna, Beatriz and Martinho, Mauricio Moller and McGrath, Jane and Muratori, Filippo and Murphy, Clodagh M. and Murphy, Declan G. M. and O'Hearn, Kirsten and Oranje, Bob and Parellada, Mara and Retico, Alessandra and Rosa, Pedro and Rubia, Katya and Shook, Devon and Taylor, Margot and Thompson, Paul M. and Tosetti, Michela and Wallace, Gregory L. and Zhou, Fengfeng and Buitelaar, Jan K.},
  title     = {Cortical and Subcortical Brain Morphometry
               Differences Between Patients With Autism
               Spectrum Disorder and Healthy Individuals
               Across the Lifespan: Results From the
               {ENIGMA} {ASD} Working Group},
  journal   = {Am. J. Psychiatry},
  volume    = {175},
  number    = {4},
  pages     = {359--369},
  year      = {2018},
  doi       = {10.1176/appi.ajp.2017.17010100},
  pmid      = {29145754}
}

@article{Haar2016,
  author    = {Haar, Shlomi and Berman, Sigal and
               Behrmann, Marlene and Dinstein, Ilan},
  title     = {Anatomical Abnormalities in Autism?},
  journal   = {Cereb. Cortex},
  volume    = {26},
  number    = {4},
  pages     = {1440--1452},
  year      = {2016},
  doi       = {10.1093/cercor/bhu242},
  pmid      = {25316335}
}

@article{Feng2018,
  author    = {Feng, Feng and Wang, Pan and Zhao, Kun and Zhou, Bo and 
               Yao, Hongxiang and Meng, Qingqing and Wang, Lei and 
               Zhang, Zengqiang and Ding, Yanhui and Wang, Luning and 
               An, Ningyu and Zhang, Xi and Liu, Yong},
  title     = {Radiomic Features of Hippocampal Subregions in {Alzheimer's} 
               Disease and Amnestic Mild Cognitive Impairment},
  journal   = {Front. Aging Neurosci.},
  volume    = {10},
  pages     = {290},
  year      = {2018},
  doi       = {10.3389/fnagi.2018.00290},
  pmid      = {30319396},
  pmcid     = {PMC6167420}
}

@article{Kinoshita2016,
  author    = {Kinoshita, Manabu and Sakai, Mio and Arita, Hideyuki and 
               Shofuda, Tomoko and Chiba, Yasuyoshi and Kagawa, Naoki and 
               Watanabe, Yoshiyuki and Hashimoto, Naoya and Fujimoto, Yasunori 
               and Yoshimine, Toshiki and Nakanishi, Katsuyuki and Kanemura, Yonehiro},
  title     = {Introduction of High Throughput Magnetic Resonance {T2}-Weighted 
               Image Texture Analysis for {WHO} Grade 2 and 3 Gliomas},
  journal   = {PLoS ONE},
  volume    = {11},
  number    = {10},
  pages     = {e0164268},
  year      = {2016},
  doi       = {10.1371/journal.pone.0164268},
  pmid      = {27716832},
  pmcid     = {PMC5055327}
}

@ARTICLE{Shao2025,
  author={Shao, Minye and Wang, Zeyu and Duan, Haoran and Huang, Yawen and Zhai, Bing and Wang, Shizheng and Long, Yang and Zheng, Yefeng},
  journal={IEEE Transactions on Medical Imaging}, 
  title={Rethinking Brain Tumor Segmentation From the Frequency Domain Perspective}, 
  year={2025},
  volume={44},
  number={11},
  pages={4536-4553},
  doi={10.1109/TMI.2025.3579213}}

@article{DiMartino2014,
  author    = {Di Martino, A. and Yan, C-G and Li, Q. and Denio, E. 
               and Castellanos, F.X. and Alaerts, K. and Anderson, J.S. 
               and Assaf, M. and Bookheimer, S.Y. and Dapretto, M. 
               and Deen, B. and Delmonte, S. and Dinstein, I. 
               and Ertl-Wagner, B. and Fair, D.A. and Gallagher, L. 
               and Kennedy, D.P. and Keown, C.L. and Keysers, C. 
               and Lainhart, J.E. and Lord, C. and Luna, B. 
               and Menon, V. and Minshew, N.J. and Monk, C.S. 
               and Mueller, S. and M{\"u}ller, R-A and Nebel, M.B. 
               and Nigg, J.T. and O'Hearn, K. and Pelphrey, K.A. 
               and Peltier, S.J. and Rudie, J.D. and Sunaert, S. 
               and Thioux, M. and Tyszka, J.M. and Uddin, L.Q. 
               and Verhoeven, J.S. and Wenderoth, N. and Wiggins, J.L. 
               and Mostofsky, S.H. and Milham, M.P.},
  title     = {The autism brain imaging data exchange: towards a 
               large-scale evaluation of the intrinsic brain 
               architecture in autism},
  journal   = {Mol. Psychiatry},
  volume    = {19},
  number    = {6},
  pages     = {659--667},
  year      = {2014},
  doi       = {10.1038/mp.2013.78}
}

@article{Mazher2025,
  author  = {Mazher, Moona and Parker, Geoff J.M. and Alexander, Daniel C.},
  title   = {Towards generalisable foundation models for brain {MRI}},
  journal = {npj Imaging},
  year    = {2026},
  doi     = {10.1038/s44303-026-00176-5}
}

@article{Oquab2023,
  author  = {Oquab, Maxime and Darcet, Timoth{\'e}e and Moutakanni, Th{\'e}o
             and Vo, Huy V. and Szafraniec, Marc and Khalidov, Vasil
             and Fernandez, Pierre and Haziza, Daniel and Massa, Francisco
             and El-Nouby, Alaaeldin and Assran, Mahmoud and Ballas, Nicolas
             and Galuba, Wojciech and Howes, Russell and Huang, Po-Yao
             and Li, Shang-Wen and Misra, Ishan and Rabbat, Michael
             and Sharma, Vasu and Synnaeve, Gabriel and Xu, Hu
             and J{\'e}gou, Herv{\'e} and Mairal, Julien and Labatut, Patrick
             and Joulin, Armand and Bojanowski, Piotr},
  title   = {{DINOv2}: Learning Robust Visual Features without Supervision},
  journal = {Trans. Mach. Learn. Res.},
  year    = {2024}
}

@misc{ixi_dataset,
  title        = {{IXI} Dataset},
  author       = {{Brain Development}},
  year         = {2007},
  howpublished = {\url{https://brain-development.org/ixi-dataset/}},
  note         = {[dataset]. Funded by EPSRC GR/S21533/02. 
                  Licensed under CC BY-SA 3.0. Accessed: June 2026}
}

@article{abide_dataset,
  author  = {Di Martino, Adriana and O'Connor, David and Chen, Bosi 
             and Alaerts, Kaat and Anderson, Jeffrey S. and Assaf, Michal 
             and Balsters, Joshua H. and Baxter, Leslie and Beggiato, Anita 
             and Bernaerts, Sylvie and Blanken, Laura M. E. and Bookheimer, Susan Y. 
             and Braden, B. Blair and Byrge, Lisa and Castellanos, F. Xavier 
             and Dapretto, Mirella and Delorme, Richard and Fair, Damien A. 
             and Fishman, Inna and Fitzgerald, Jacqueline and Gallagher, Louise 
             and Keehn, R. Joanne Jao and Kennedy, Daniel P. and Lainhart, Janet E. 
             and Luna, Beatriz and Mostofsky, Stewart H. and M{\"u}ller, Ralph-Axel 
             and Nebel, Mary Beth and Nigg, Joel T. and O'Hearn, Kirsten 
             and Solomon, Marjorie and Toro, Roberto and Vaidya, Chandan J. 
             and Wenderoth, Nicole and White, Tonya and Craddock, R. Cameron 
             and Lord, Catherine and Leventhal, Bennett and Milham, Michael P.},
  title   = {Enhancing studies of the connectome in autism using the 
             autism brain imaging data exchange {II}},
  journal = {Sci. Data},
  volume  = {4},
  number  = {1},
  pages   = {170010},
  year    = {2017},
  doi     = {10.1038/sdata.2017.10},
  issn    = {2052-4463},
}

@misc{oasis3_dataset,
  author  = {LaMontagne, Pamela J. and Benzinger, Tammie L.S. and Morris, John C. 
             and Keefe, Sarah and Hornbeck, Russ and Xiong, Chengjie 
             and Grant, Elizabeth and Hassenstab, Jason and Moulder, Krista 
             and Vlassenko, Andrei and Raichle, Marcus E. and Cruchaga, Carlos 
             and Marcus, Daniel},
  title   = {{OASIS-3}: Longitudinal neuroimaging, clinical, and cognitive dataset 
             for normal aging and {Alzheimer} disease [dataset]},
  year    = {2019},
  howpublished = {medRxiv preprint},
  doi     = {10.1101/2019.12.13.19014902},
}

@misc{BraTS24_dataset,
 author       = {de Verdier, Maria Correia and Saluja, Rachit and Gagnon, Louis and LaBella, Dominic and Baid, Ujjwall and Tahon, Nourel Hoda and Foltyn-Dumitru, Martha and Zhang, Jikai and Alafif, Maram and Baig, Saif and Chang, Ken and D'Anna, Gennaro and Deptula, Lisa and Gupta, Diviya and Haider, Muhammad Ammar and Hussain, Ali and Iv, Michael and Kontzialis, Marinos and Manning, Paul and Moodi, Farzan and Nunes, Teresa and Simon, Aaron and Sollmann, Nico and Vu, David and Adewole, Maruf and Albrecht, Jake and Anazodo, Udunna and Chai, Rongrong and Chung, Verena and Faghani, Shahriar and Farahani, Keyvan and Fathi Kazerooni, Anahita and Iglesias, Eugenio and Kofler, Florian and Li, Hongwei and Linguraru, Marius George and Menze, Bjoern and Moawad, Ahmed W. and Velichko, Yury and Wiestler, Benedikt and Altes, Talissa and Basavasagar, Patil and Bendszus, Martin and Brugnara, Gianluca and Cho, Jaeyoung and Dhemesh, Yaseen and Fields, Brandon K.K. and Garrett, Filip and Gass, Jaime and Hadjiiski, Lubomir and Hattangadi-Gluth, Jona and Hess, Christopher and Houk, Jessica L. and Isufi, Edvin and Layfield, Lester J. and Mastorakos, George and Mongan, John and Nedelec, Pierre and Nguyen, Uyen and Oliva, Sebastian and Pease, Matthew W. and Rastogi, Aditya and Sinclair, Jason and Smith, Robert X. and Sugrue, Leo P. and Thacker, Jonathan and Vidic, Igor and Villanueva-Meyer, Javier and White, Nathan S. and Aboian, Mariam and Conte, Gian Marco and Dale, Anders and Sabuncu, Mert R. and Seibert, Tyler M. and Weinberg, Brent and Abayazeed, Aly and Huang, Raymond and Turk, Sevcan and Rauschecker, Andreas M. and Farid, Nikdokht and Vollmuth, Philipp and Nada, Ayman and Bakas, Spyridon and Calabrese, Evan and Rudie, Jeffrey D.},
  title   = {The 2024 {Brain} {Tumor} {Segmentation} ({BraTS}) challenge: 
             Glioma segmentation on post-treatment {MRI} [dataset]},
  year    = {2024},
  howpublished    = {arXiv Preprint},
  doi     = {10.48550/arXiv.2405.18368},
}

@article{mood_dataset,
  author  = {Zimmerer, David and Full, Peter M. and Isensee, Fabian and Jäger, Paul and Adler, Tim and Petersen, Jens and Köhler, Gregor and Ross, Tobias and Reinke, Annika and Kascenas, Antanas and Jensen, Bjørn Sand and O’Neil, Alison Q. and Tan, Jeremy and Hou, Benjamin and Batten, James and Qiu, Huaqi and Kainz, Bernhard and Shvetsova, Nina and Fedulova, Irina and Dylov, Dmitry V. and Yu, Baolun and Zhai, Jianyang and Hu, Jingtao and Si, Runxuan and Zhou, Sihang and Wang, Siqi and Li, Xinyang and Chen, Xuerun and Zhao, Yang and Marimont, Sergio Naval and Tarroni, Giacomo and Saase, Victor and Maier-Hein, Lena and Maier-Hein, Klaus},
  title   = {{MOOD} 2020: A Public Benchmark for Out-of-Distribution 
             Detection and Localization on Medical Images},
  journal = {IEEE Trans. Med. Imaging},
  volume  = {41},
  number  = {10},
  pages   = {2728--2738},
  year    = {2022},
  doi     = {10.1109/TMI.2022.3170077},
}

@article{ucsf_dataset,
  title={The University of California San Francisco Preoperative Diffuse Glioma {MRI} Dataset},
  author={Calabrese, Evan and Villanueva-Meyer, Javier E and Rudie, Jeffrey D and Rauschecker, Andreas M and Baid, Ujjwal and Bakas, Spyridon and Cha, Soonmee and Mongan, John T and Hess, Christopher P},
  journal={Radiol. Artif. Intell.},
  volume={4},
  number={6},
  pages={220058},
  year={2022},
  publisher={Radiological Society of North America},
  doi={10.1148/ryai.220058}
}

@misc{scan_dataset,
  title        = {{SCAN}: Standardized Centralized {Alzheimer's} 
                  and Related Dementias Neuroimaging [dataset]},
  author       = {{National Alzheimer's Coordinating Center}},
  year         = {2020},
  howpublished = {\url{https://scan.naccdata.org}},
  note         = {NIA/NIH Grant U24 AG067418. Accessed: June 2026}
}

@inproceedings{brainjepa,
  author    = {Dong, Zijian and Li, Ruilin and Wu, Yilei and Nguyen, Thuan Tinh and Chong, Joanna Su Xian and Ji, Fang and Tong, Nathanael Ren Jie and Chen, Christopher Li Hsian and Zhou, Juan Helen},
  title     = {Brain-{JEPA}: Brain Dynamics Foundation Model with Gradient Positioning and Spatiotemporal Masking},
  booktitle = {Advances in Neural Information Processing Systems (NeurIPS)},
  volume    = {37},
  pages     = {86048--86073},
  year      = {2024},
  doi       = {10.48550/arXiv.2409.19407}
}

@misc{huang,
  author       = {Huang, Haoxu and Chen, Long and Chen, Jingyun and Hyun, Jinu and Loftus, James Ryan and Melmed, Kara and Orringer, Daniel and Frontera, Jennifer and Dehkharghani, Seena and Masurkar, Arjun and Razavian, Narges},
  title        = {Learning Sparse Latent Predictive Foundation Model for Multimodal Neuroimaging},
  year         = {2026},
  howpublished = {arXiv Preprint},
  doi          = {10.48550/arXiv.2606.14957}
}

@InProceedings{kornblith2019similarity,
  title     = {Similarity of Neural Network Representations Revisited},
  author    = {Kornblith, Simon and Norouzi, Mohammad and Lee, Honglak and Hinton, Geoffrey},
  booktitle = {Proceedings of the 36th International Conference on Machine Learning},
  pages     = {3519--3529},
  year      = {2019},
  editor    = {Chaudhuri, Kamalika and Salakhutdinov, Ruslan},
  volume    = {97},
  series    = {Proceedings of Machine Learning Research},
  month     = {09--15 Jun},
  publisher = {PMLR}
}

@inproceedings{garrido2023rankme,
author = {Garrido, Quentin and Balestriero, Randall and Najman, Laurent and LeCun, Yann},
title = {RankMe: assessing the downstream performance of pretrained self-supervised representations by their rank},
year = {2023},
publisher = {JMLR.org},
booktitle = {Proceedings of the 40th International Conference on Machine Learning},
articleno = {440},
numpages = {46},
series = {ICML'23},
doi = {10.5555/3618408.3618848}
}

@article{Tak2026,
  author = {Tak, Divyanshu and
            Garomsa, Biniam A. and
            Zapaishchykova, Anna and
            Chaunzwa, Tafadzwa L. and
            Climent Pardo, Juan Carlos and
            Ye, Zezhong and
            Zielke, John and
            Ravipati, Yashwanth and
            Pai, Suraj and
            Vajapeyam, Sri and
            Mahootiha, Maryam and
            Parker, Mitchell and
            Pike, Luke R. G. and
            Smith, Ceilidh and
            Familiar, Ariana M. and
            Liu, Kevin X. and
            Prabhu, Sanjay and
            Arnaout, Omar and
            Bandopadhayay, Pratiti and
            Nabavizadeh, Ali and
            Mueller, Sabine and
            Aerts, Hugo JWL and
            Huang, Raymond Y. and
            Poussaint, Tina Y. and
            Kann, Benjamin H.},
  title   = {A generalizable foundation model for analysis of human brain {MRI}},
  journal = {Nature Neuroscience},
  year    = {2026},
  volume  = {29},
  number  = {4},
  pages   = {945--956},
  doi     = {10.1038/s41593-026-02202-6}
}


\clearpage
\includepdf[pages=-,pagecommand={}]{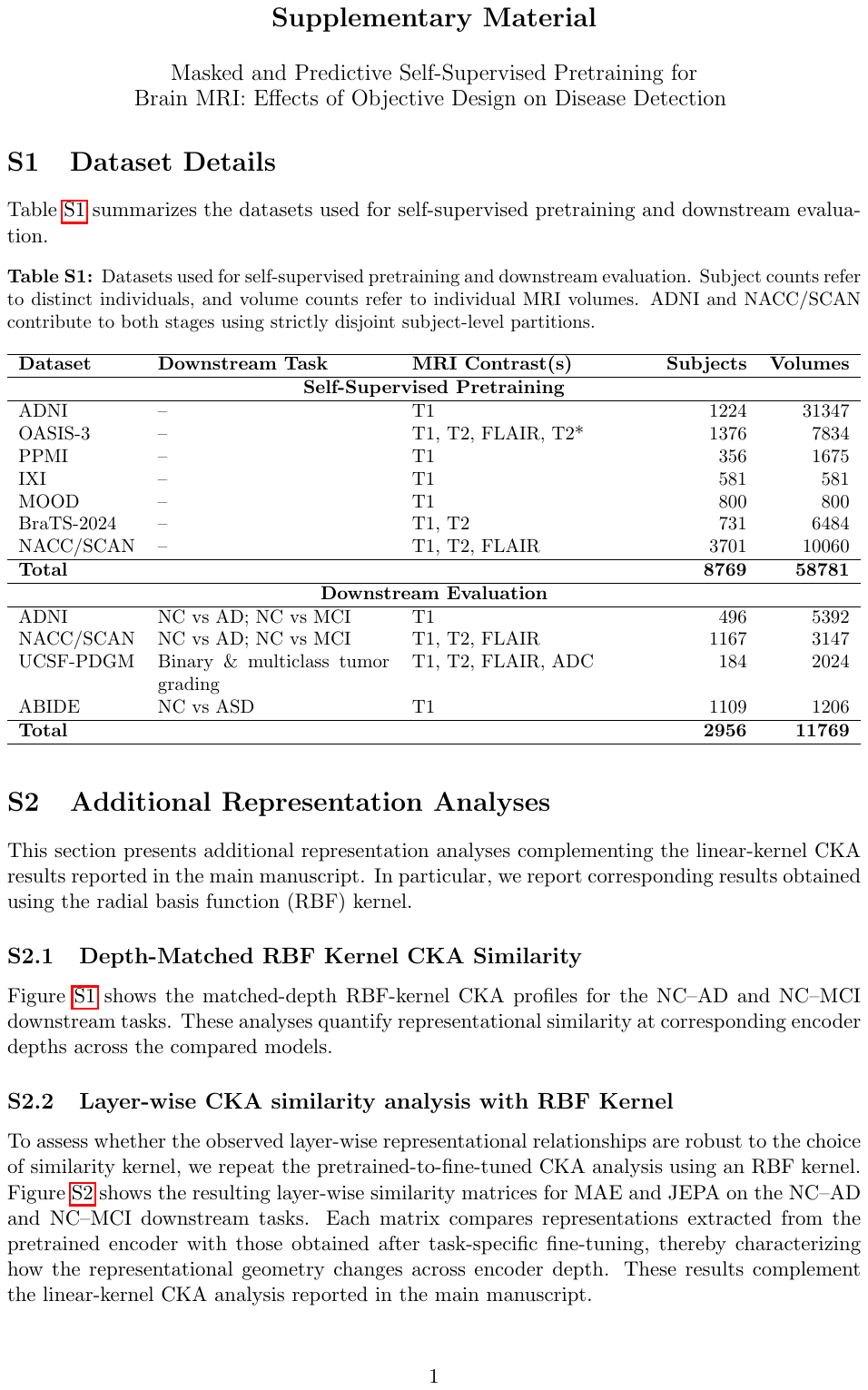}
\end{document}